\documentclass[runningheads]{llncs}

\usepackage{eccv}

\usepackage{eccvabbrv}
\usepackage{algorithm}
\usepackage{algpseudocode}
\usepackage{amsmath}
\usepackage{amssymb}
\usepackage{graphicx}
\usepackage{booktabs}
\usepackage{multirow}

\usepackage{diagbox}
\usepackage{pifont}%
\newcommand{\cmark}{\textcolor{green}{\checkmark}} %
\newcommand{\halfcmark}{\textcolor{blue}{\checkmark\kern-1.1ex\raisebox{.7ex}{\rotatebox[origin=c]{125}{--}}}} %
\newcommand{\xmark}{\textcolor{red}{\texttimes}}  %
\newcommand{\cmstar}{\textcolor{blue}{\checkmark*}} %
\newcommand{\gcmark}{\textcolor{gray}{\checkmark}} %
\newcommand{\gxmark}{\textcolor{gray}{\texttimes}}  %
\newcommand{\gcmstar}{\textcolor{gray}{\checkmark*}} %

\usepackage{arydshln}

\usepackage[accsupp]{axessibility}  %
\newcommand{\method}{{ReViV}\xspace}

\usepackage{hyperref}

\usepackage{orcidlink}

\begin{document}

\title{ReViV: Reconstructing the Viewer and the View in 4D from Monocular Egocentric Video} 

\titlerunning{ReViV: Reconstructing the Viewer and the View in 4D from Ego Video}

\author{Xiaozhong Lyu\inst{1}\thanks{Equal contribution; order interchangeable. See \nameref{sec:contrib}.}\orcidlink{0009-0008-6857-3311} \and
Gen Li\inst{1}\textsuperscript{$\star$}\orcidlink{0009-0005-9782-7649} \and
Zhiyin Qian\inst{1}\orcidlink{0009-0008-3975-0016} \and \newline
Xucong Zhang\inst{1,2}\orcidlink{0000-0002-8368-3542} \and
Marc Pollefeys\inst{1,3}\orcidlink{0000-0003-2448-2318} \and
Siyu Tang\inst{1}\orcidlink{0000-0002-1015-4770}
}

\authorrunning{X.~Lyu et al.}

\institute{ETH Zurich, Switzerland \and
Delft University of Technology, Netherlands \and
Microsoft, Switzerland}

\maketitle

\begin{abstract}
Egocentric devices, such as wearable front-facing cameras, provide a unique perspective for capturing the continuous interaction between a human viewer and the surrounding environment. 
A holistic and efficient multimodal model capable of reconstructing this 4D representation is therefore highly desirable. 
However, existing approaches often rely on auxiliary inputs such as pre-computed camera trajectories, treat scene perception and human ego-motion modeling as separate problems despite their strong interdependency, and suffer from slow inference time.
To address these limitations, we present \method, the first unified framework for holistic egocentric 4D reconstruction that extracts both viewer and view dynamics from a single monocular RGB video.
We formulate the task as learning the full joint probability distribution over multimodal signals, including RGB video, camera trajectory, gaze direction, full-body motion, hand motion, and depth.
Powered by a Masked Generative Egocentric Transformer, \method operates within a single feed-forward architecture to simultaneously reconstruct the temporally consistent 4D reconstruction across the viewer and the view with fast inference speed.
Extensive experiments on diverse benchmarks, including HoloAssist, HOT3D, ARCTIC, Aria Digital Twin, and TACO, demonstrate that \method achieves state-of-the-art accuracy and efficiency across holistic ego-body, hand, and gaze reconstruction, camera tracking, while maintaining highly competitive egocentric depth estimation, without relying on heavy task-specific priors. Code and models are fully open-sourced: \url{https://reviv4d.github.io/}.

\keywords{Egocentric Vision \and 4D Reconstruction \and Human Motion}
\end{abstract}   
\section{Introduction}

\begin{figure}[t]
    \centering
    \includegraphics[width=1.0\linewidth]{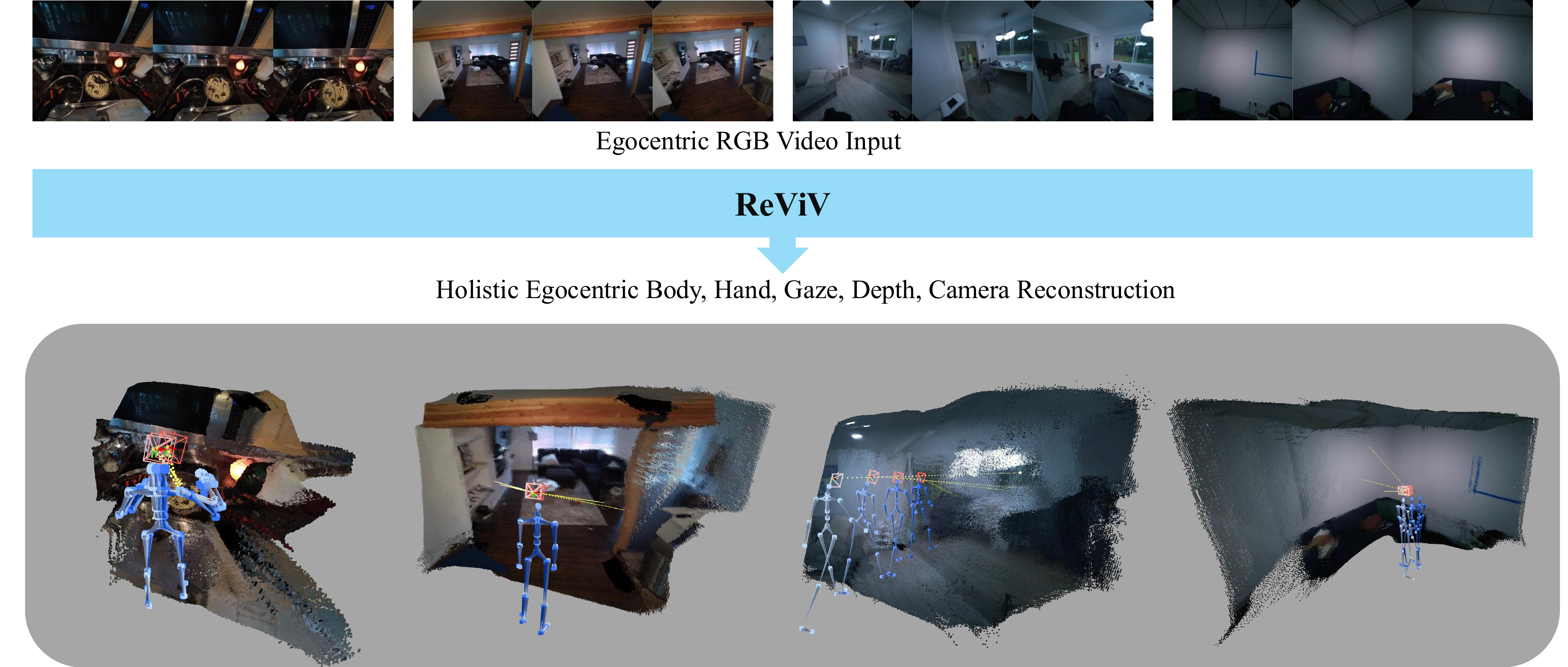}
    \caption{We present \textbf{\method}, a unified framework for holistic egocentric 4D reconstruction. Given a monocular egocentric RGB video, our method jointly estimates human-centric modalities, including \textbf{\textcolor{blue}{body}}, \textbf{\textcolor{blue}{hand}} pose, and \textbf{\textcolor{Goldenrod}{gaze}}, alongside scene-aware \textbf{\textcolor{red}{camera trajectory}} and \textbf{depth estimation}. These predictions are integrated into a temporally consistent viewer–view reconstruction from a monocular egocentric video.
    }
    \label{fig:teaser}
\end{figure}

Egocentric vision lies at the core of embodied intelligence, as it captures the world directly from the first-person perspective through which humans perceive, act, and interact \cite{he2026bridging, rodin2021predicting}. With the rapid proliferation of wearable devices such as smart glasses, body-mounted cameras, and head-mounted displays, egocentric video has become increasingly prevalent in augmented and virtual reality \cite{tome2020selfpose}, assistive robotics \cite{xin2023learning}, and human–computer interaction. In these scenarios, systems must not only understand the surrounding environment, but also interpret the wearer’s own actions, gestures, and intentions in real time \cite{jiang2017seeing, ng2020you2me, Grauman_2024_CVPR}. 

Despite the unique perspective offered by egocentric vision, reconstructing the viewer’s actions and scene remains insufficiently explored \cite{grauman2022ego4d, damen2020epic}. 
Existing egocentric methods typically focus on either environmental reconstruction \cite{li2025egom2p, zhang2025egogaussian, egomono4d} or human pose estimation \cite{li2023ego, yuan2019ego, luo2021dynamics, li2024egogen, guzov-jiang2025hmd2, yi2025egoallo, patel2025uniegomotion}.
As a result, scene dynamics and human motion are inferred independently, ignoring the temporal constraints that naturally couple the observer and the observed world.
Moreover, many of the egocentric human pose estimation approaches depend on auxiliary inputs, such as on-device SLAM trajectories \cite{patel2025uniegomotion, yi2025egoallo}, precomputed 3D point clouds \cite{yi2025egoallo}, or dedicated hand-tracking modules \cite{yi2025egoallo}, limiting their applicability to specialized hardware setups.
By treating scene understanding and human reconstruction in isolation, current approaches
often result in temporally inconsistent view geometry and the viewer's motion \cite{duan20244d, yang2024deformable}. Therefore, a holistic framework that reconstructs the viewer’s human motion while jointly modeling the contextual 4D reconstructions, such as depth sequences, gaze, and camera trajectories, from monocular egocentric video remains an open and challenging problem.

Recent advances in 4D reconstruction have achieved impressive results in third-person or multi-view settings, reconstructing dynamic scenes and human motion from multiple synchronized cameras \cite{duan20244d, yang2024deformable}. Methods such as Shape of Motion \cite{wang2025shape} enable long-range dynamic reconstruction by representing scenes with persistent 3D Gaussians and motion bases, and feed-forward architectures further improve efficiency and generalization by avoiding costly test-time optimization \cite{feng2025st4rtrack, xiao2025spatialtrackerv2, xu20254dgt, chen2025human3r}. However, these advances rely on third-person viewpoints or multi-view capture, leaving the problem of reconstructing the wearer’s own body dynamics in monocular egocentric settings largely unresolved.

To address this challenge, we present \method (see Fig.~\ref{fig:teaser}), a unified framework for \underline{\textbf{Re}}constructing the \underline{\textbf{Vi}}ewer and the \underline{\textbf{V}}iew from a single monocular egocentric RGB video. Rather than treating body motion, hand dynamics, gaze, camera trajectory, and depth as independent prediction tasks, ReViV formulates egocentric reconstruction as masked generative modeling over multimodal signals. At inference time, ReViV conditions on RGB observations and predicts the missing viewer-centric and scene-centric modalities in a single feed-forward pass. Since monocular depth is inherently ambiguous in scale, ReViV predicts temporally consistent affine-invariant scene depth and uses a lightweight alignment step to place the reconstructed viewer and view into a metric 4D coordinate system when suitable geometric anchors are available.

At the core of our framework is a Masked Generative Egocentric Transformer (MGET). To model the complex and partially unobserved kinematics of the viewer, we introduce a unified tokenization scheme that encodes spatiotemporal signals from diverse modalities into compact latent representations. MGET is trained with a unified multi-task objective that learns intrinsic human motion dynamics and cross-modal correlations by predicting randomly masked tokens conditioned on available context. By masking subsets of multimodal tokens during training, the model learns to infer missing kinematics and scene geometry from visible cues, capturing the underlying joint distribution across modalities. 

Extensive experiments on multiple benchmarks demonstrate that \method achieves state-of-the-art performance in holistic ego-body, hand, and gaze reconstruction while maintaining competitive accuracy in depth estimation and camera tracking. By scaling our multimodal dataset to 7B unique tokens and optimizing over 500B training tokens, the model learns robust cross-modal dependencies that yield temporally consistent viewer–view dynamics. \method's feed-forward architecture enables notably faster inference than optimization-based methods~\cite{huang2025vipe, egomono4d}. Our contributions are:
\begin{itemize}
\item A unified framework for egocentric 4D reconstruction from a single monocular RGB video, holistically modeling scene, body, hand, and gaze.
\item A large-scale feed-forward pretrained model that explicitly models human motion, improving scalability and generalization in egocentric settings.
\item Comprehensive benchmarking on downstream egocentric tasks, demonstrating that \method achieves state-of-the-art accuracy in human body, hand, and gaze reconstruction along with highly competitive depth estimation.
\end{itemize}

\section{Related Work}

\subsection{4D Reconstruction}

Dynamic 4D reconstruction recovers temporally coherent 3D structures of dynamic scenes over time.  
Early works mainly focused on capturing 3D motion trajectories of objects \cite{wang2025continuous, lu2025align3r}.  
These approaches have evolved from static 3D scene reconstruction toward dynamic scene modeling using synchronized multi-view video inputs \cite{luiten2024dynamic, wu20244d, duan20244d, yang2024deformable}.  
Recent advances further relax the multi-view requirement by performing 4D reconstruction from monocular videos \cite{chu2024dreamscene4d, lei2025mosca, qingming2025modgs, stearns2024dynamic, wang2025gflow}.  

To model long-term dynamics, \cite{wang2025shape} represents dynamic scenes as persistent 3D Gaussians with compact motion bases, achieving efficient long-range reconstruction.  
However, Gaussian-based representations still rely on per-scene test-time optimization, limiting their generalization.  
To address this, recent feed-forward methods jointly reconstruct and track dynamic scenes, offering fast and generalized solutions without optimization at inference time \cite{feng2025st4rtrack, xiao2025spatialtrackerv2, xu20254dgt}.  
Despite these advances, existing 4D reconstruction methods mainly target third-person or multi-view settings, leaving egocentric domain largely unexplored.

\subsection{Scene Reconstruction from Egocentric Videos}

Scene reconstruction from egocentric perspectives poses unique challenges due to the limited field of view, frequent camera motion, and strong occlusions.  
EgoM2P \cite{li2025egom2p} introduces a multitask pretraining framework for egocentric vision that jointly models RGB, depth, gaze, and camera trajectories, enabling multimodal scene understanding.  
EgoGaussian \cite{zhang2025egogaussian} extends this idea by simultaneously reconstructing 3D scenes and dynamically tracking object motion directly from monocular egocentric RGB inputs.  
EgoMono4D \cite{egomono4d} applies self-supervised learning for point cloud sequence reconstruction to the label-scarce egocentric field.
These works demonstrate the potential of egocentric video for holistic scene modeling, yet they focus mainly on scene or object-level geometry, without explicitly reconstructing full human motion and interaction dynamics.

\subsection{Human Motion Estimation from Egocentric Videos}

Estimating human motion from egocentric videos has been explored in several directions.  
Some works employ fisheye cameras to predict full-body pose~\cite{xu2019mo, jiang2021egocentric, tome2019xr, wang2021estimating, li2026egoposeformer, deshmukh20263dpsm}, though image distortion and self-occlusion remain major challenges.  
EgoEgo \cite{li2023ego} estimates 3D human motion by first predicting head movement from the egocentric video and then inferring the full-body motion from it, however, it does not recover hand poses.  
Ego-Pose \cite{yuan2019ego} estimates and forecasts a person’s pose sequence from egocentric input using reinforcement learning, while \cite{luo2021dynamics} introduces an object-aware 3D egocentric pose estimation framework.  
More recently, generative models have significantly advanced egocentric pose estimation. 
HMD$^2$ \cite{guzov-jiang2025hmd2} and EgoAllo \cite{yi2025egoallo} employ conditional diffusion models to generate full-body kinematics; additionally, EgoAllo estimates hand movements by integrating off-the-shelf hand priors \cite{pavlakos2024reconstructing} through diffusion guidance.
Similarly, UniEgoMotion \cite{patel2025uniegomotion} proposes a unified diffusion framework for motion reconstruction and forecasting. While these methods achieve impressive kinematic accuracy, they fundamentally rely on supplementary inputs, such as pre-computed SLAM trajectories, 3D point clouds, or external hand trackers. This reliance on specialized hardware limits their applicability to casual, in-the-wild monocular videos.

Furthermore, gaze cues have been integrated into egocentric motion understanding.  
\cite{yungaze2025} forecasts future gaze in the 3D scene from 2D visual inputs, and \cite{pani2025gaze} proposes a gaze-regularized attention mechanism that enhances vision-language models (VLMs) for egocentric behavior understanding, particularly for activity recognition and future prediction.  

Together, these efforts highlight the growing interest in egocentric human motion and scene modeling, yet a unified approach for 4D reconstruction that jointly models human body, hand, and gaze dynamics along with scene geometry remains an open challenge. 
In this work, we propose \method, the first unified generative framework for 4D reconstruction in the egocentric domain.

\section{Data Engine}

\begin{table}[t]
    \centering
    \caption{\textbf{Large-Scale Multi-Modal Pretraining Dataset.} Building upon the scene-centric foundation of EgoM2P \cite{li2025egom2p}, our dataset includes dense hand and full-body kinematics to capture holistic human-scene interactions. By scaling the total pretraining corpus from 4B to 7B unique tokens, \method establishes a new data foundation for egocentric 4D understanding. (Note: \cmark: available, \xmark: unavailable. \cmstar: our generated pseudo-labels. Grey: baseline usage. Color: newly integrated datasets/modalities.}
    \setlength{\tabcolsep}{2.5pt} %
    \renewcommand{\arraystretch}{1.} %
    \begin{tabular}{l|cccc:cc}
        \toprule
        \diagbox[width=8em, height=2em]{\tiny\textbf{Datasets}}{\tiny\textbf{Modalities}} 
        & RGB & Depth & Gaze & Camera & Hand & Body\\
        \midrule
        EgoExo4D \cite{Grauman_2024_CVPR} & \gcmark & \gxmark   & \gcmark & \gcmark & \xmark & \xmark\\
        HoloAssist \cite{HoloAssist2023}    & \gcmark & \gcmstar & \gcmark & \gcmark & \cmark & \xmark\\
        HOT3D (Aria) \cite{banerjee2024hot3d} & \gcmark & \gcmstar  & \gcmark & \gcmark & \cmark & \xmark\\
        HOT3D (Quest) \cite{banerjee2024hot3d} & \gcmark & \gcmstar  & \gxmark & \gcmark & \cmark & \xmark\\
        ARCTIC \cite{fan2023arctic}         & \gcmark & \gcmstar   & \gxmark & \gcmark & \cmark & \xmark\\
        TACO \cite{liu2024taco}                & \gcmark & \gcmstar  & \gxmark & \gcmark & \cmark & \xmark\\
        H2O \cite{Kwon_2021_ICCV}           & \gcmark & \gcmark    & \gxmark & \gcmark & \cmark & \xmark\\
        EgoGen \cite{li2024egogen} & \gcmark & \gcmark  & \gxmark & \gcmark & \xmark & \xmark\\
        \cdashline{1-5}
        Nymeria \cite{ma2024nymeria} & \cmark & \cmstar  & \cmark & \multicolumn{1}{c}{\cmark} & \xmark & \cmark \\
        \bottomrule
    \end{tabular}
    \label{tab:datasets}
\end{table}

One major limitation of prior work lies in the lack of a unified data organization for multimodal pretraining, which prevents training a single framework across heterogeneous egocentric datasets.
To train \method at scale (Sec. \ref{sec:method}), we construct an automated multi-modal data engine that harmonizes spatial representations and compensates for missing modalities across diverse datasets (Tab.~\ref{tab:datasets}). Building upon the scene-centric foundation of EgoM2P \cite{li2025egom2p}, we extend the corpus with dense hand and full-body kinematics, increasing the pretraining set from 4B to a unified 7B-token dataset tailored for holistic egocentric 4D modeling.

\subsection{Temporally Consistent Geometric Pseudo-Labeling}
Many egocentric datasets lack dense 3D geometry due to hardware constraints or incomplete sensor streams. To address this, we generate high-quality, temporally consistent video depth pseudo-labels using Video Depth Anything \cite{video_depth_anything}. This automated pipeline enables us to scale geometric supervision beyond the constrained hand-object interactions used in EgoM2P \cite{li2025egom2p} to diverse, unconstrained real-world environments. The resulting large-scale depth annotations provide consistent geometric signals over time, allowing \method to learn robust 4D structural priors. As shown in our evaluations, this large-scale geometric pretraining improves video depth prediction performance in complex scenes compared to existing baselines.

\subsection{Unified Kinematic Representation}
Egocentric datasets differ significantly in coordinate conventions and sensor setups, leading to inconsistencies in kinematic annotations. To enable unified modeling, we standardize all motion signals into two complementary reference frames that disentangle local manipulation from global navigation.

\noindent\textbf{Camera-Space Hand Kinematics.} We project the hand joints into the camera space of their current respective frames. While dependent on the immediate camera pose, this provides a trajectory-invariant representation relative to the user's viewpoint, encouraging the model to learn interaction priors independent of global body translation.

\noindent\textbf{Gravity-Aligned Global Body Kinematics.} Anchoring trajectories to a raw initial camera pose introduces pitch and roll artifacts. Instead, we establish a stable world reference frame by taking the first frame's camera pose and projecting its up-vector to align opposite to gravity. This creates a ground-parallel reference frame that isolates true global navigation from the camera's initial tilt.
\begin{figure*}[t]
    \centering
    \includegraphics[ width=1.0\linewidth]{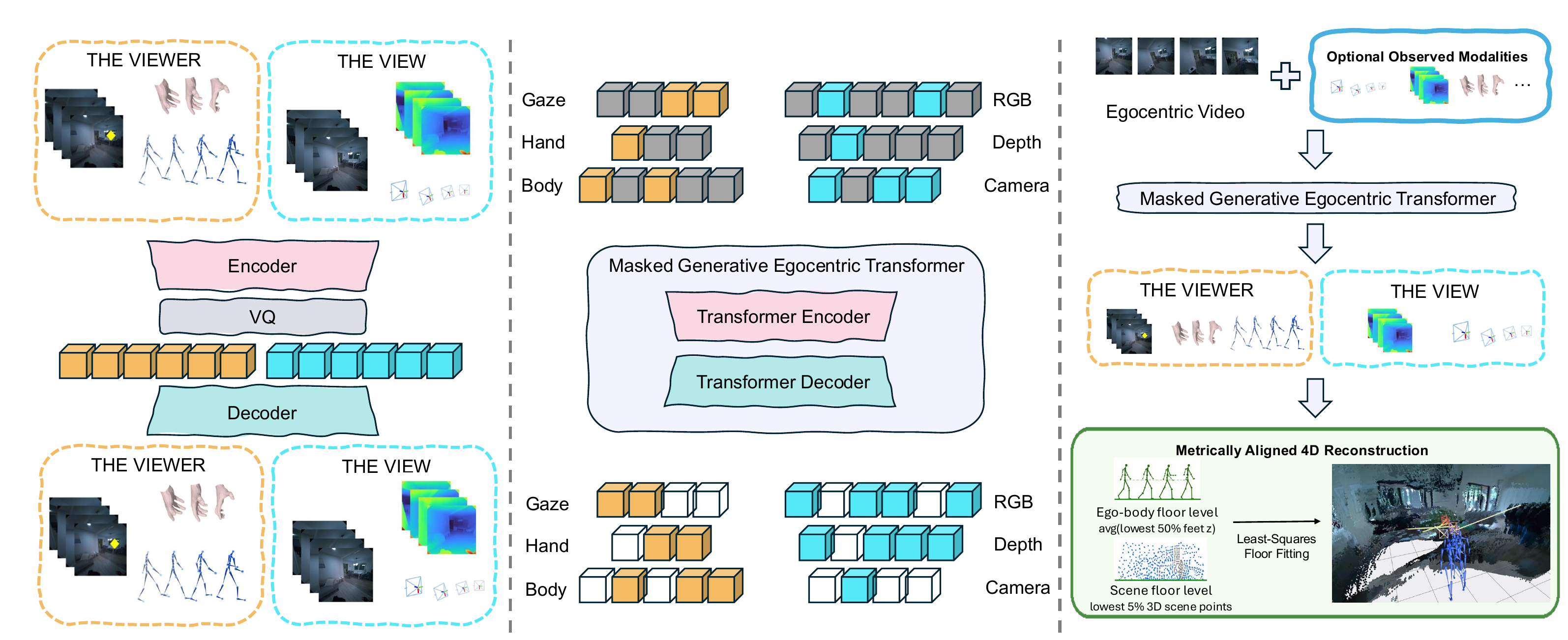}
    \caption{\method Architecture. \textbf{(Left)} Modality-specific VQ-VAEs discretize heterogeneous viewer (gaze, hand, body) and view (RGB, depth, camera) continuous modalities into a unified token sequence. \textbf{(Middle)} A Masked Generative Egocentric Transformer (MGET) learns the joint probability distribution by predicting randomly masked tokens (grey blocks), capturing both intra- and cross-modal dynamics. \textbf{(Right)} During inference, MGET conditions on available observed tokens (e.g., RGB) to iteratively decode and reconstruct the unobserved human and scene states. Floor fitting aligns the reconstruction to a metric 4D coordinate system.
    }
    \label{fig:method}
\end{figure*}

\section{Method}
\label{sec:method}

We propose a unified framework to jointly reconstruct human-centric (gaze, body, hand motion) and scene-centric (camera trajectories, depth maps) features from monocular egocentric video (Fig \ref{fig:method}). Because the viewer's body is heavily occluded in egocentric views, deterministic estimation is inherently ambiguous. To address this, we formulate the task as a generative modeling problem and approximate the multimodal joint distribution through masked token prediction.

\subsection{Problem Formulation}
Let $\mathcal{X}=\{\mathbf{x}_{\text{rgb}}\}$ represent the observed egocentric video context, and $\mathcal{Y}=\{\mathbf{y}_{\text{hand}}, \mathbf{y}_{\text{body}}, \mathbf{y}_{\text{gaze}}, \mathbf{y}_{\text{depth}}, \mathbf{y}_{\text{cam}}\}$ denote the human and scene dynamic states to reconstruct.
Directly learning a deterministic mapping $f: \mathcal{X} \rightarrow \mathcal{Y}$ is ill-posed due to the severe occlusion of the ego-body, which renders the underlying solution space highly multimodal. To address this, we cast the problem as learning the \textit{joint probability distribution} across all observed and unobserved modalities: 

\begin{equation}
    p(\mathcal{X}, \mathcal{Y}) = p(\mathbf{x}_{\text{rgb}}, \mathbf{y}_{\text{hand}}, \mathbf{y}_{\text{body}}, \mathbf{y}_{\text{gaze}}, \mathbf{y}_{\text{depth}}, \mathbf{y}_{\text{cam}}).
\end{equation}

In practice, we approximate this joint distribution through masked multimodal token prediction, which enables the model to learn both intra-modal temporal dynamics and cross-modal correlations. During inference, this allows us to flexibly sample from the conditional distribution $p(\mathcal{Y} \mid \mathcal{X})$ to reconstruct the human and scene states.

\noindent\textbf{Metric Alignment of 4D Reconstruction.} To achieve a metric-aligned 4D reconstruction, we introduce a lightweight alignment step to map the relative scene depth into the metric domain. Specifically, we fit a stable floor plane from the reconstructed scene geometry and use it to regularize the metric scale, with details in supplementary Sec. D. For RGB inputs that do not contain a visible floor, we leverage off-the-shelf metric depth predictions from VIPE~\cite{huang2025vipe} as an optional geometric anchor. These alignment steps allow the reconstructed viewer and view to be placed within a shared, metric-aligned 4D coordinate system.

\subsection{Unified Discrete Representation}
Modeling the joint distribution $p(\mathcal{X}, \mathcal{Y})$ in the continuous domain is challenging due to the heterogeneity of the modalities, which range from dense pixel arrays to sparse kinematic vectors. We therefore leverage a unified discrete latent interface to homogenize the modalities.

For each modality $m \in \mathcal{X} \cup \mathcal{Y}$ with continuous input dimension $D_m$, we employ a modality-specific Vector Quantized Variational Autoencoder (VQ-VAE) to learn a quantization function $Q_m: \mathbb{R}^{D_m} \rightarrow \mathcal{Z}_m$. This function maps the continuous signal to a sequence of $L_m$ discrete tokens $\mathbf{z}_m = (z_{m,1}, \dots, z_{m,L_m})$. These tokens are indices drawn from a learned codebook $\mathcal{C}_m \in \mathbb{R}^{K_m \times d}$, where $K_m$ denotes the vocabulary size for modality $m$ and $d$ is the latent feature dimension.
The full multi-modal state is represented as a unified sequence of tokens $\mathbf{Z}$:

\begin{equation}
    \mathbf{Z}=[\mathbf{z}_{\text{rgb}}, \mathbf{z}_{\text{depth}}, \mathbf{z}_{\text{cam}}, \mathbf{z}_{\text{gaze}}, \mathbf{z}_{\text{hand}}, \mathbf{z}_{\text{body}}]
\end{equation}

For human body and hand tokenization, to address the severe scarcity of high-quality, unified whole-body datasets, we design a decoupled, dual-stream tokenization framework capable of leveraging disjoint body-only and hand-only motion captures. Given an input 3D joint sequence $\mathbf{y}_m \in \mathbb{R}^{N\times J \times 3}, m\in\{\text{hand},\text{body}\}$ consisting of $N$ frames and $J$ joints, our body and hand tokenizers independently apply 2D convolutions to extract local kinematic priors. These features are subsequently processed by 12-block Transformer encoders, projecting the raw kinematics into a dense, continuous latent space of 210-dimensional embeddings.
To quantize these representations while ensuring high dictionary utilization across highly varied datasets, we leverage a spherical quantization codebook. By projecting the continuous embeddings onto a unit sphere and pairing this with an Exponential Moving Average (EMA) codebook update strategy, we effectively prevent codebook collapse and maintain an expressive discrete latent space for both full-body and fine-grained hand domains. Symmetric 12-block Transformer decoders then reconstruct the continuous sequences, optimized via standard $L_2$ reconstruction and codebook commitment losses. For the dense visual modalities, we quantize RGB and depth videos using Cosmos Tokenizers \cite{agarwal2025cosmos}. Furthermore, to capture a broader distribution of other ego-state dynamics, we re-train the gaze and camera trajectory tokenizers from EgoM2P \cite{li2025egom2p} on our scaled-up 7B-token dataset (Tab. \ref{tab:datasets}). This discretization transforms the continuous regression problem into a sequence modeling task defined over modality-specific discrete vocabularies ($\mathcal{C}_m$), ensuring that predictions for each modality remain constrained to their respective codebook spaces.

\subsection{Masked Generative Egocentric Transformer}

We approximate the joint distribution $p(\mathbf{Z})$ using a Masked Generative Egocentric Transformer (MGET) parameterized by $\theta$. Specifically, we adapt an encoder-decoder architecture based on T5 \cite{2020t5}, comprising 12 Transformer blocks in both the encoder and decoder, with a hidden dimension of 768. Because the unified sequence $\mathbf{Z}$ comprises highly heterogeneous data, we first map the discrete tokens to continuous feature vectors via modality-specific embedding layers. To explicitly inject structural and semantic context, we add learnable modality-type embeddings and fixed sinusoidal spatiotemporal positional encodings to the sequence before processing. To compensate for the fine-grained pixel details inevitably lost during discrete quantization, we introduce an additional trainable Vision Transformer (ViT) that directly encodes the raw egocentric video into continuous spatially detailed features $\mathbf{x}_{\text{ViT}}$. These features are incorporated as additional visual context, yielding $\mathcal{X}=\{\mathbf{x}_{rgb}, \mathbf{x}_{\text{ViT}}\}$.

Following the paradigm of masked modeling \cite{chang2022maskgit, yu2023magvit}, we define a training objective that learns the inter-modal dependencies across the unified token sequence $\mathbf{Z}$ representing the full state $\mathcal{X} \cup \mathcal{Y}$. During training, we randomly sample a binary mask $\mathbf{M} \in \{0, 1\}^{|\mathbf{Z}|}$, partitioning the sequence into masked tokens $\mathbf{Z}_\mathcal{M}$ and visible tokens $\mathbf{Z}_{\mathcal{V}}$. By randomly sampling the mask $\mathbf{M}$ across all modalities~\cite{bachmann20244m, li2025egom2p}, the network is forced to learn an ensemble of arbitrary conditional distributions $p(\mathbf{Z}_{\mathcal{M}} \mid \mathbf{Z}_{\mathcal{V}})$. Because a joint distribution can be factorized into a sequence of conditionals, optimizing over all possible mask configurations allows this objective to act as an efficient proxy for learning the underlying joint distribution $p(\mathbf{Z}) \approx p(\mathcal{X}, \mathcal{Y})$: 

\begin{equation}
    \mathcal{L}_{\text{mask}}(\theta) = - \mathbb{E}_{\mathbf{Z}, \mathbf{M}} \left[ \sum_{i \in \mathcal{M}, z_i \notin \mathbf{x}_{\text{ViT}}} \log p_\theta(z_i \mid \mathbf{Z}_{\mathcal{V}}) \right].
\end{equation}

Crucially, although optimized with a single cross-entropy loss, this random multimodal masking strategy serves as a multi-task training objective. First, by masking partial sequences within a single modality, the network learns intra-modal motion dynamics (e.g., $p(\mathbf{z}_{\text{body}, \mathcal{M}} \mid \mathbf{z}_{\text{body}, \mathcal{V}}), p(\mathbf{z}_{\text{hand}, \mathcal{M}} \mid \mathbf{z}_{\text{hand}, \mathcal{V}})$). This forces the model to encode strong temporal kinematic priors, enabling it to generate continuous and plausible motions without relying on explicit anatomical or smoothness regularizers. Second, by masking across different modalities, the network learns robust cross-modal correlations (e.g., inferring unobserved kinematics from visible context, $p(\mathbf{z}_{\text{body}, \mathcal{M}} \mid \mathbf{z}_{\text{rgb}})$). Together, these two implicit tasks enable MGET to capture human kinematics and environment-conditioned interaction patterns directly from the multimodal training distribution.

The Transformer processes the input sequence to capture spatiotemporal dependencies to output latent representations. These features are projected through modality-specific linear heads to predict probability distributions over their respective vocabularies $\mathcal{C}_m$. At inference time, we achieve our primary reconstruction goal by formulating a specific masking condition that provides the observed egocentric video context as the visible tokens ($\mathbf{Z}_{\mathcal{V}} = \mathbf{Z}_{\mathcal{X}}$) and treat the human and scene state modalities as fully masked ($\mathbf{Z}_{\mathcal{M}} = \mathbf{Z}_{\mathcal{Y}}$). We then employ an iterative parallel decoding strategy to progressively sample masked tokens from these distributions, reconstructing the complete, temporally coherent 4D state.
More implementation details are in Supplementary Material Sec. B and C.

\section{Experiment}

\noindent\textbf{Evaluation Details.} For all experiments, we split each sequence into 2-second clips. For video modalities (RGB and depth), we downsample the original 30 FPS to 8 FPS, and resize spatial resolution to 256$\times$ 256. For non-video modalities, including camera, gaze, body and hand, we retain the original 30 FPS.

\subsection{Egocentric Body Motion Reconstruction}
\begin{table*}[t]
\centering
\caption{
\textbf{Egocentric Body Motion Reconstruction on ADT~\cite{Pan_2023_ICCV}.} \method achieves SOTA performance and efficiency purely from monocular video. Even without explicit camera tracking, \method yields superior motion quality (FID), pose accuracy (PA-MPJPE), and pose similarity compared to baselines utilizing either estimated (VIPE) or ground-truth camera trajectories. ↓: lower is better; ↑: higher is better.
}
\label{tab:body_motion_comparison}
\resizebox{\textwidth}{!}{
\begin{tabular}{l|c|ccccc}
\toprule
Methods & Input Cam Traj 
& GA-MPJPE $\downarrow$ 
& PA-MPJPE $\downarrow$ 
& Similarity $\uparrow$ 
& FID $\downarrow$  & Time (s) $\downarrow$\\
\midrule

\multirow{3}{*}{EgoAllo~\cite{yi2025egoallo}}
& None & -- & -- & -- & --  & --\\
& VIPE~\cite{huang2025vipe} & 227.6 & 167.5 & 0.529 & 1.069 & 101.0\\
& GT Cam & 198.4 & 136.8 & 0.556 & 1.160 & 72.5\\
\hdashline

\multirow{3}{*}{UniEgoMotion~\cite{patel2025uniegomotion}}
& None & -- & -- & -- & -- & --\\
& VIPE~\cite{huang2025vipe} & 130.5 & 98.4 & 0.572 & 1.087 & 37.6\\
& GT Cam & \textbf{105.8} & 88.7 & 0.591 & 1.098 & 9.1\\
\hdashline

\multirow{1}{*}{Ours}
& None & \underline{111.5} & \textbf{88.6} & \textbf{0.751} & \textbf{0.442} & \textbf{0.7} \\

\bottomrule
\end{tabular}
}
\end{table*}
\noindent\textbf{Baselines.}
We compare \method against EgoAllo~\cite{yi2025egoallo} and UniEgoMotion~\cite{patel2025uniegomotion}, two recent SOTA diffusion-based models for egocentric full-body motion recovery.

\noindent\textbf{Evaluation Protocol.}
Since our large-scale pretraining corpus lacks ground-truth SMPL(-X)~\cite{SMPL:2015, SMPL-X:2019} annotations, we cannot retrain the baseline models under identical supervision. To ensure a fair comparison and assess generalization to unseen domains, we conduct all quantitative evaluations on the Aria Digital Twin dataset \cite{Pan_2023_ICCV}, which was strictly excluded from our training data. Specifically, we use 3,185 video clips with corresponding 3D body annotations.
Unlike \method, existing baselines cannot operate purely from monocular RGB and require explicit camera trajectories. To ensure a fair comparison when ground-truth trajectories are unavailable, we provide these baselines with consistent camera poses estimated using the state-of-the-art camera tracking method VIPE~\cite{huang2025vipe}.

\noindent\textbf{Metrics.} Absolute joint position errors can be heavily biased by global translation offsets, which come from both the ambiguity of monocular camera tracking and the dynamic articulation between the head-mounted camera and the body's root joint. To isolate true pose accuracy, we report two aligned variants of the Mean Per Joint Position Error (MPJPE). \texttt{GA-MPJPE} applies a single Procrustes alignment (optimizing rotation, translation, and scale) across the entire motion sequence. In contrast, \texttt{PA-MPJPE} applies this alignment independently on a per-frame basis to capture strictly local pose accuracy.

Beyond per-joint errors, we evaluate motion realism and distributional similarity. Following prior generative motion models, we project the sequences into a latent space using the pretrained TMR motion encoder~\cite{petrovich2023tmr}. From these embeddings, we compute the Fréchet Inception Distance (\texttt{FID}) to measure distributional alignment, and the cosine \texttt{Similarity} between paired embeddings to quantify motion-level correspondence.

\noindent\textbf{Results.}
As shown in Table~\ref{tab:body_motion_comparison}, \method establishes a new SOTA for monocular egocentric body motion reconstruction. Compared to baselines using estimated camera trajectories from VIPE \cite{huang2025vipe}, our method dominates across all metrics while being 100$\times$ faster than EgoAllo and over 10$\times$ faster than UniEgoMotion during inference. We attribute this to our unified framework: unlike baselines that suffer from compounding camera-tracking errors, \method directly models the joint distribution $p(\mathcal{X},\mathcal{Y})$, bypassing error-prone intermediate tracking entirely.

Notably, even without camera trajectories as input, \method outperforms baselines that rely on ground-truth cameras across local pose accuracy (\texttt{PA-MPJPE}), semantic correspondence (\texttt{Similarity}), and motion realism (\texttt{FID}). This advantage stems from our discrete tokenization and multimodal masking objective, which forces the model to learn robust, biomechanically plausible kinematic priors rather than over-relying on rigid trajectory inputs. While baselines equipped with perfect external tracking yield slightly better global alignment (\texttt{GA-MPJPE}), \method remains highly competitive, demonstrating that global ego-motion can be effectively learned implicitly from pure monocular video. 
See additional qualitative visualizations in Supplementary Material Sec. E.2.

\subsection{Egocentric Hand Motion Reconstruction}

\begin{table*}[t]
    \centering
    \setlength{\tabcolsep}{4pt}
    \caption{\textbf{Quantitative Comparison of Egocentric Hand Motion Reconstruction.} \method achieves state-of-the-art accuracy across four benchmarks while operating orders of magnitude faster than continuous optimization baselines (e.g., Dyn-HaMR). Metrics evaluated include Global-Aligned (GA-MPJPE), Root-Aligned (RA-MPJPE), and Procrustes-Aligned (PA-MPJPE) errors (mm), computed per hand in the camera frame. All clips are evaluated without visibility filtering, including occluded and out-of-view hands. ↓ indicates lower is better.}
    
    \label{tab:egocentric_reconstruction}
    
    \resizebox{\textwidth}{!}{
    \begin{tabular}{l|c|ccc|ccc}
    \toprule
    Methods & Time (s) $\downarrow$ 
    & \multicolumn{3}{c|}{HoloAssist \cite{HoloAssist2023}} 
    & \multicolumn{3}{c}{HOT3D \cite{banerjee2024hot3d}} \\
    & 
    & GA-MPJPE $\downarrow$ & RA-MPJPE $\downarrow$ & PA-MPJPE $\downarrow$
    & GA-MPJPE $\downarrow$ & RA-MPJPE $\downarrow$ & PA-MPJPE $\downarrow$ \\
    \midrule
    HaMeR \cite{pavlakos2024reconstructing} 
    & 72 
    & 59.5 & 57.9 & 32.9 
    & 94.4 & 74.1 & 48.1 \\
    
    Dyn-HaMR \cite{yu2025dynhamr} 
    & 280 
    & 49.6 & 31.5 & 23.0 
    & 107.7 & \textbf{46.9} & 32.0 \\
    
    \textbf{Ours} 
    & \textbf{0.7} 
    & \textbf{23.5} & \textbf{29.7} & \textbf{10.5} 
    & \textbf{38.2} & \underline{48.1} & \textbf{13.6} \\
    \bottomrule
    \end{tabular}
    }

    \rule{0pt}{0.5em}

    \resizebox{\textwidth}{!}{
    \begin{tabular}{l|c|ccc|ccc}
    \toprule
    Methods & Time (s) $\downarrow$ 
    & \multicolumn{3}{c|}{ARCTIC \cite{fan2023arctic}} 
    & \multicolumn{3}{c}{TACO \cite{liu2024taco}} \\
    & 
    & GA-MPJPE $\downarrow$ & RA-MPJPE $\downarrow$ & PA-MPJPE $\downarrow$
    & GA-MPJPE $\downarrow$ & RA-MPJPE $\downarrow$ & PA-MPJPE $\downarrow$ \\
    \midrule
    HaMeR \cite{pavlakos2024reconstructing} 
    & 72  
    & 95.9 & 39.2 & 28.6 
    & 52.0 & 39.3 & 29.1 \\
    
   Dyn-HaMR \cite{yu2025dynhamr} 
    & 280 
    & 76.9 & 33.3 & 22.4 
    & 48.0 & 32.5 & 23.5 \\
    
    \textbf{Ours} 
    & \textbf{0.7} 
    & \textbf{35.1}& \textbf{33.0}  & \textbf{13.7} 
    & \textbf{20.3} & \textbf{25.2} & \textbf{9.4} \\
    \bottomrule
    \end{tabular}
    }

\end{table*}

\noindent\textbf{Baselines.}
We compare \method against two recent state-of-the-art (SOTA) methods for egocentric hand motion reconstruction: HaMeR~\cite{pavlakos2024reconstructing} and Dyn-HaMR~\cite{yu2025dynhamr}. 
We use the official implementations of both methods and follow their recommended evaluation settings whenever applicable.

\noindent\textbf{Evaluation Protocol.} We evaluate on four public egocentric hand benchmarks: HoloAssist~\cite{HoloAssist2023}, HOT3D~\cite{banerjee2024hot3d}, ARCTIC~\cite{fan2023arctic}, and TACO~\cite{liu2024taco}. Each validation sequence is split into non-overlapping 2-s clips (60 frames at 30\,fps). Ground-truth joints are taken directly from each dataset's 3D annotations. We discard only sequence boundaries and frames without valid ground truth: the first and last 10 frames of each ARCTIC sequence, short trailing segments, unannotated hands in HOT3D, and hands not tracked by the HoloAssist headset. All remaining frames are kept, including those in which a hand is occluded or out of view. In contrast, existing benchmarks typically evaluate only visible hands, \eg, frames with at least two ground-truth joints projected inside the image~\cite{zhang2025hawor}, hands whose root lies inside the image (ARCTIC), or hands above a visibility threshold (HOT3D). Yet occluded and out-of-view hands are common in egocentric video, and applications such as AR and robot learning require continuous hand trajectories; excluding them removes the most challenging cases and evaluates an easier task.

All joints (21 per hand) are predicted in the camera coordinate frame of the clip. Errors are computed separately for the left and right hand of each clip over all 60 frames, and then averaged over the two hands. \texttt{RA-MPJPE} subtracts the wrist in every frame and measures articulation, \texttt{GA-MPJPE} applies one similarity alignment per hand per clip and measures the trajectory, and \texttt{PA-MPJPE} applies a per-frame similarity alignment and measures local pose.

To assess computational efficiency, we record the average inference time per clip. Since the Dyn-HaMR baseline relies on computationally intensive test-time optimization, evaluating it across the massive HoloAssist dataset is prohibitively expensive. Therefore, to ensure a fair and tractable comparison, we randomly sample 3,000 clips from the HoloAssist~\cite{HoloAssist2023} validation set and report both accuracy and runtime statistics on this representative subset.

\noindent\textbf{Results.}
Table~\ref{tab:egocentric_reconstruction} shows \method outperforms both frame-based (HaMeR) and dynamic (Dyn-HaMR) baselines across four diverse datasets. Notably, \method achieves SOTA local pose accuracy (\texttt{PA-MPJPE}) and global alignment (\texttt{GA-MPJPE}) while being 100$\times$ faster than HaMeR and over 400$\times$ faster than Dyn-HaMR.

\begin{figure}[t]
    \centering
    \includegraphics[width=1.0\linewidth]{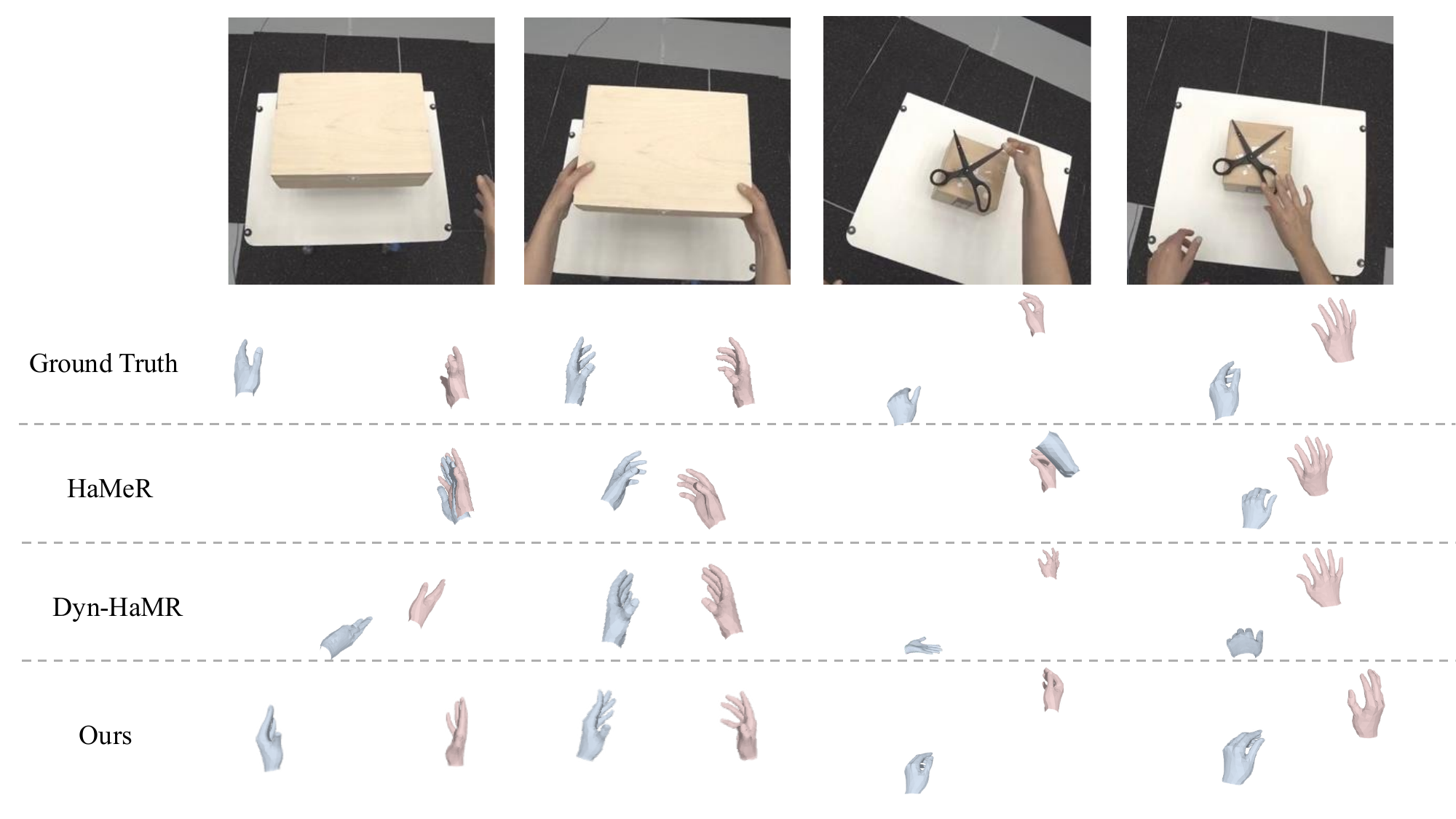}
    \caption{\textbf{Qualitative Results of Egocentric Hand Reconstruction.} \method learns robust spatiotemporal motion priors to generate plausible hand motions, even under severe occlusion or when the hands are out of view. We visualize sampled, non-consecutive frames across two distinct sequences to demonstrate temporal stability.}
    \label{fig:hand}
\end{figure}

Furthermore, qualitative visualizations shown in Fig.~\ref{fig:hand} highlight \method's robustness to severe occlusions common in egocentric vision. While baselines typically collapse or predict broken kinematics under these conditions, \method leverages strong spatiotemporal priors learned from the joint distribution. This enables \method to produce biomechanically plausible, temporally consistent hand movements across temporal gaps (e.g., between frames 4 and 10 in the left sequence, and frames 2 and 6 in the right), even when the hands are entirely unobservable from the egocentric view.

\begin{table*}[t]
\centering
\scriptsize
\setlength{\tabcolsep}{4pt}
\captionof{table}{\textbf{Egocentric camera tracking, gaze estimation and depth estimation on ADT.} We additionally report the average runtime per clip for each method over the camera tracking task, where EgoM2P and our method achieve real-time performance.}
\label{tab:ego_merged}
\begin{tabular}{l|c|ccc|c|cc}
\toprule
& & \multicolumn{3}{c|}{Camera} & Gaze & \multicolumn{2}{c}{Depth} \\
Methods & Time (s) $\downarrow$ & ATE $\downarrow$ & RTE $\downarrow$ & RRE $\downarrow$ & MSE $\downarrow$ & Abs Rel $\downarrow$ & $\delta_{1.25}$ $\uparrow$ \\
\midrule
EgoM2P~\cite{li2025egom2p}   & \textbf{0.7} & 0.030          & \textbf{0.009} & 1.290          & 0.0311          & 0.458          & 30.3          \\
EgoMono4D~\cite{egomono4d}   & 14.2         & 0.051          & 0.015          & 1.307          & -               & \textbf{0.150} & \textbf{83.9} \\
VIPE~\cite{huang2025vipe}    & 25.8         & \textbf{0.005} & \textbf{0.009} & 1.307          & -               & -              & -             \\
Ours                         & \textbf{0.7} & \underline{0.015}          & \textbf{0.009} & \textbf{1.279} & \textbf{0.0211} & \underline{0.265}          & \underline{56.5}          \\
\bottomrule
\end{tabular}
\end{table*}
\subsection{Egocentric Camera Tracking}

\noindent\textbf{Baselines.}
We compare \method with egocentric-focused 4D reconstruction methods (EgoM2P~\cite{li2025egom2p} and EgoMono4D~\cite{egomono4d}), as well as the SOTA general-scene 4D reconstruction method VIPE~\cite{huang2025vipe}.
EgoMono4D relies on off-the-shelf optical flow predictions and reformulates camera tracking as depth alignment across confidence-aware correspondences. Similarly, VIPE performs dense bundle adjustment over off-the-shelf predictions of dense optical flow, sparse tracks, and metric depth. In contrast, \method does not rely on explicit geometric constraints and predicts camera poses directly from egocentric RGB input end-to-end.

\noindent\textbf{Evaluation Protocol.}
We evaluate on all sequences from the unseen ADT dataset. For camera tracking, we report standard error metrics, including Absolute Translation Error (ATE), Relative Translation Error (RTE), and Relative Rotation Error (RRE). All predicted camera trajectories are aligned to the ground truth using a Sim(3) transformation estimated over the entire clip. 

\noindent\textbf{Results.}
See Tab.~\ref{tab:ego_merged} for quantitative results. Our method outperforms all baselines except VIPE on the ATE metric, while achieving the lowest RTE and RRE among all methods. The performance gap with VIPE on ATE can be attributed to its use of dense bundle adjustment over the entire input clip, whereas our method predicts camera poses in a single feedforward pass without geometric post-processing. Qualitative results are in the Supplementary Material Sec. E.

\subsection{Egocentric Gaze Estimation}

We compare \method with EgoM2P~\cite{li2025egom2p} on the ADT dataset to predict 2D gaze locations from egocentric videos. Both the predicted gaze locations and ground truth are normalized to $[0,1]$ with respect to the input resolution. We report the Mean Squared Error (MSE) as the evaluation metric.
As reported in Tab.~\ref{tab:ego_merged}, our method achieves significantly lower error compared to EgoM2P on unseen data. Qualitative results are provided in the Supplementary Material Sec. E.

\subsection{Egocentric Video Depth Estimation}

\noindent\textbf{Baselines.}
We compare \method against egocentric-focused 4D reconstruction methods EgoM2P~\cite{li2025egom2p} and EgoMono4D~\cite{egomono4d} for depth estimation. EgoMono4D leverages UniDepth~\cite{piccinelli2024unidepth}, a specialized monocular depth estimation model, by finetuning on egocentric data. In contrast, our method is trained from scratch and does not rely on expert priors.

\noindent\textbf{Evaluation Protocol.}
We use all sequences with depth annotations on ADT~\cite{Pan_2023_ICCV} and filter out clips where the timestamp difference between RGB and depth frames exceeds 10 ms to avoid misalignment. For all methods, predictions are aligned with the ground truth with a clip-wise scale and translation factor. We then report relative metrics, including Absolute Relative Error (Abs Rel) and the percentage of predicted depths within a $1.25$ factor of the ground truth ($\delta_{1.25}$).

\noindent\textbf{Results.}
The quantitative results are reported in Tab.~\ref{tab:ego_merged}. \method significantly outperforms EgoM2P on the unseen dataset but falls behind EgoMono4D on both metrics. We attribute this gap to two main factors: (1) EgoMono4D benefits from strong depth-specific priors inherited from UniDepth pretrained weights, whereas ours is trained without depth expert initialization; and (2) the discrete tokenization from the Cosmos tokenizer~\cite{agarwal2025cosmos} introduces quantization errors that can affect fine-grained depth prediction.
While there is a gap, our method is substantially more efficient, achieving over \textbf{20}$\times$ faster inference than EgoMono4D.
\subsection{Ablation Studies}
\label{sec:ablation}

\newcommand{\modelFull}{\textbf{ReViV}}
\newcommand{\modelBodyData}{ReViV-BodyData}
\newcommand{\modelBodyTask}{ReViV-BodySpec.}
\newcommand{\modelHandData}{ReViV-HandData}
\newcommand{\modelHandTask}{ReViV-HandSpec.}
\newcommand{\modelAlpha}{ReViV-1to1Mask}
\newcommand{\modelUniform}{ReViV-Uniform}
\newcommand{\modelTiny}{ReViV-Tiny}
\newcommand{\modelFiveK}{ReViV-w/oNy}
\newcommand{\modelThreeK}{ReViV-w/oNy\&Holo}

\begin{table}[t]
    \centering
    \scriptsize
    \renewcommand{\arraystretch}{0.95}
    \setlength{\tabcolsep}{4pt}
    \caption{\textbf{Ablation on Ego Body Motion, Camera Tracking $\&$ Depth.}}
    \label{tab:ablation_main}
    \begin{tabular}{@{}lcccc|cc|cc@{}}
        \toprule
        & \multicolumn{4}{c|}{Body Motion} & \multicolumn{2}{c|}{Camera} & \multicolumn{2}{c}{Depth} \\
        \cmidrule(lr){2-5} \cmidrule(lr){6-7} \cmidrule(lr){8-9}
        Methods & GA $\downarrow$ & PA $\downarrow$ & Sim $\uparrow$ & FID $\downarrow$ & ATE $\downarrow$ & RRE $\downarrow$ & Rel $\downarrow$ & $\delta_{1.25}$ $\uparrow$ \\
        \midrule
        \modelFull & \textbf{111.5} & \textbf{88.6} & \textbf{0.751} & \textbf{0.442} & \textbf{0.015} & 1.279 & \textbf{0.265} & \textbf{56.5} \\
        \modelBodyData & 134.0 & 109.3 & 0.645 & 0.478 & -- & -- & -- & -- \\
        \modelBodyTask & 200.3 & 139.9 & 0.545 & 0.623 & -- & -- & -- & -- \\
        \modelFiveK & -- & -- & -- & -- & 0.031 & 1.293 & 0.432 & 32.0 \\
        \modelThreeK & -- & -- & -- & -- & 0.032 & 1.335 & 0.452 & 30.1 \\
        \modelAlpha & 158.3 & 120.6 & 0.615 & 0.548 & 0.036 & \textbf{1.267} & 0.401 & 36.8 \\
        \modelUniform & 172.7 & 119.8 & 0.598 & 0.665 & 0.053 & 1.529 & 0.494 & 26.7 \\
        \modelTiny & 143.1 & 116.7 & 0.634 & 0.486 & 0.026 & 1.269 & 0.337 & 44.7 \\
        \bottomrule
    \end{tabular}
\end{table}

We conduct systematic ablation studies to validate each core design choice of \method, as summarized in Tab.~\ref{tab:ablation_main}. Complementary hand reconstruction ablations are provided in the Supplementary Material Sec. F.

\noindent\textbf{Task-Specific vs. Unified Model.}
Training a task-specific body specialist (\modelBodyTask) with only RGB input suffers a massive accuracy drop across all metrics compared to \method. Isolated single-modality training fails to exploit the rich cross-modal supervision signals captured by our joint formulation.

\noindent\textbf{Model Capacity.}
A lightweight variant (\modelTiny, $\sim$127M parameters) exhibits noticeably degraded performance across body motion, camera tracking, and depth estimation. This confirms that the more parameters is critical for approximating the complex joint distribution underlying our large-scale multimodal dataset, and that reconstruction quality scales with model capacity.

\noindent\textbf{Masking Strategy.}
We trained an independent 1-to-1 modality masking by setting Dirichlet $\alpha$ to $0.001$ (\modelAlpha), missing deep cross-modal info. Another ablation using uniform sampling on all modalities (\modelUniform), which causes sparse body tokens with unbalanced dense video tokens, collapsing accuracy. Our sampling strategy mitigates both by producing all possible mask permutations for an ensemble of arbitrary conditional distributions.

\noindent\textbf{Data Scaling.}
To disentangle data volume, we incrementally removed training datasets (Nymeria, and Nymeria + HoloAssist), yielding \modelFiveK~and \modelThreeK~. Experiment results confirm performance scales with data volume. Models (\modelBodyData) only trained on datasets with body annotations degrade accuracy compared to our full model. This demonstrates that even without explicit annotations, joint training on massive datasets significantly enhances reconstruction by enriching generalized cross-modal priors.

\section{Conclusion}

In this work, we presented \method, the first unified generative framework to bridge the longstanding gap between egocentric scene reconstruction and unobserved human kinematics. By casting the inherently ambiguous task of ego-body estimation as a joint probability learning problem, we demonstrated that a single monocular RGB video is sufficient to reconstruct both the dynamic 4D environment and the full-body, hand, and gaze behaviors of the camera wearer. At the core of our approach is the Masked Generative Egocentric Transformer (MGET), which successfully aligns these heterogeneous modalities into a shared, temporally coherent latent space. Through large-scale pretraining, \method establishes a new foundational baseline for holistic egocentric perception, achieving state-of-the-art performance without relying on specialized hardware or pre-computed SLAM. Ultimately, by unifying the viewer and the view within a single feed-forward architecture, this work provides a stepping stone for next-generation embodied AI systems capable of natural, human-centric reasoning.

\paragraph{Limitations and Future Directions.}
We leverage VQ-VAEs to resolve the heterogeneity of egocentric multimodalities. By mapping these diverse continuous signals into a unified discrete token space, we enable a single Transformer to jointly model both the viewer and the view, benefiting from the stable scaling laws that drive modern foundation models. However, this unified discrete representation introduces an inherent trade-off for dense regression tasks. Specifically, the quantization process inevitably discards some high-frequency spatial details, which slightly limits the model's capacity for depth estimation compared to task-specific continuous regression works. Exploring the integration of continuous generative priors, such as conditional diffusion models, into the decoding stage could be a future direction. This would allow our framework to retain its robust multimodal reasoning capabilities while effectively recovering the fine-grained geometric details necessary for high-fidelity 4D scene reconstruction.

\subsubsection*{Acknowledgements.}
\label{sec:contrib}
Xiaozhong Lyu and Gen Li contributed equally to this work. Gen Li formulated the core idea, developed the main codebase, implemented the hand tokenizer, trained the main model (without the body component), and led the paper writing. Xiaozhong Lyu drove the system scale-up and finalization by expanding the multimodal dataset to 7B tokens, developing the body tokenizer, retraining the gaze and camera tokenizers, training the final full-scale main model, conducting experiments, and creating visualizations.

Xiaozhong Lyu was supported by an ETH Zurich Research Grant. Gen Li was supported by a Microsoft Spatial AI Zurich Lab PhD scholarship.
This work was also supported as part of the Swiss AI Initiative by a grant from the Swiss National Supercomputing Centre (CSCS) under project IDs a143, a144 on Alps.
We thank Zinuo You and Malte Prinzler for proofreading.

\bibliographystyle{splncs04}
\bibliography{main}

\clearpage
\appendix

\title{
ReViV: Reconstructing the Viewer and the View in 4D from Monocular Egocentric Video
--Supplementary Material--
}

\titlerunning{ReViV: Reconstructing the Viewer and the View in 4D from Ego Video}

\author{Xiaozhong Lyu\inst{1}\thanks{Equal contribution; order interchangeable on CVs. See Acknowledgements.}\orcidlink{0009-0008-6857-3311} \and
Gen Li\inst{1}\textsuperscript{$\star$}\orcidlink{0009-0005-9782-7649} \and
Zhiyin Qian\inst{1}\orcidlink{0009-0008-3975-0016} \and \newline
Xucong Zhang\inst{1,2}\orcidlink{0000-0002-8368-3542} \and
Marc Pollefeys\inst{1,3}\orcidlink{0000-0003-2448-2318} \and
Siyu Tang\inst{1}\orcidlink{0000-0002-1015-4770}
}

\authorrunning{X.~Lyu et al.}

\institute{ETH Zurich, Switzerland \and
Delft University of Technology, Netherlands \and
Microsoft, Switzerland}

\maketitle

\section{Tokenizer Details}

We unify continuous spatiotemporal multimodalities by tokenizing them into discrete representational spaces. For RGB and depth videos, we follow the pipeline established by EgoM2P \cite{li2025egom2p} and employ the Cosmos tokenizer (with a codebook size of 64,000) to extract visual tokens. For low-dimensional modalities including camera trajectory, gaze dynamics, and body/hand motions, we design and train modality-specific vector-quantized autoencoders (VQ-VAEs). The detailed hyperparameters for these tokenizers are summarized in Tab. \ref{tab:tokenizer_settings}.

We adopt the VQ-VAE architecture from EgoM2P \cite{li2025egom2p} for camera trajectory ($\mathbf{y}_{\text{cam}}$) and gaze dynamics ($\mathbf{y}_{\text{gaze}}$). However, to ensure robust generalization across diverse spatial environments and human behaviors, we significantly scale up the models by re-training both tokenizers from scratch on our newly-curated large-scale database in Sec. 3 of the main paper. For both modalities, the input sequences are passed through a temporal 1D convolution with a kernel size of 2, which downsamples the sequence temporally by a factor of 2 while mapping the channel dimension to 768. The features are then processed by a 12-layer Transformer encoder (ViT-B). Following EgoM2P, we account for invalid values in gaze signals (e.g., hardware occlusion or tracking loss) by applying a binary mask during training and computing a Masked Mean Squared Error (MSE) reconstruction loss, allowing the network's deep smoothness to naturally impute missing values. We set the codebook sizes to 512 for gaze and 256 for camera trajectories.

To effectively capture the complex spatiotemporal articulation of human whole-body poses, we introduce dedicated tokenizers specifically designed for body and hand kinematics. Recognizing that human motion exhibits strong correlations across both the temporal domain and spatial skeletal topology, instead of relying purely on temporal convolutions, we apply 2D convolutions over the continuous pose sequences. This operation jointly compresses the data with a temporal downsampling factor of 2 and a spatial (joint) compression factor of 3, projecting the complex skeletal structure into a unified 768-dimensional latent space.
These representations are subsequently encoded using a 12-block Transformer (ViT-B) with self-attention. For reconstruction, we calculate the standard MSE loss. To capture the higher degrees of freedom inherent in human skeletal movements, we expand the representational capacity of these tokenizers by setting the codebook sizes to 2048 for body motion and 1024 for hand motions. To leverage the bilateral symmetry of human hands, we train separate tokenizers for the left and right hands using an augmented dataset. Specifically, we effectively double the training volume for each tokenizer by incorporating mirrored motions from the opposite hand—supplementing the left-hand data with mirrored right-hand trajectories, and vice versa. Each tokenizer is trained for approximately 72 hours on 4 NVIDIA H100 GPUs. As shown in the Tab. \ref{tab:tokenizer_settings}, we choose hyperparameters by training multiple versions of tokenizers and select the best model with the lowest validation loss.

\begin{table}[t]
    \centering
    \scriptsize
    \begin{tabular}{l |c c c c}
        \toprule
        Configuration & Body & Hands & Gaze & Camera\\
        \midrule
        Codebook size & 2048 & 1024 & 512 & 256 \\
        Temporal compression & 2 & 2 & 2 & 2 \\
        Space compression & 3 & 3 & - & - \\
        Code latent dimension & 32 & 32 & 32 & 32 \\
        EMA dead code threshold & 2 & 2 & 2 & 2 \\
        Codebook EMA & 0.99& 0.99& 0.99 & 0.99 \\
        $l_2$-normalized codes \cite{yu2022vectorquantized} & \checkmark & \checkmark & \checkmark & \checkmark \\
        Codebook weight & 1.0 & 1.0 & 1.0 & 1.0 \\
        Commitment weight $\beta$ & 1.0 & 1.0 & 1.0 & 1.0 \\
        \midrule
        Loss function & MSE& MSE& Masked MSE & MSE \\
        Base learning rate &5e-6 &5e-6& 5e-5 & 2.5e-5\\
        Learning rate & 2e-5 & 2e-5& 1e-4 & 5e-5\\
        Batch size & 64 & 64 & 128 & 128 \\
        \midrule
        Encoder architecture &  \multicolumn{4}{c}{ViT-B}\\
        Decoder architecture & \multicolumn{4}{c}{ViT-B} \\
        
        Optimizer & \multicolumn{4}{c}{AdamW \cite{loshchilov2018decoupled}} \\
        Opt. momentum & \multicolumn{4}{c}{$\beta_1, \beta_2 = 0.9, 0.95$} \\
        Weight decay & \multicolumn{4}{c}{0.05} \\
        Max gradient norm & \multicolumn{4}{c}{1}  \\
        Learning rate sched. & \multicolumn{4}{c}{Cosine decay} \\
        Training epochs &   \multicolumn{4}{c}{200}   \\
        Warmup epochs & \multicolumn{4}{c}{5} \\
        Data type & \multicolumn{4}{c}{float32} \\
        \bottomrule
    \end{tabular}
    \caption{\textbf{Tokenizer training settings.}}
    \label{tab:tokenizer_settings}
\end{table}

\section{Masked Generative Egocentric Transformer Details}

We provide the training algorithm for Masked Generative Egocentric Transformer (MGET) shown in Alg. \ref{alg:pretraining}. Please note that in Alg. \ref{alg:pretraining}, ``rgb'' denotes discrete video tokens, whereas ``ViT'' represents continuous video embeddings.

\begin{algorithm}[ht]
\caption{Multimodal Masked Pretraining for MGET}
\begin{algorithmic}[1]
    \State \textbf{Input:} Datasets $\{D_1, \dots, D_N\}$, training modalities $\mathcal{M} = \{\text{rgb}, \text{ViT}, \text{depth}, \text{cam}, \text{gaze}, \text{hand}, \text{body}\}$, max sequence length $L_{\text{max}} = 2048$
    \State \textbf{Output:} Optimized MGET parameters $\theta$
    
    \State \textit{/* Dataset and Token Proportion Sampling */}
    \State $i \sim \text{Categorical}(p_1, \dots, p_N)$ \Comment{Sample dataset proportional to total size}
    \State $\alpha \sim \text{Uniform}(\{(0.01, \dots), (0.1, \dots), (1, \dots), (10, \dots)\})$ \Comment{Sample Dirichlet concentration parameters}
    \State $\boldsymbol{\pi} \sim \text{Dirichlet}(\alpha')$ \Comment{Sample token allocation ratios across modalities}
    
    \State $\mathbf{Z}_{\mathcal{V}} \gets \emptyset, \mathbf{Z}_{\mathcal{M}} \gets \emptyset$ \Comment{Init sets for visible input and masked target tokens}
    
    \State \textit{/* Token Allocation and Masking */}
    \For{$j \in \mathcal{M}$}
        \State $T_j \gets L_{\text{max}} \times \boldsymbol{\pi}_j$ \Comment{Max visible tokens for modality $j$}
        \State $\mathbf{z}_{j, \mathcal{V}} \gets \text{Sample}(D_{i, j}, T_j)$ \Comment{Sample visible input tokens}
        \State $\mathbf{z}_{j, \mathcal{M}} \gets \text{Sample}(D_{i, j} \setminus \mathbf{z}_{j, \mathcal{V}}, T_j)$ \Comment{Sample mutually exclusive targets}
        \State $\mathbf{Z}_{\mathcal{V}} \gets \mathbf{Z}_{\mathcal{V}} \cup \mathbf{z}_{j, \mathcal{V}}$
        \State $\mathbf{Z}_{\mathcal{M}} \gets \mathbf{Z}_{\mathcal{M}} \cup \mathbf{z}_{j, \mathcal{M}}$
    \EndFor
    
    \State \textit{/* Modality-Specific Embedding and Context Injection */}
    \State $\mathbf{Z}_{\text{emb}} \gets \text{Embed}(\mathbf{Z}_{\mathcal{V}} \cup \mathbf{Z}_{\mathcal{M}}) + \text{PosTypeEncodings}()$
    
    \State \textit{/* MGET Forward Pass */}
    \State $\mathbf{C}_{\text{enc}} \gets \text{Encoder}_{\theta}(\mathbf{Z}_{\text{emb}}[\mathbf{Z}_{\mathcal{V}}])$ \Comment{Unrestricted global self-attention on input}
    \State $\hat{\mathbf{P}} \gets \text{Decoder}_{\theta}(\mathbf{Z}_{\text{emb}}[\mathbf{Z}_{\mathcal{M}}], \mathbf{C}_{\text{enc}})$ \Comment{Intra-modal self-attn \& inter-modal cross-attn}
    
    \State \textit{/* Optimization */}
    \State $\mathcal{L}_{\text{mask}} \gets \text{CrossEntropyLoss}(\hat{\mathbf{P}}, \mathbf{Z}_{\mathcal{M}})$
    \State $\theta \gets \text{AdamW}(\theta, \nabla_{\theta} \mathcal{L}_{\text{mask}})$ \Comment{Update model weights}
    
    \State \textbf{Return:} $\theta$
\end{algorithmic}
\label{alg:pretraining}
\end{algorithm}

We train the MGET model on the unified discrete token sequences spanning all modalities using the AdamW optimizer with a base learning rate of $1\times10^{-4}$ and a global batch size of 1024.
In total, the training process consumes approximately 500B tokens. We adopt a warm-up schedule over the first 10 epochs to stabilize optimization.
Full model training requires approximately 24 hours on 256 NVIDIA H100 GPUs.

\begin{figure}[ht]
\centering
\setlength{\tabcolsep}{1pt}
\renewcommand{\arraystretch}{0.5}
\begin{tabular}{cccccc}
    & Input & Ours & EgoM2P~\cite{li2025egom2p} & EgoMono4D~\cite{egomono4d} & VIPE~\cite{huang2025vipe} \\[2pt]
    \rotatebox{90}{\parbox{0.14\linewidth}{\centering Frame 05}} &
    \includegraphics[width=0.14\linewidth]{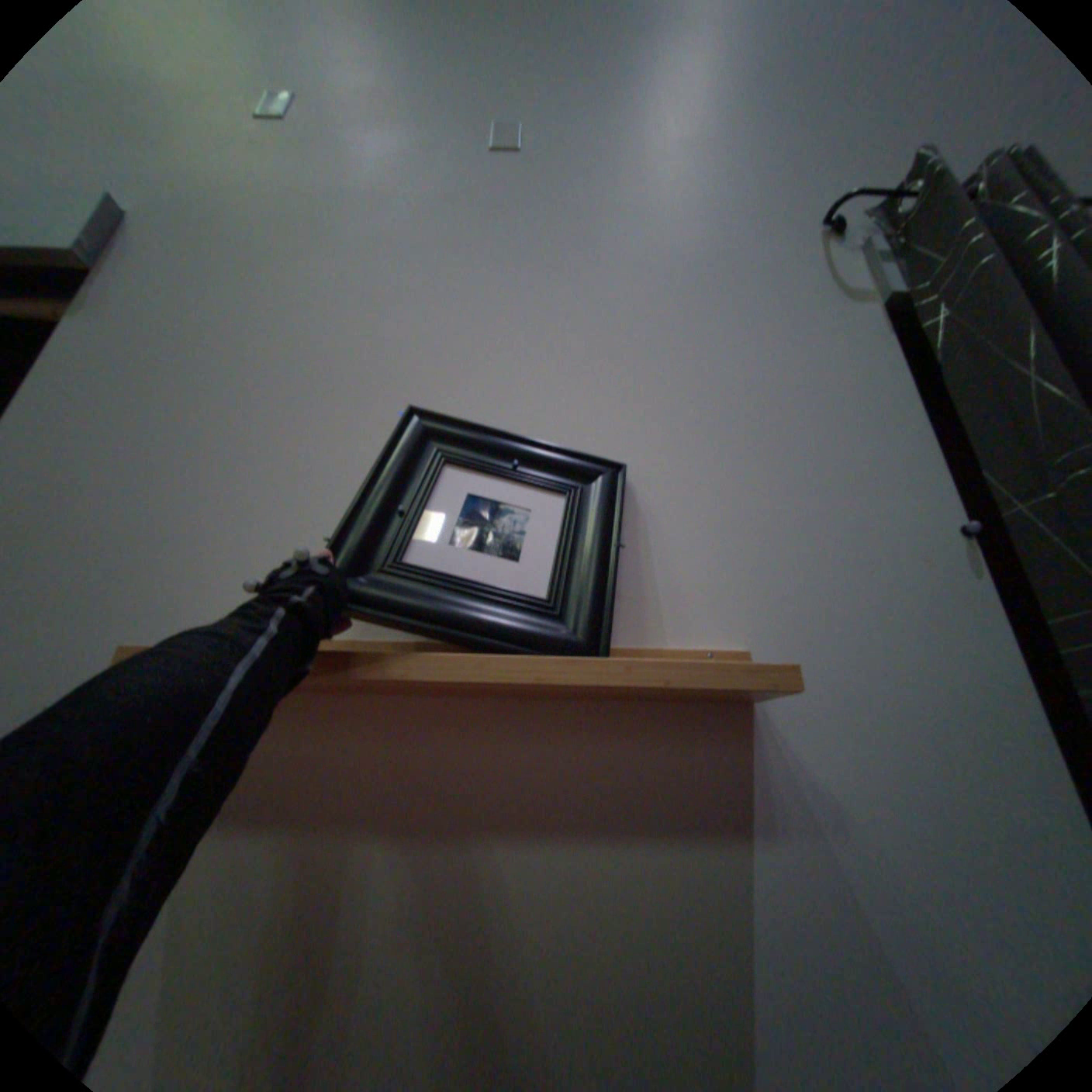} &
    \includegraphics[width=0.14\linewidth]{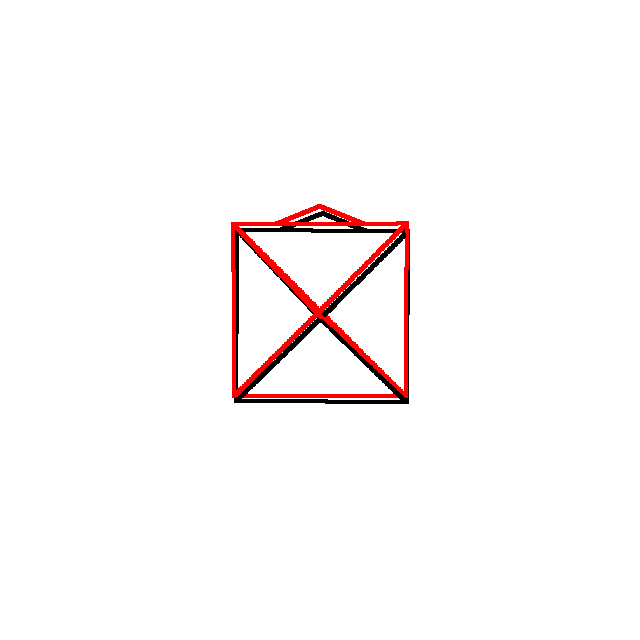} &
    \includegraphics[width=0.14\linewidth]{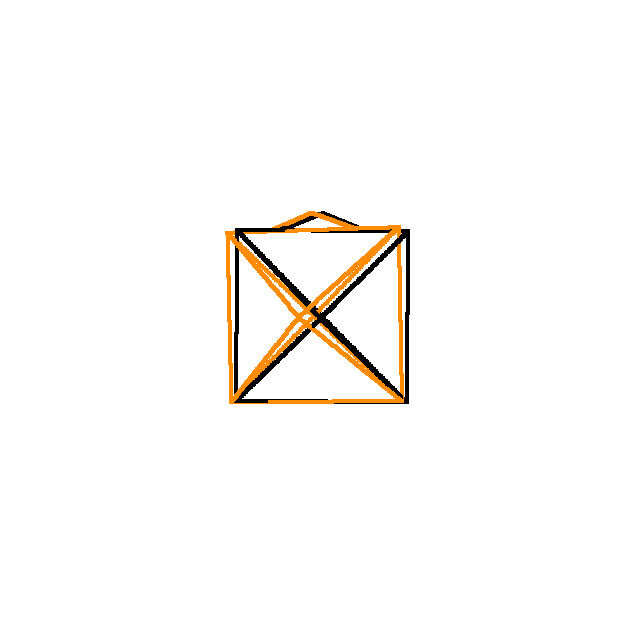} &
    \includegraphics[width=0.14\linewidth]{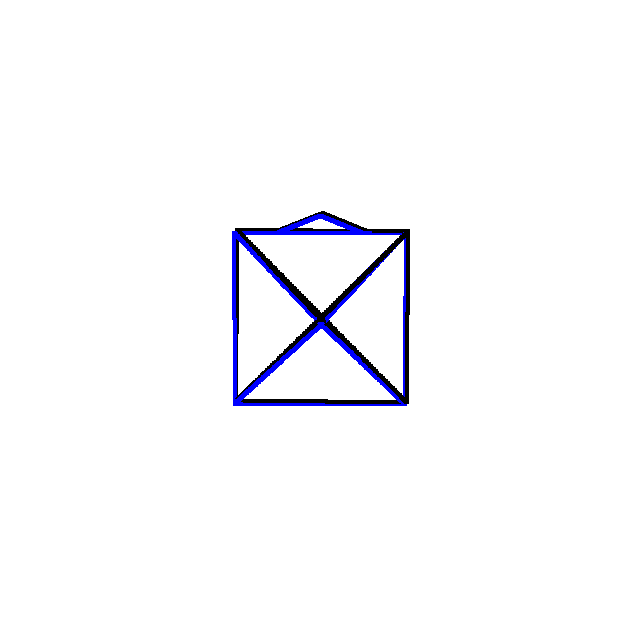} &
    \includegraphics[width=0.14\linewidth]{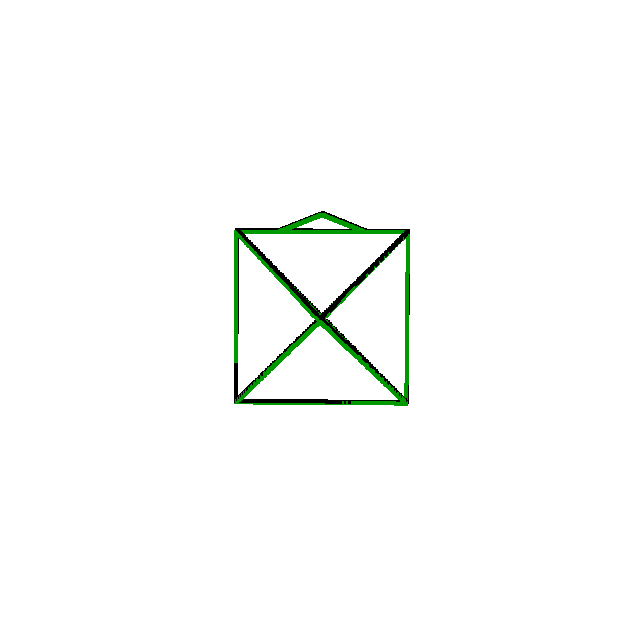} \\
    \rotatebox{90}{\parbox{0.14\linewidth}{\centering Frame 25}} &
    \includegraphics[width=0.14\linewidth]{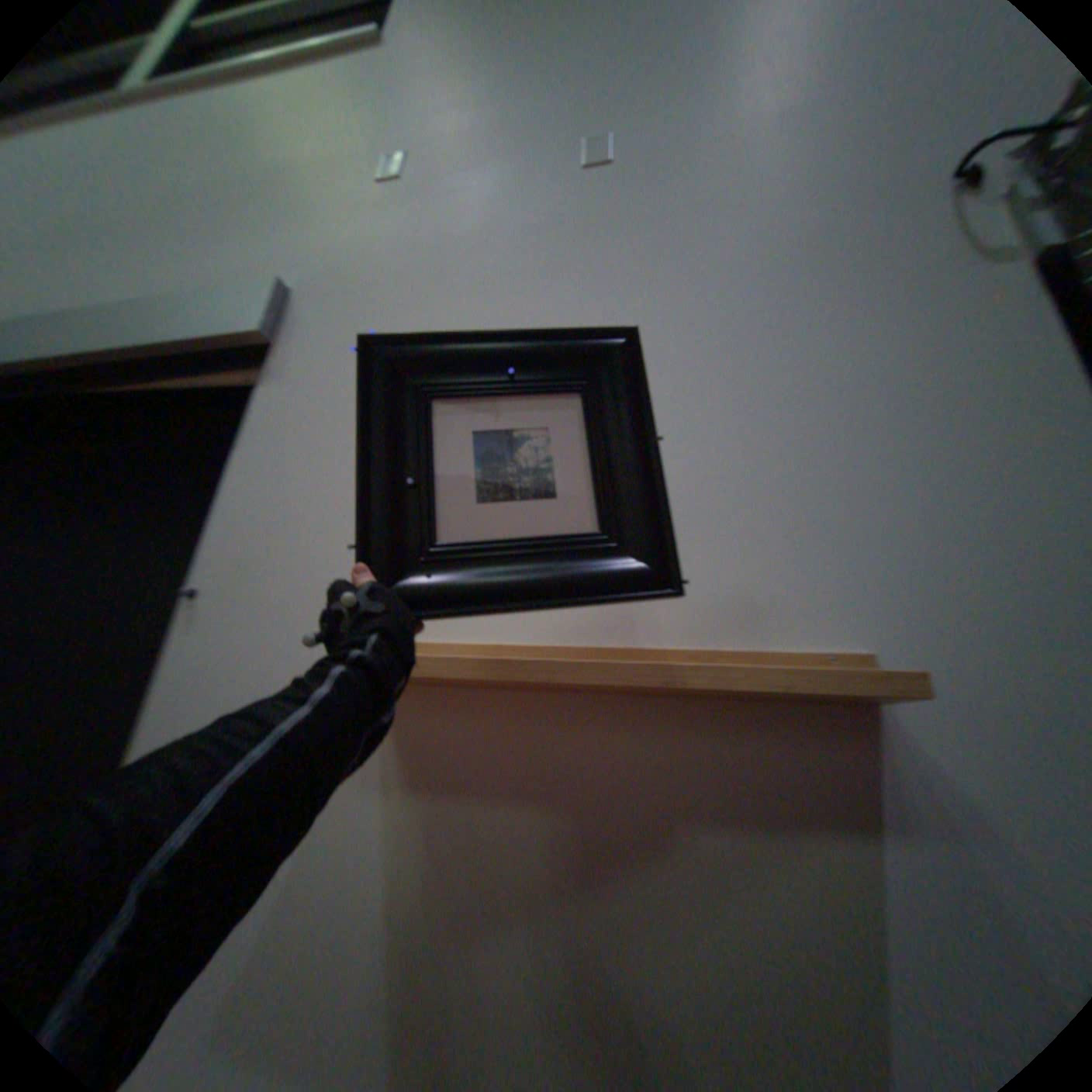} &
    \includegraphics[width=0.14\linewidth]{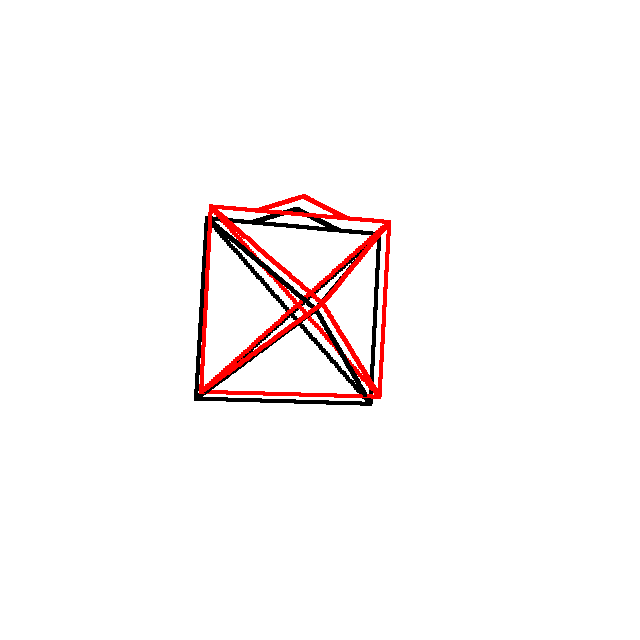} &
    \includegraphics[width=0.14\linewidth]{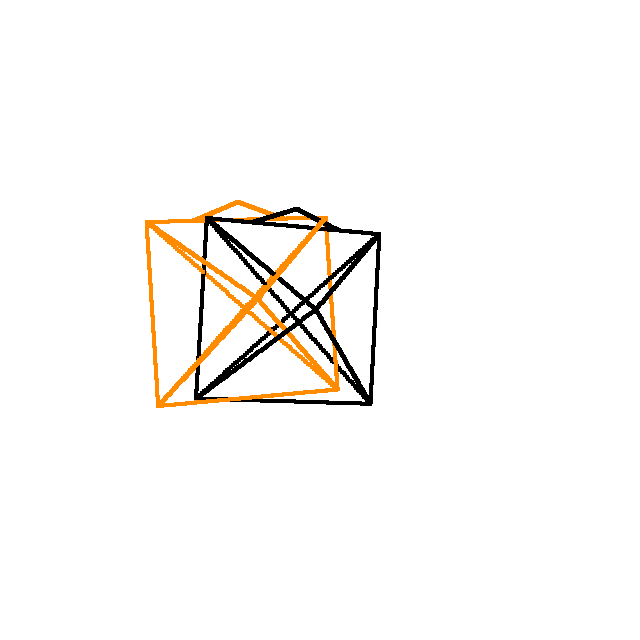} &
    \includegraphics[width=0.14\linewidth]{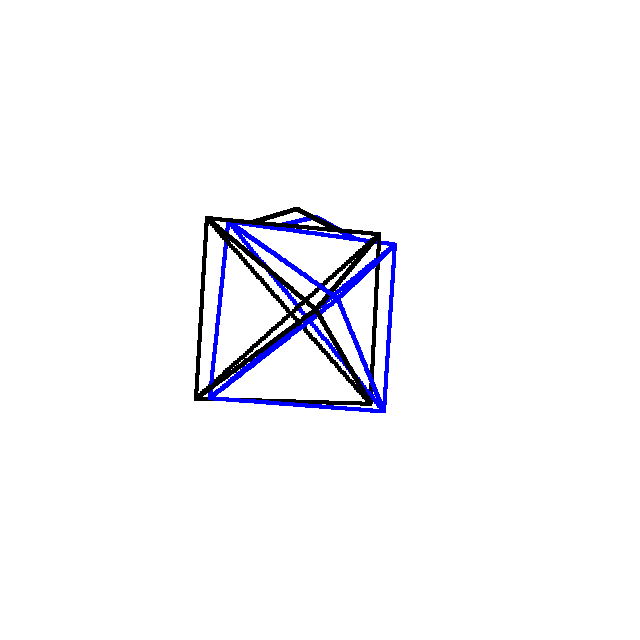} &
    \includegraphics[width=0.14\linewidth]{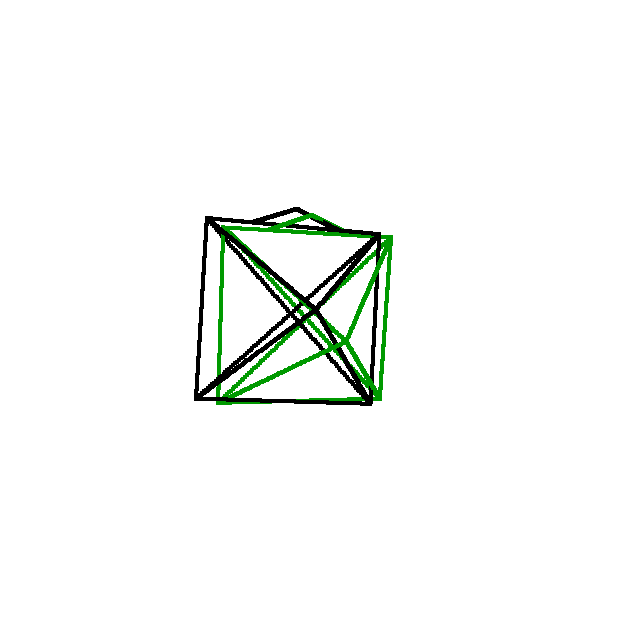} \\
    \rotatebox{90}{\parbox{0.14\linewidth}{\centering Frame 45}} &
    \includegraphics[width=0.14\linewidth]{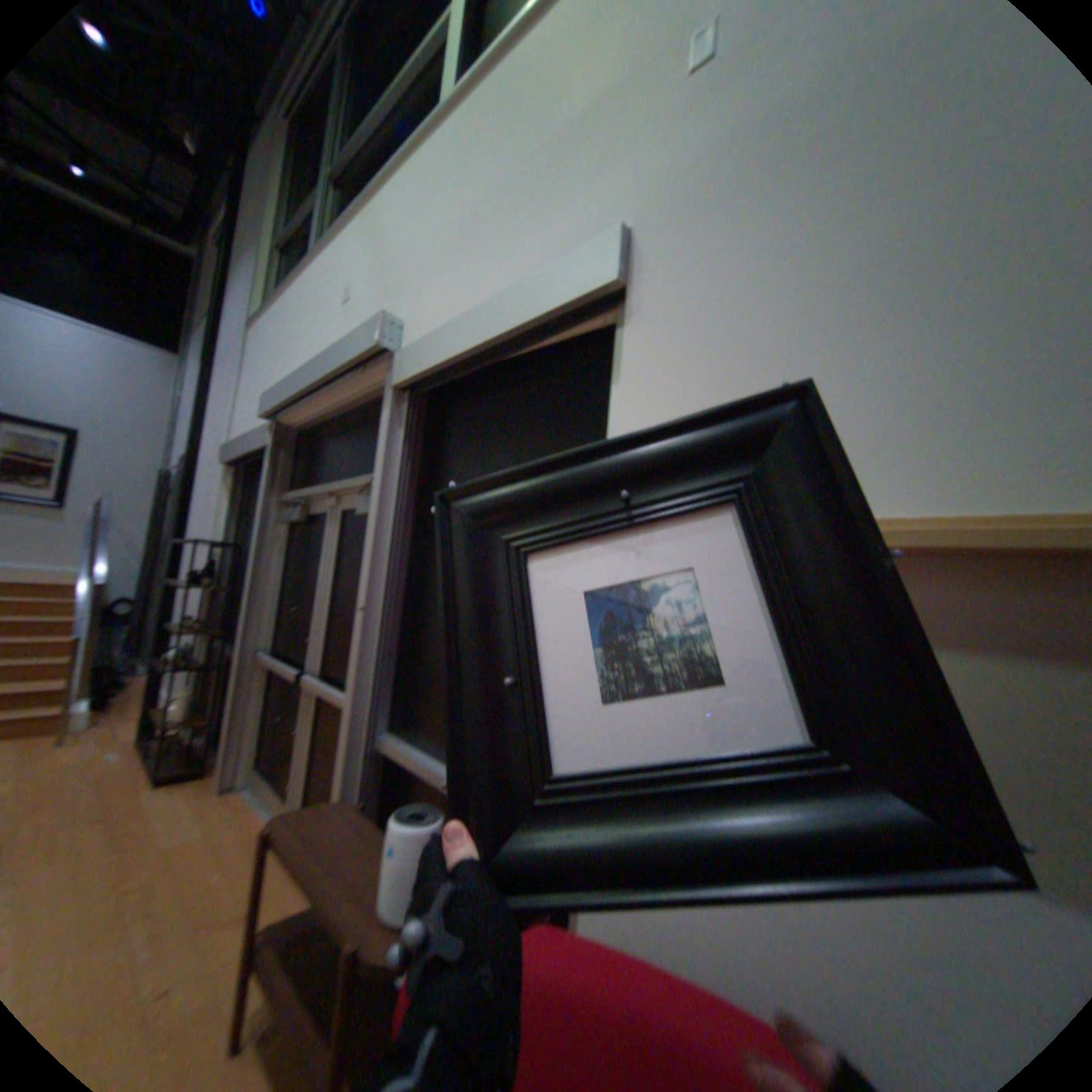} &
    \includegraphics[width=0.14\linewidth]{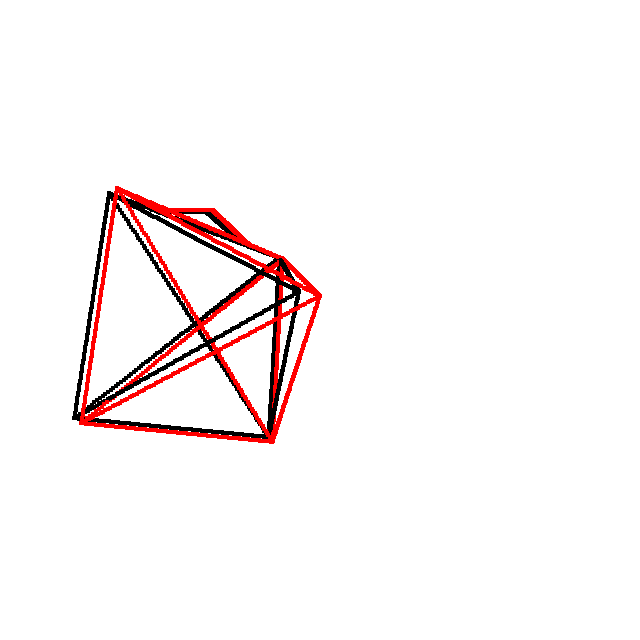} &
    \includegraphics[width=0.14\linewidth]{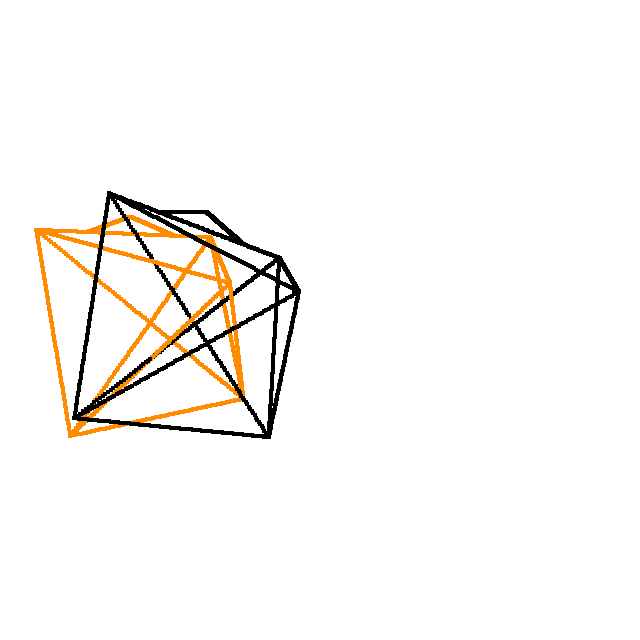} &
    \includegraphics[width=0.14\linewidth]{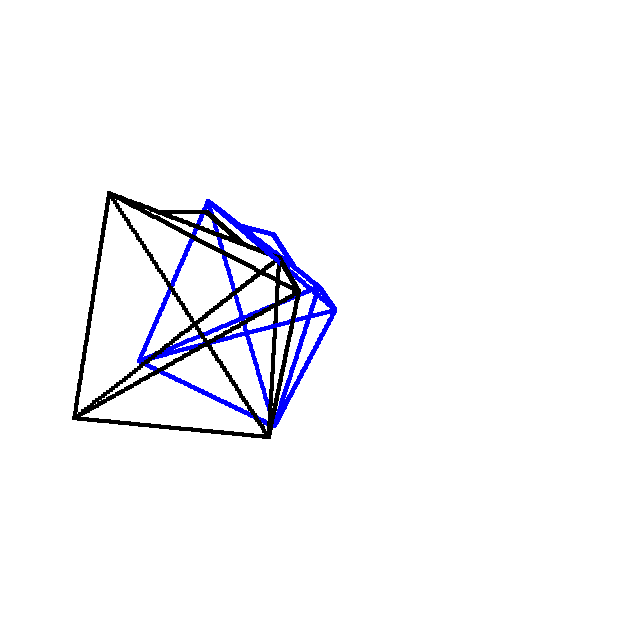} &
    \includegraphics[width=0.14\linewidth]{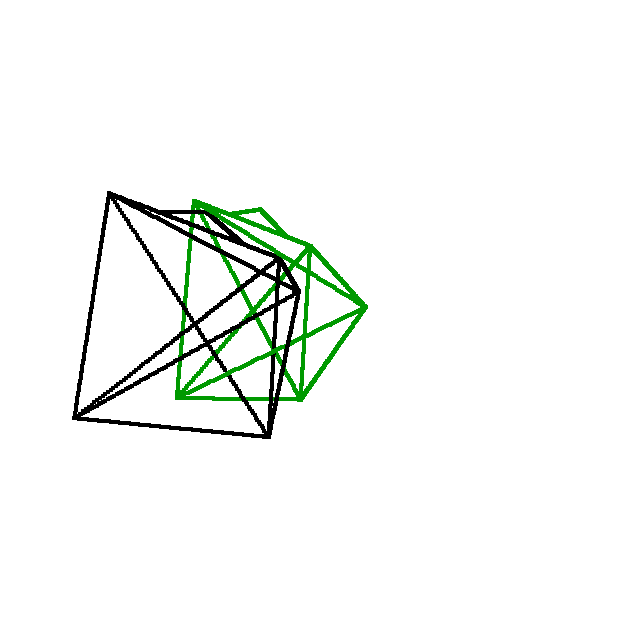} \\
\end{tabular}
\begin{tabular}{cccccc}
    & Input & Ours & EgoM2P~\cite{li2025egom2p} & EgoMono4D~\cite{egomono4d} & VIPE~\cite{huang2025vipe} \\[2pt]
    \rotatebox{90}{\parbox{0.14\linewidth}{\centering Frame 05}} &
    \includegraphics[width=0.14\linewidth]{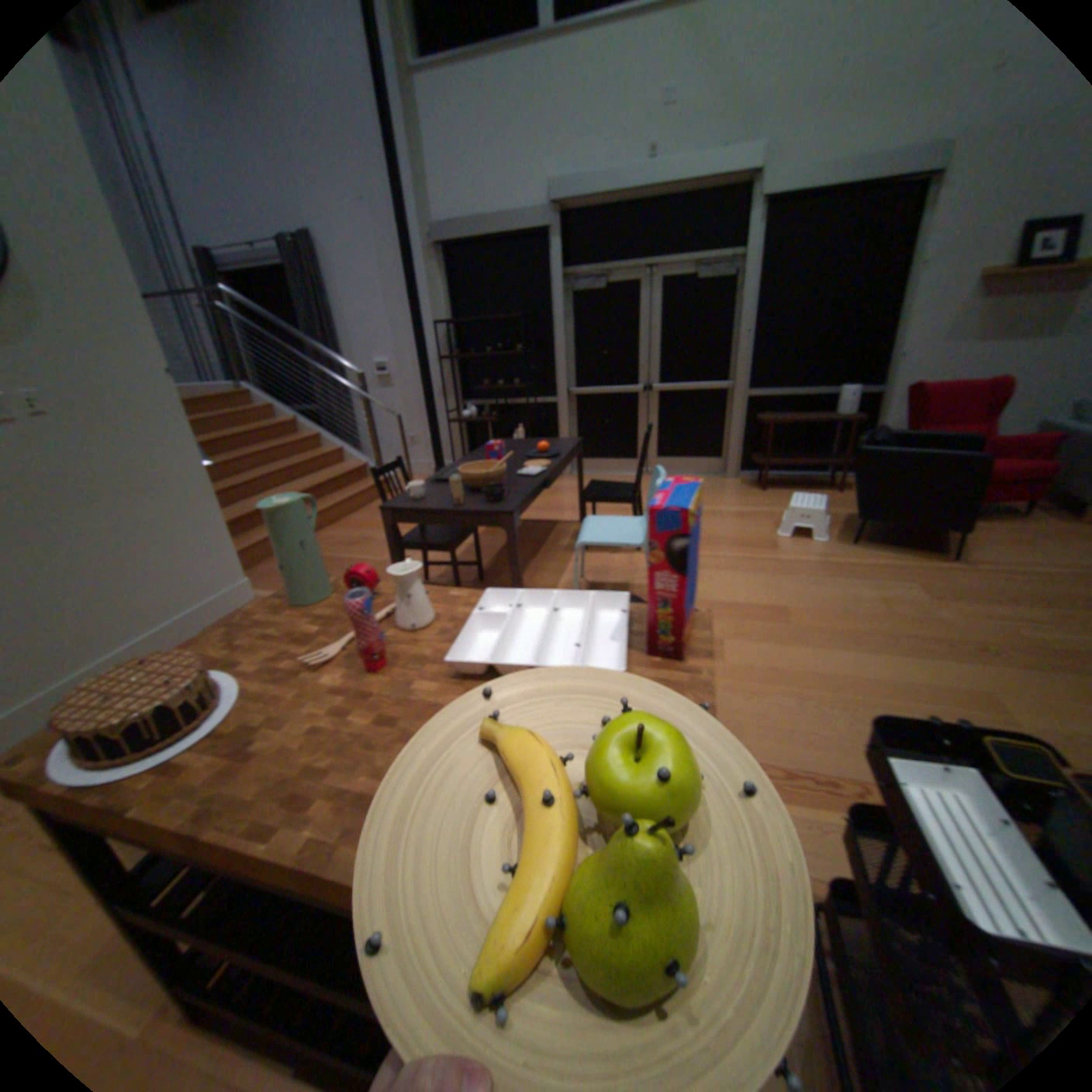} &
    \includegraphics[width=0.14\linewidth]{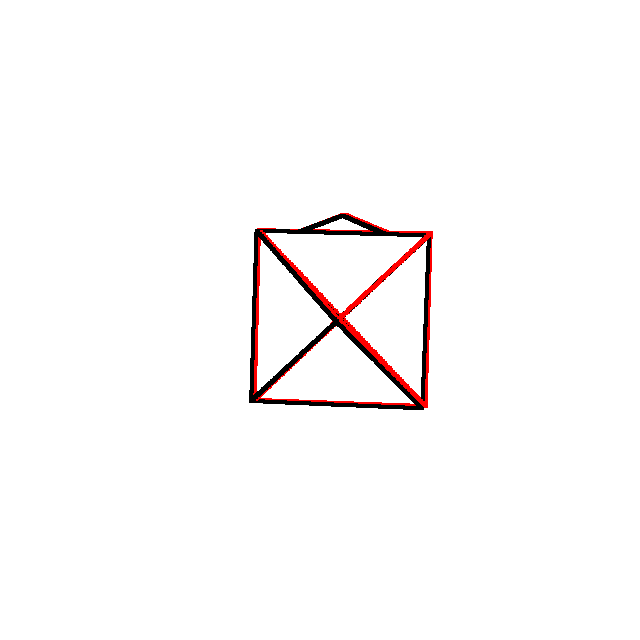} &
    \includegraphics[width=0.14\linewidth]{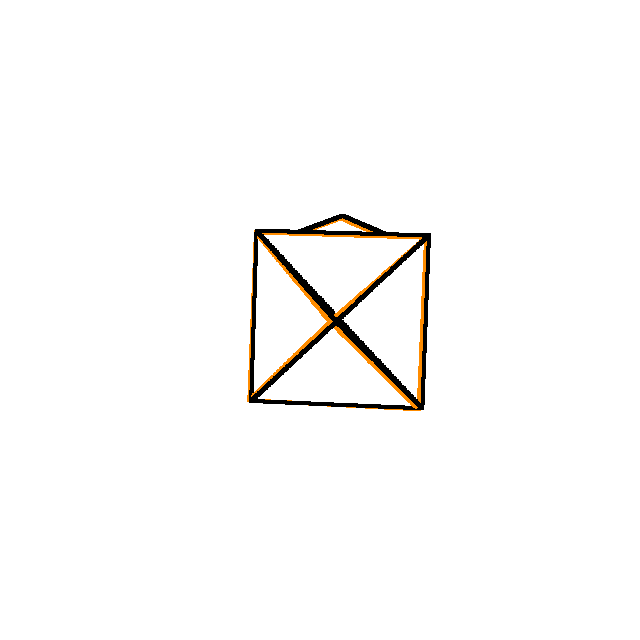} &
    \includegraphics[width=0.14\linewidth]{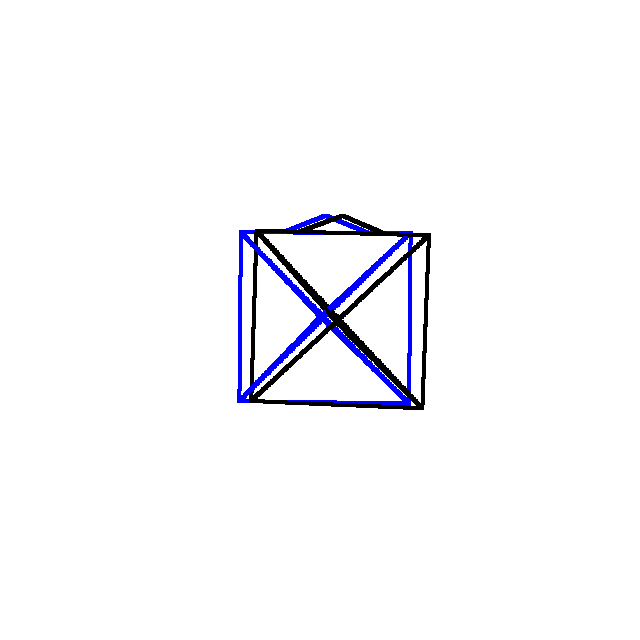} &
    \includegraphics[width=0.14\linewidth]{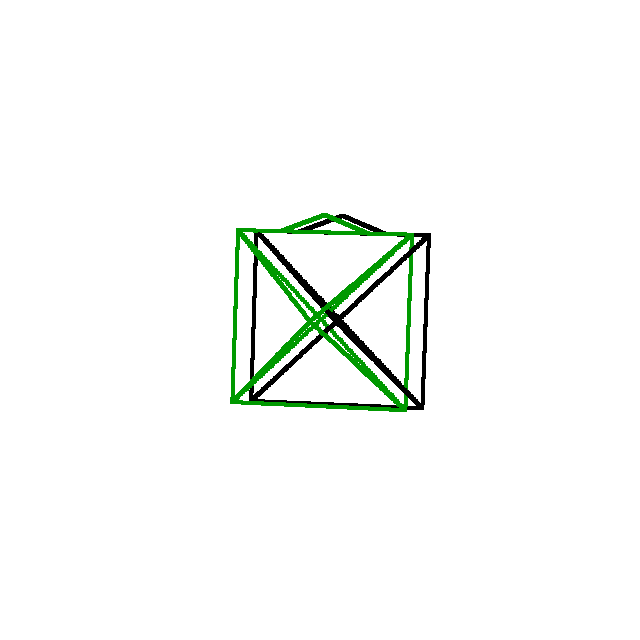} \\
    \rotatebox{90}{\parbox{0.14\linewidth}{\centering Frame 25}} &
    \includegraphics[width=0.14\linewidth]{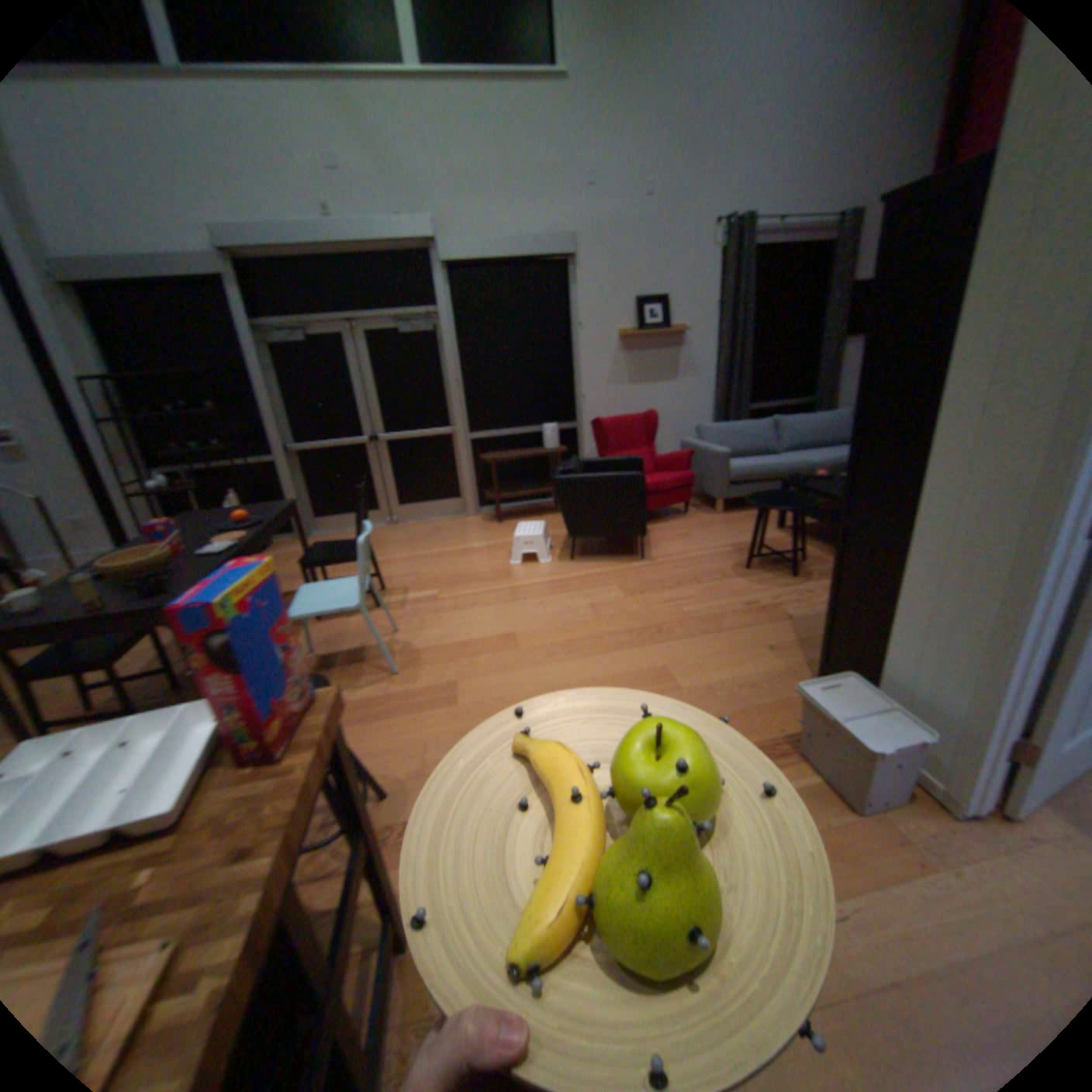} &
    \includegraphics[width=0.14\linewidth]{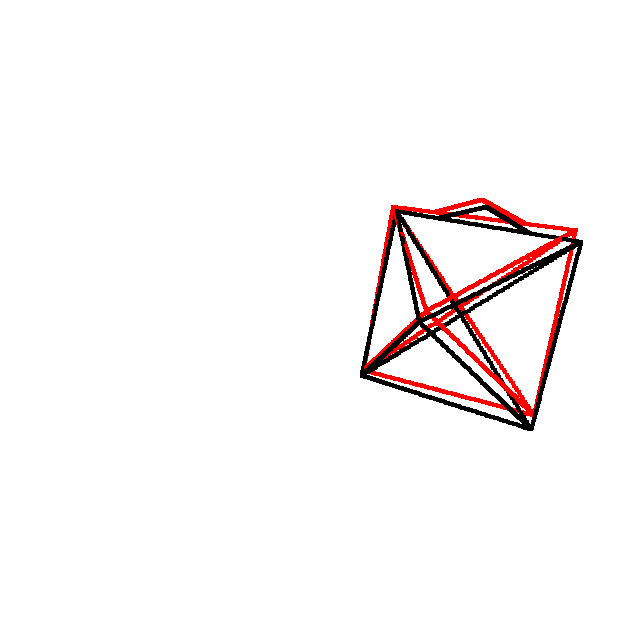} &
    \includegraphics[width=0.14\linewidth]{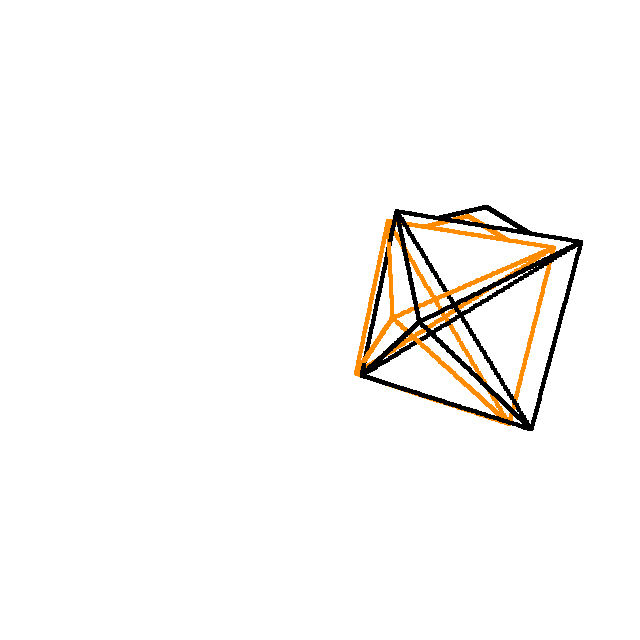} &
    \includegraphics[width=0.14\linewidth]{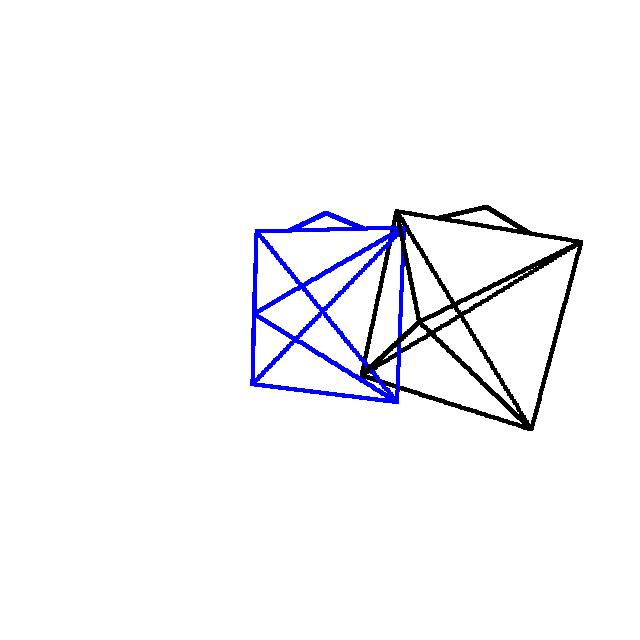} &
    \includegraphics[width=0.14\linewidth]{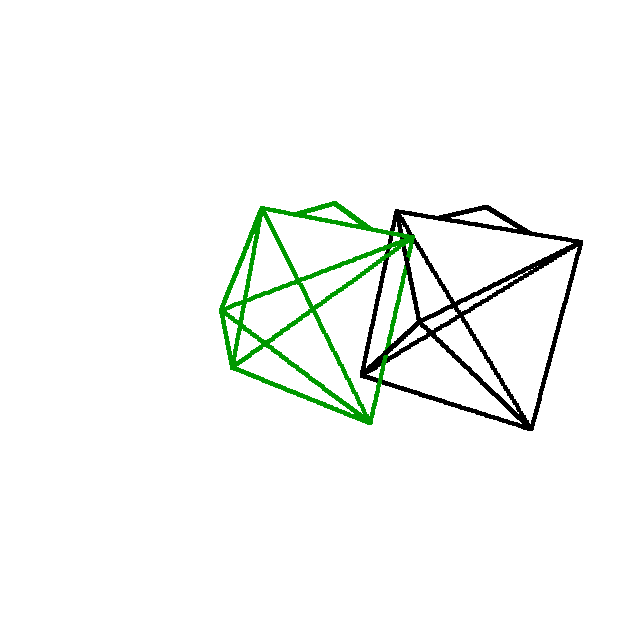} \\
    \rotatebox{90}{\parbox{0.14\linewidth}{\centering Frame 45}} &
    \includegraphics[width=0.14\linewidth]{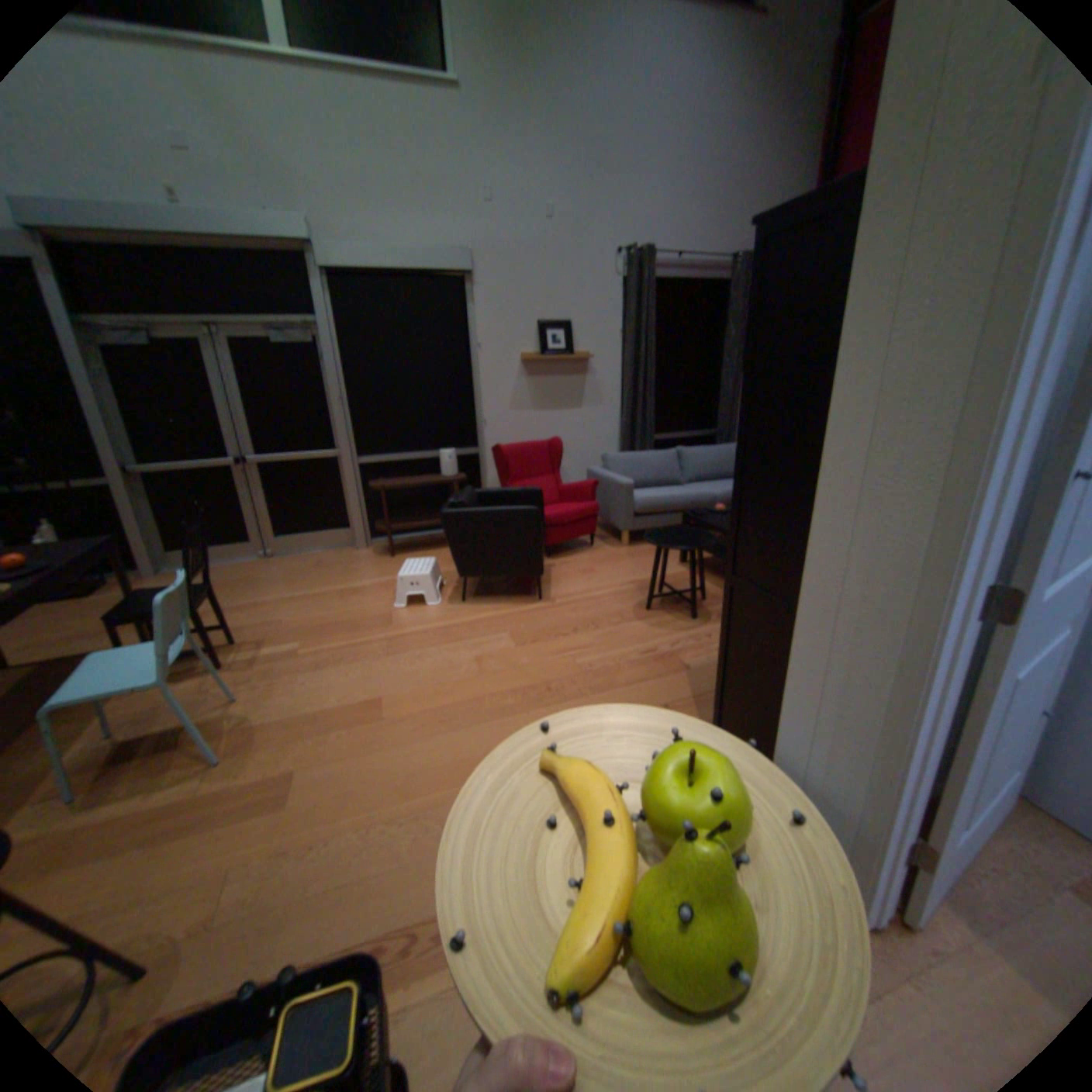} &
    \includegraphics[width=0.14\linewidth]{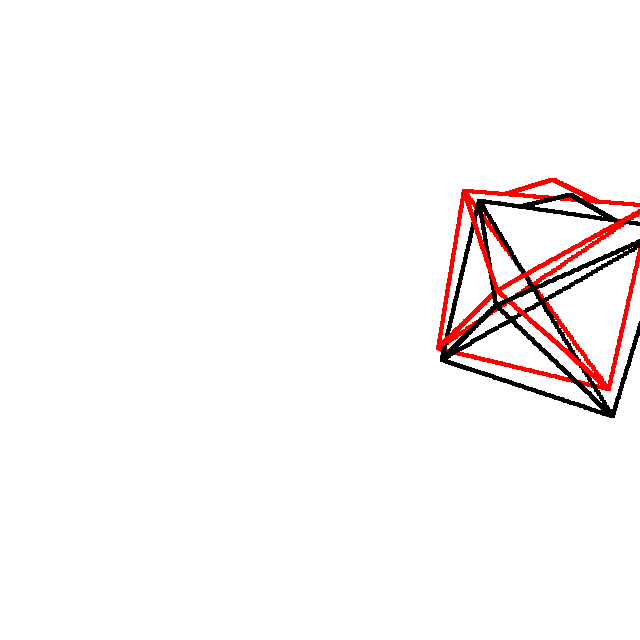} &
    \includegraphics[width=0.14\linewidth]{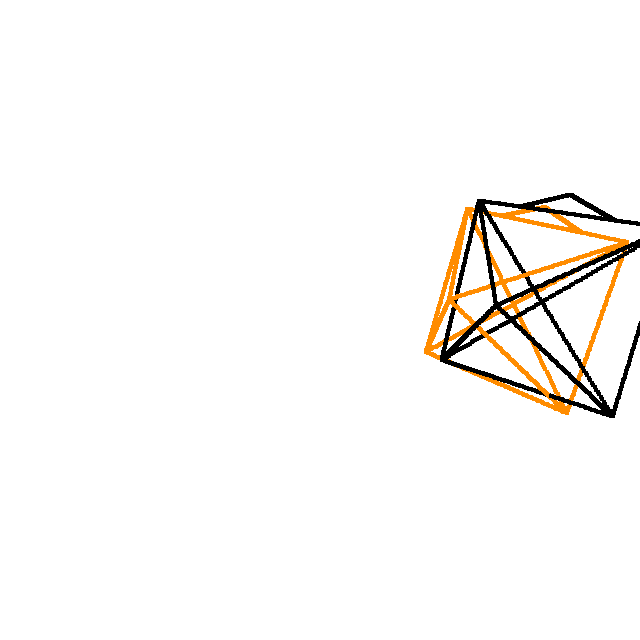} &
    \includegraphics[width=0.14\linewidth]{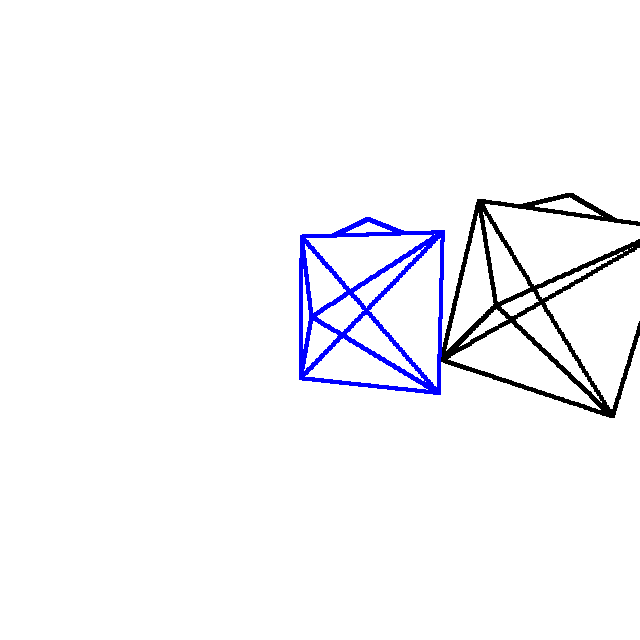} &
    \includegraphics[width=0.14\linewidth]{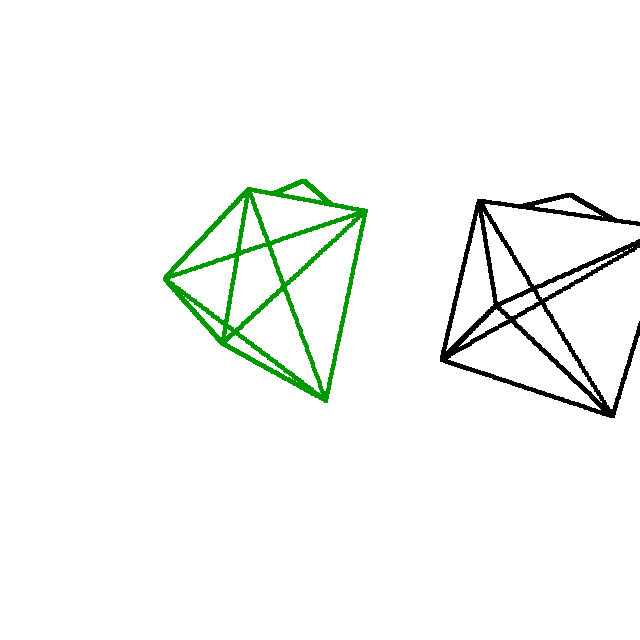} \\
\end{tabular}
\caption{\textbf{Qualitative visualization of egocentric camera tracking results.} Ground truth and predictions are shown with black and colored wireframes, respectively. }
\label{fig:cam_wireframe}
\end{figure}

\section{Inference Details}

Following the order-agnostic, multi-step parallel decoding approach of EgoM2P \cite{li2025egom2p}, we iteratively reconstruct all $n$ masked human and scene state tokens ($\mathbf{Z}_{\mathcal{M}}$) over $s=3$ steps. In each step, we uniformly sample $n/s$ unpredicted target tokens and estimate their distributions using two forward passes: a conditional pass (conditioned on both the visible video context $\mathbf{Z}_{\mathcal{V}}$ and any previously predicted state tokens) and an unconditional pass (which masks the video context but retains the previously predicted states). The final tokens are then drawn from these combined logits using classifier-free guidance ($w=2$) and nucleus sampling (top-$p=0.8$).

\section{Ego-body anchored depth scale optimization}

We detail a straightforward but effective formulation for estimating the metric scale of the predicted relative depth. This approach explicitly aligns the scene depth scale with the reconstructed human scale, providing a physically grounded metric reconstruction in egocentric settings.

Given the predicted ego-body motion moving within the world space, we can roughly approximate the ground plane location. Specifically, we extract the $z$-coordinates of the predicted feet joints across the video sequence and compute the average of the minimum 50\% of these $z$-values to represent the ego-body floor level. 

Simultaneously, we lift the relative depth predictions into unscaled 3D scene point clouds. Assuming the lowest visible points in the scene also lie on the floor, we isolate the points corresponding to the minimum 5\% of the $z$-values in the unscaled scene point cloud. 

To align the scene to the metric ego-body space, we apply a least-squares alignment to solve for a scale factor $s$ (along with a standard translation shift $t$) by minimizing the squared distance between the lowest 5\% of points in the scene point cloud and the estimated ego-body floor level. 

For sequences where the floor is not visible, we additionally consider an off-the-shelf metric depth anchor. We leverage metric depth predictions, denoted as $\mathbf{d}_{\text{vipe}}$, obtained from VIPE~\cite{huang2025vipe} as a geometric anchor. For each video, we estimate a global scale factor $s$ and translation shift $t$ by minimizing the discrepancy between the reconstructed relative depth $\mathbf{y}_{\text{depth}}$ and the metric anchor $\mathbf{d}_{\text{vipe}}$. Together, these lightweight alignment strategies provide complementary ways to map the reconstructed viewer and view into a metrically consistent 4D coordinate system.

\section{Additional Experiment Results}
\begin{figure}[t]
    \begin{minipage}[t]{0.49\columnwidth}
        \begin{subfigure}{0.48\linewidth}
            \includegraphics[width=\linewidth]{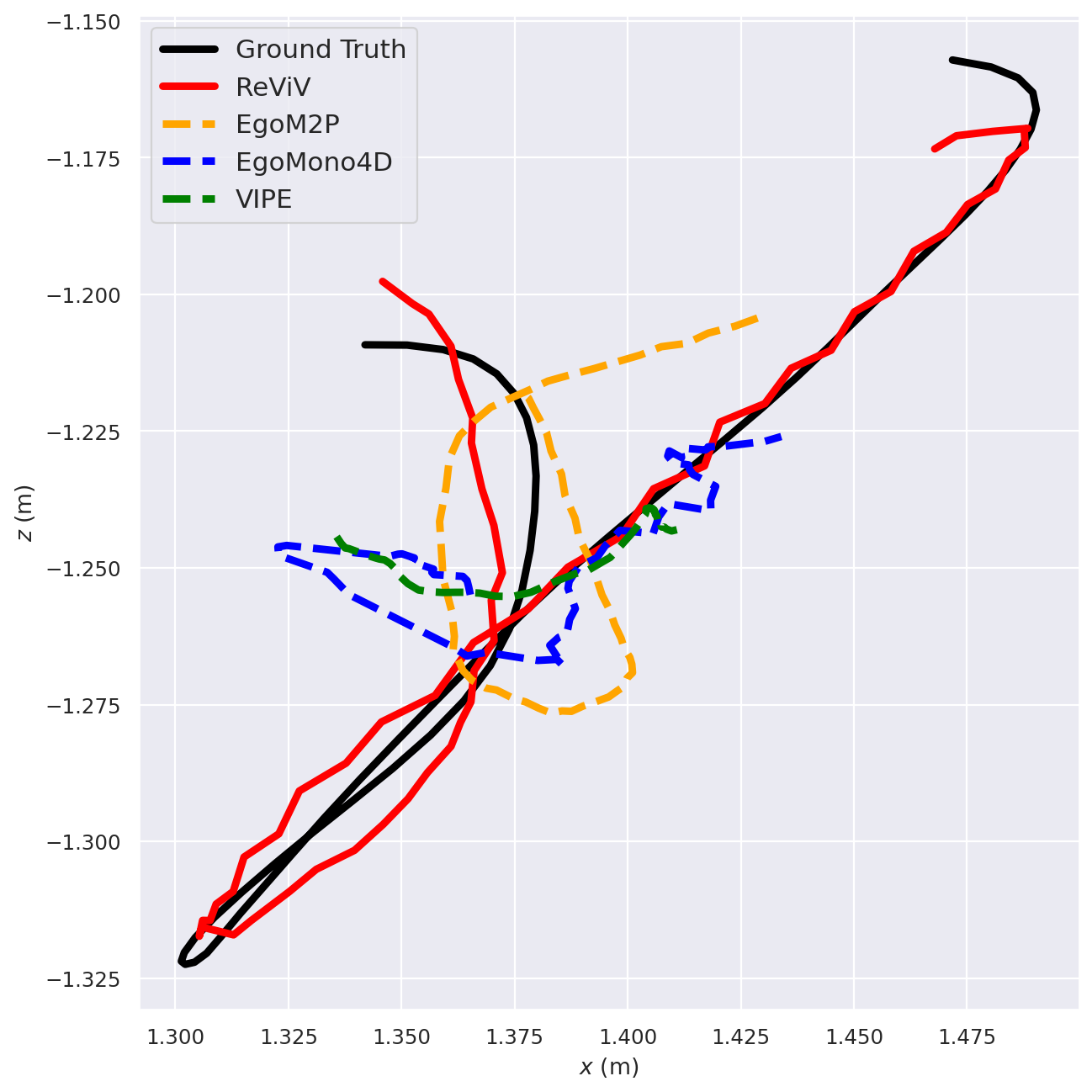}
        \end{subfigure}
        \hfill
        \begin{subfigure}{0.48\linewidth}
            \includegraphics[width=\linewidth]{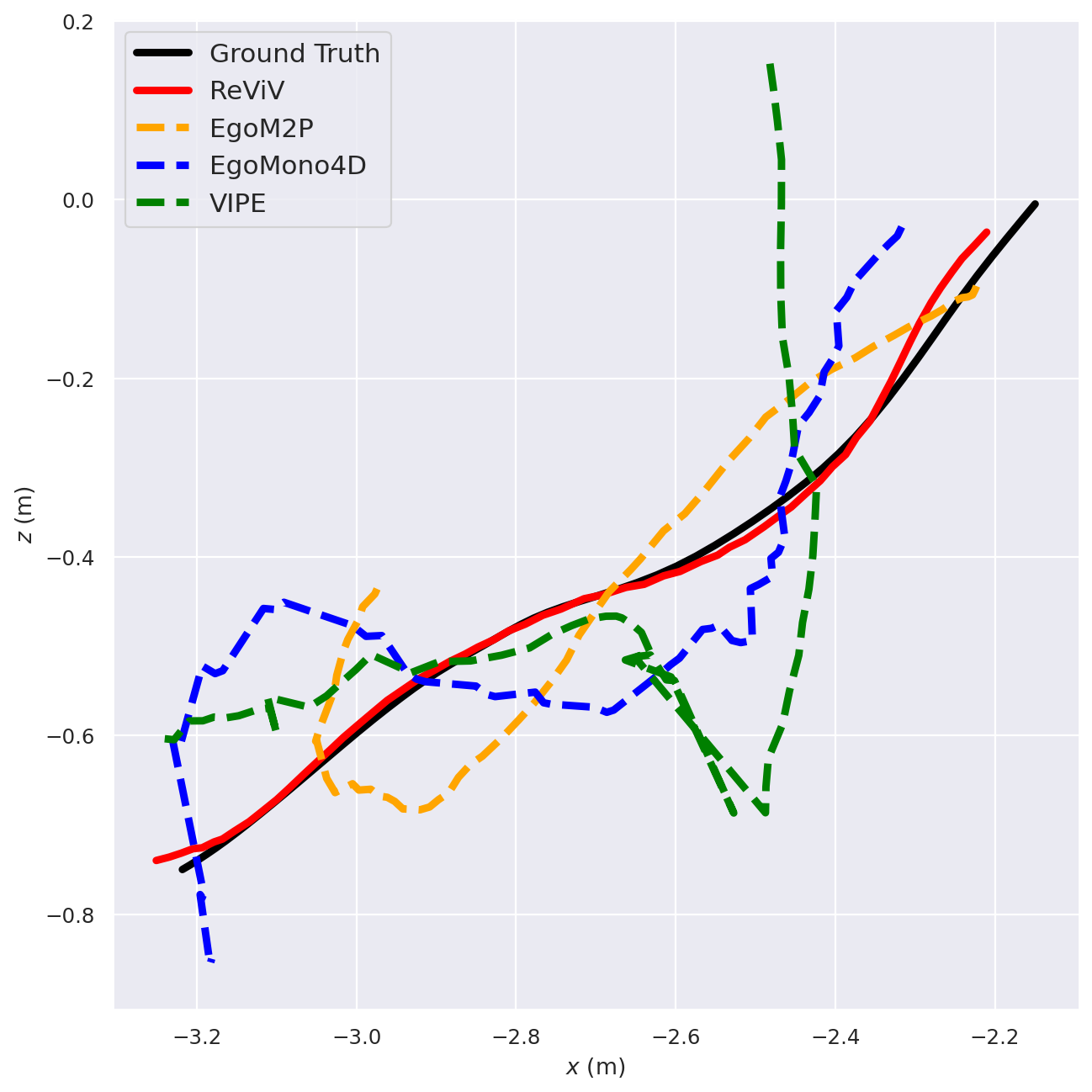}
        \end{subfigure}
        \caption{\textbf{Comparison of camera trajectories projected on 2D plane.}  }
        \label{fig:cam_traj}
    \end{minipage}
    \hfill
    \begin{minipage}[t]{0.48\columnwidth}
        \begin{subfigure}{0.48\linewidth}
            \includegraphics[width=\linewidth]{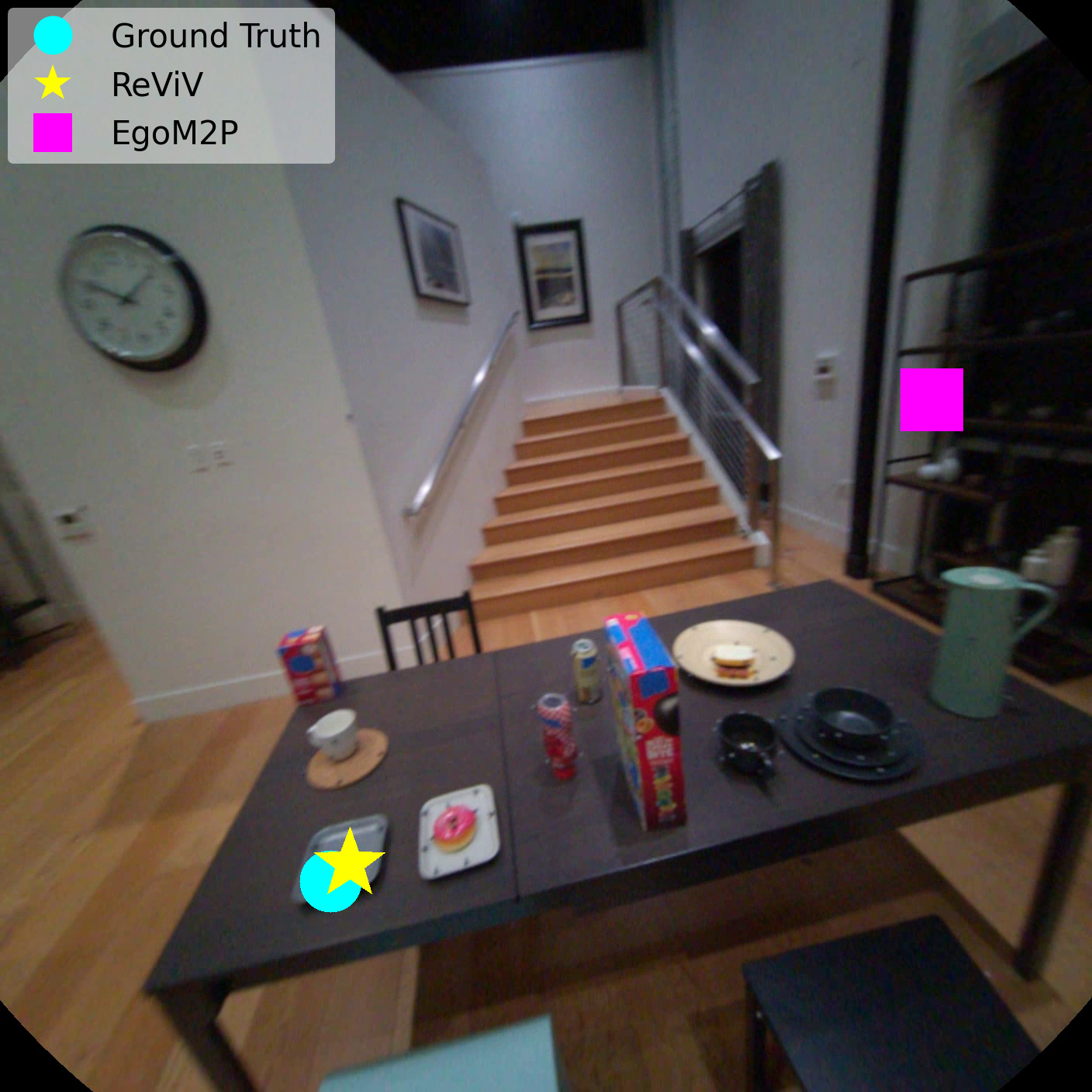}
        \end{subfigure}
        \hfill
        \begin{subfigure}{0.48\linewidth}
            \includegraphics[width=\linewidth]{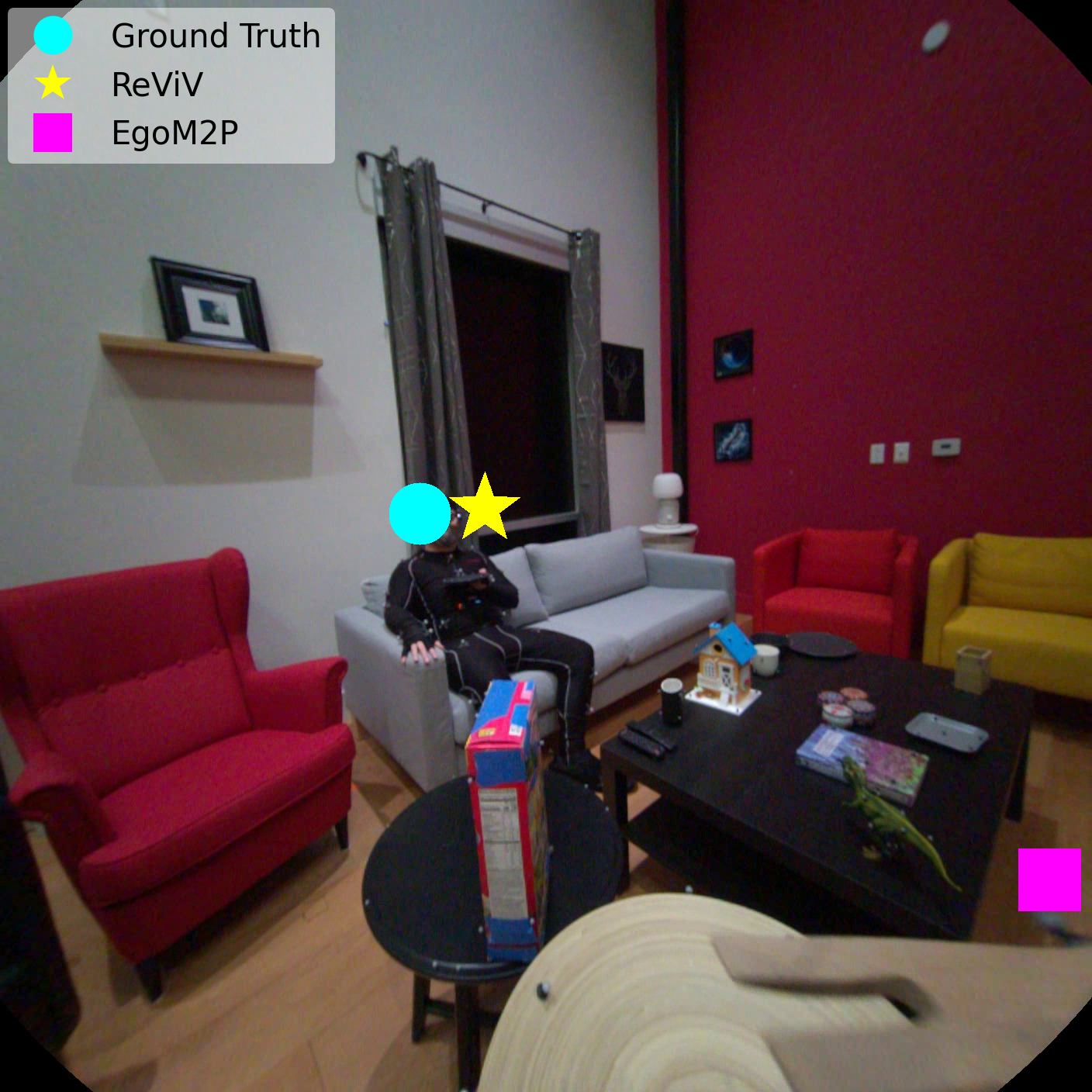}
        \end{subfigure}
        \caption{\textbf{Qualitative visualization of egocentric gaze estimation results.}}
        \label{fig:gaze}
    \end{minipage}
\end{figure}

\subsection{Egocentric Camera Tracking}
As illustrated in Fig.~\ref{fig:cam_traj}, \method produces smooth and accurate camera trajectories that closely follow the ground truth, even in challenging egocentric scenarios with rapid head movements and limited camera parallax. In contrast, baselines often exhibit drift, trajectory deviations, or temporal jitter.

In Fig. \ref{fig:cam_wireframe}, we provide additional qualitative visualizations of our egocentric camera tracking performance over time using 3D camera frustums. The ground truth camera poses are denoted by black wireframes, while the predicted poses are shown in color (red for our method, ReViV). As demonstrated across different timesteps (Frames 05, 25, and 45) in diverse indoor sequences, ReViV yields camera poses that remain tightly aligned with the ground truth throughout the temporal window. In contrast, the baseline methods (EgoM2P, EgoMono4D, and VIPE) struggle to maintain spatial consistency. As the sequences progress, these baselines increasingly exhibit noticeable translational drift and rotational misalignment from the black ground truth wireframes, whereas our approach successfully preserves accurate tracking.

\subsection{Egocentric Gaze Estimation}
Qualitatively, Fig.~\ref{fig:gaze} demonstrates \method produces closer gaze predictions, 
indicating a stronger capability to infer human intent from egocentric observations.

\subsection{Egocentric Body Motion Reconstruction}

\begin{figure}[htbp]
\centering
\setlength{\tabcolsep}{1pt}
\renewcommand{\arraystretch}{0.5}

\begin{tabular}{cccccc}
    & Input & GT & Ours & \parbox[b]{0.18\linewidth}{\centering EgoAllo \cite{yi2025egoallo} w/ VIPE cam} & \parbox[b]{0.18\linewidth}{\centering UniEgoMotion \cite{patel2025uniegomotion} w/ VIPE cam} \\[2pt]
    \rotatebox{90}{\parbox{0.18\linewidth}{\centering Frame 05}} &
    \includegraphics[width=0.18\linewidth]{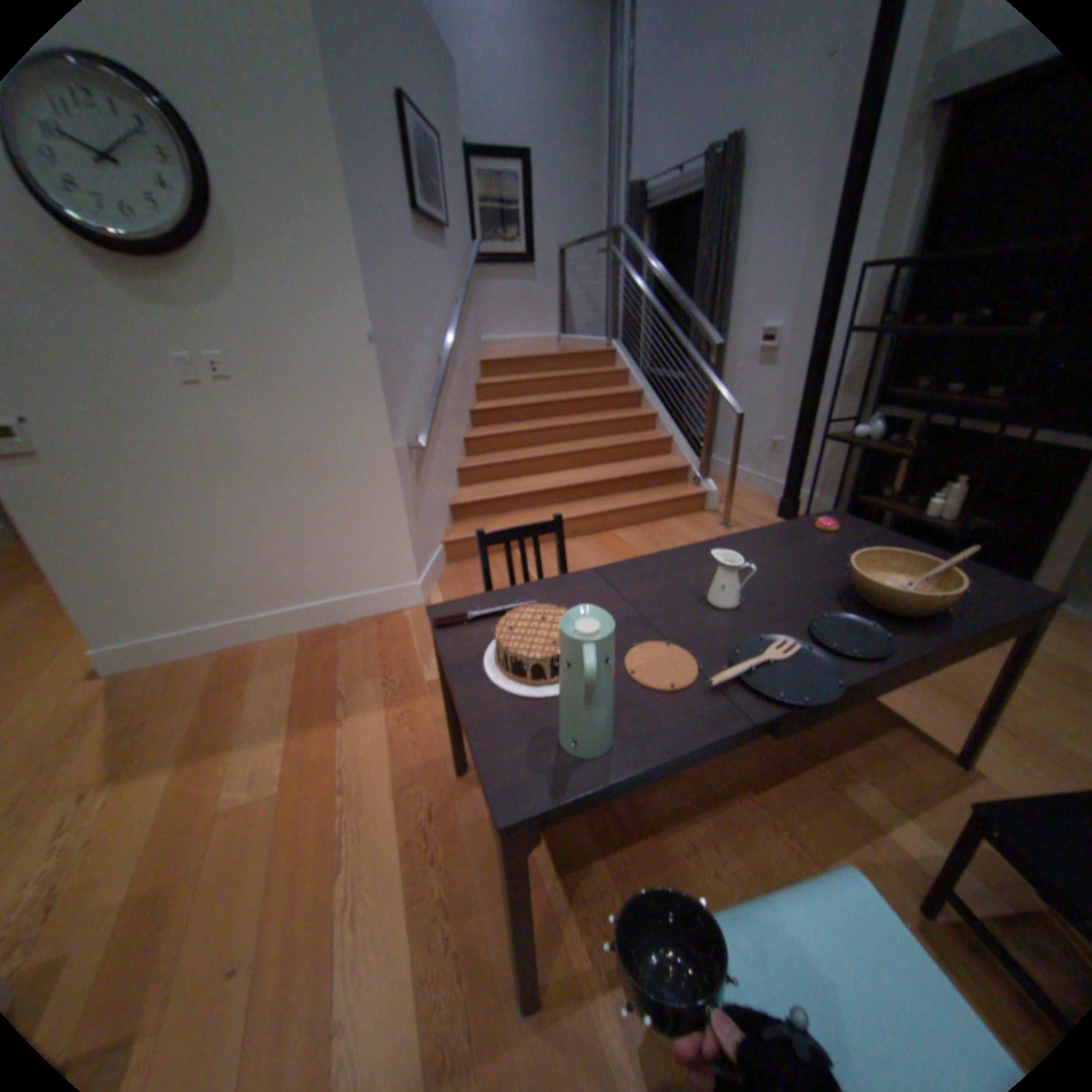} &
    \includegraphics[width=0.18\linewidth]{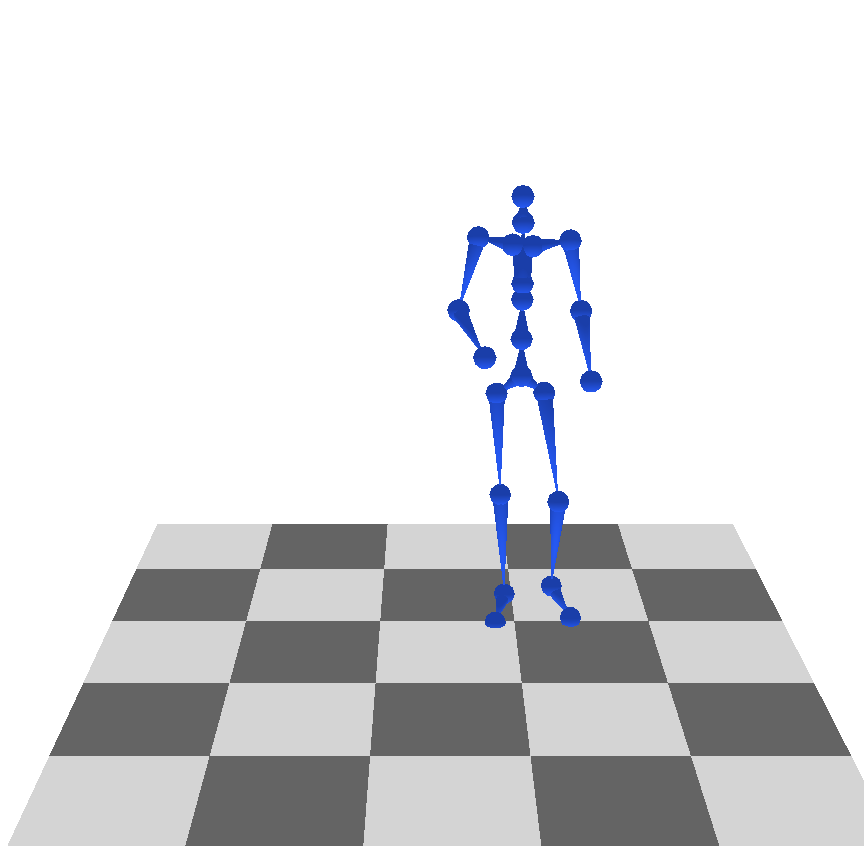} &
    \includegraphics[width=0.18\linewidth]{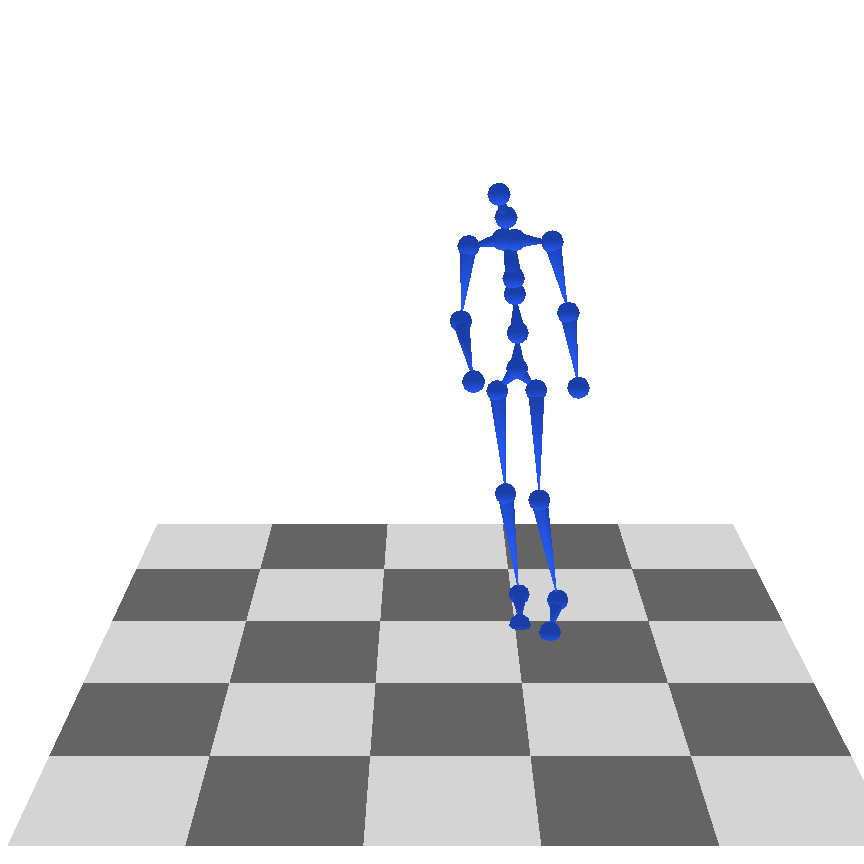} &
    \includegraphics[width=0.18\linewidth]{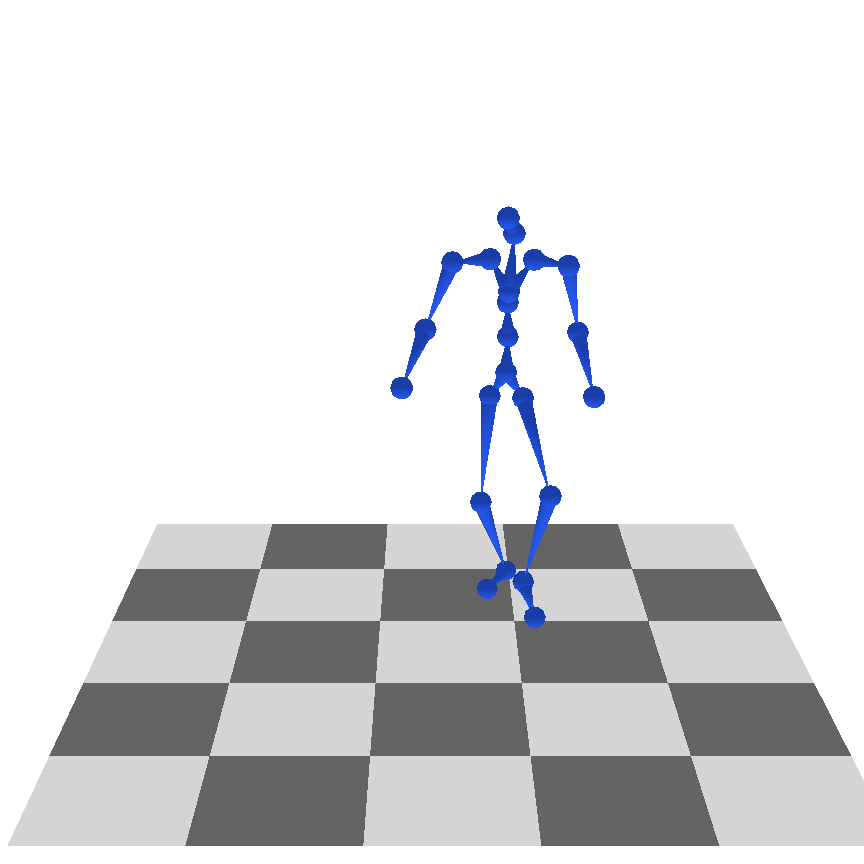} &
    \includegraphics[width=0.18\linewidth]{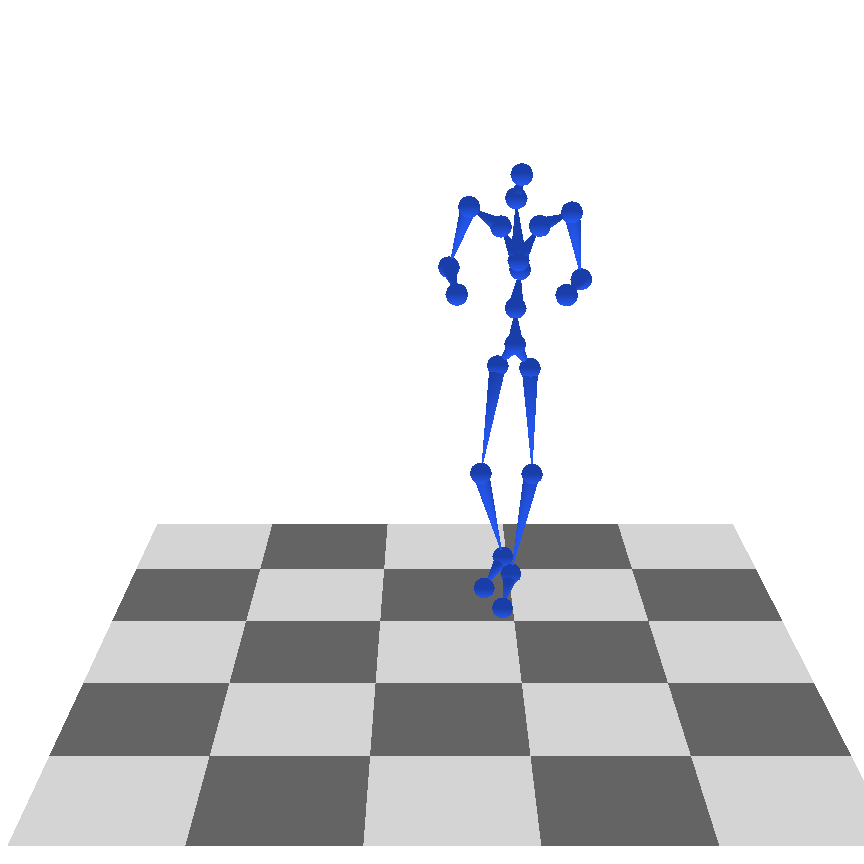} \\
    \rotatebox{90}{\parbox{0.18\linewidth}{\centering Frame 30}} &
    \includegraphics[width=0.18\linewidth]{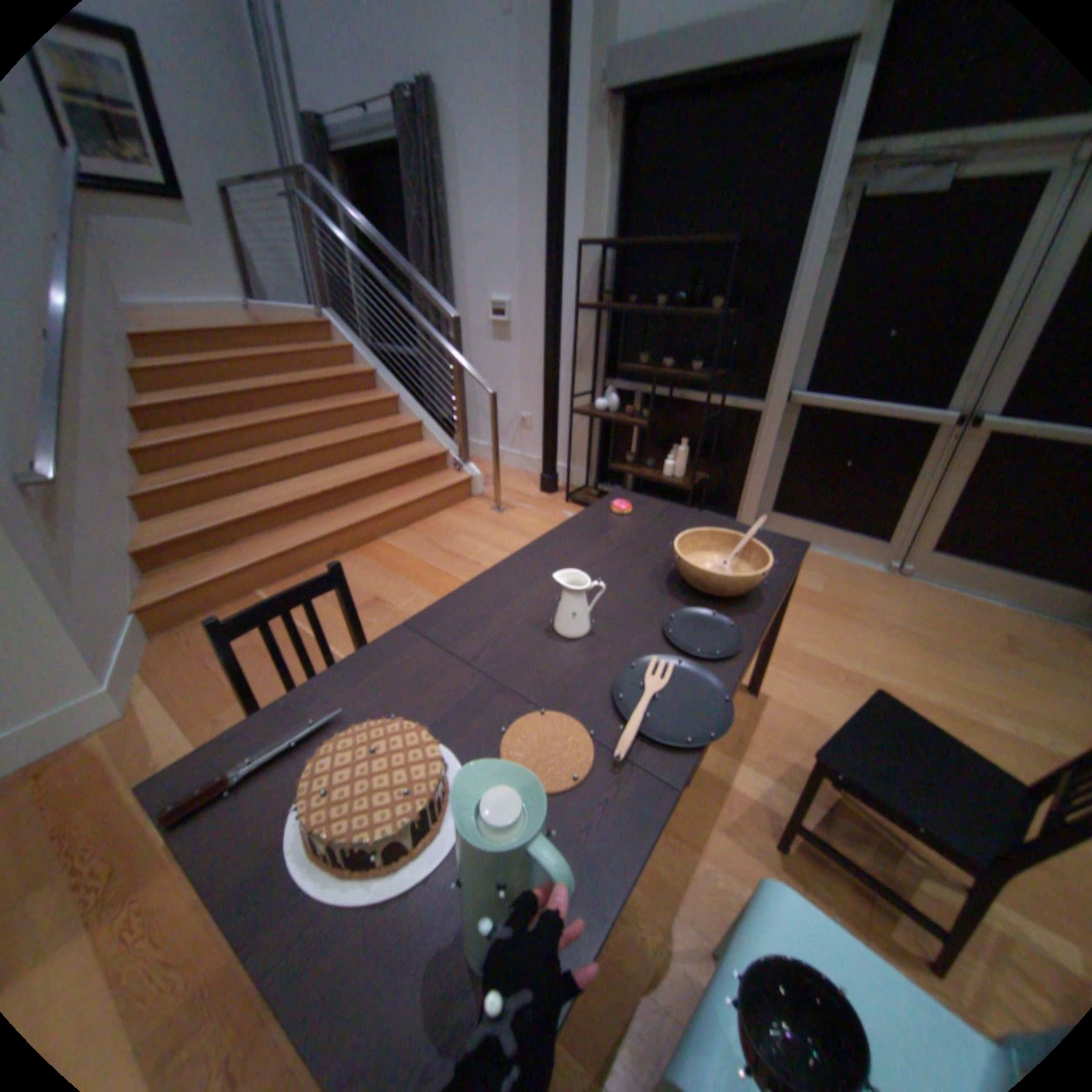} &
    \includegraphics[width=0.18\linewidth]{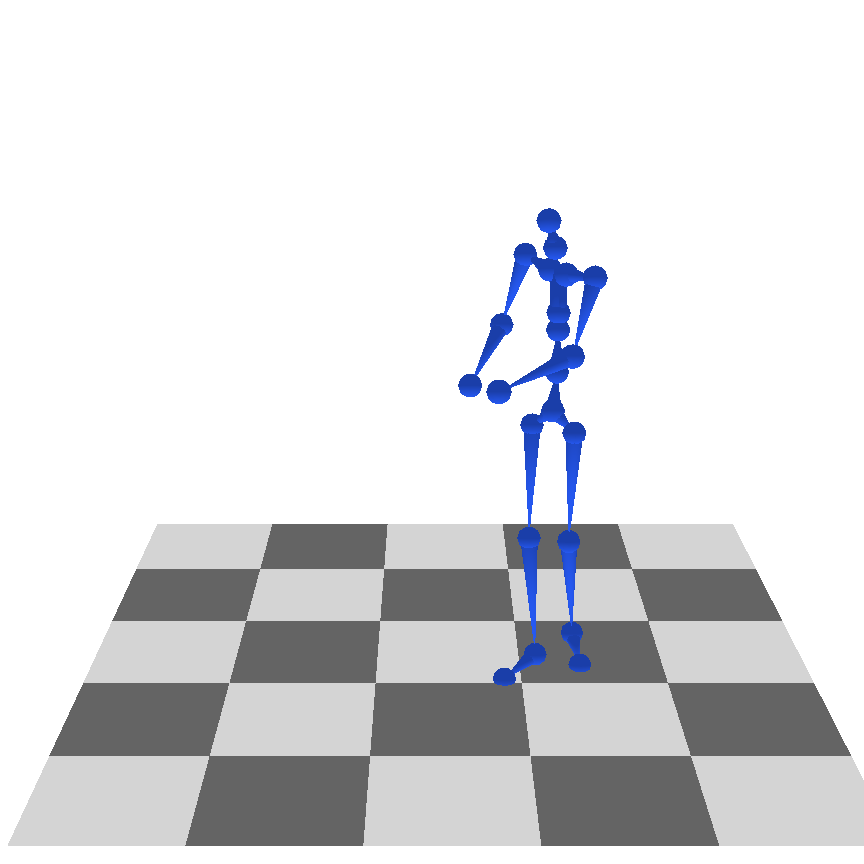} &
    \includegraphics[width=0.18\linewidth]{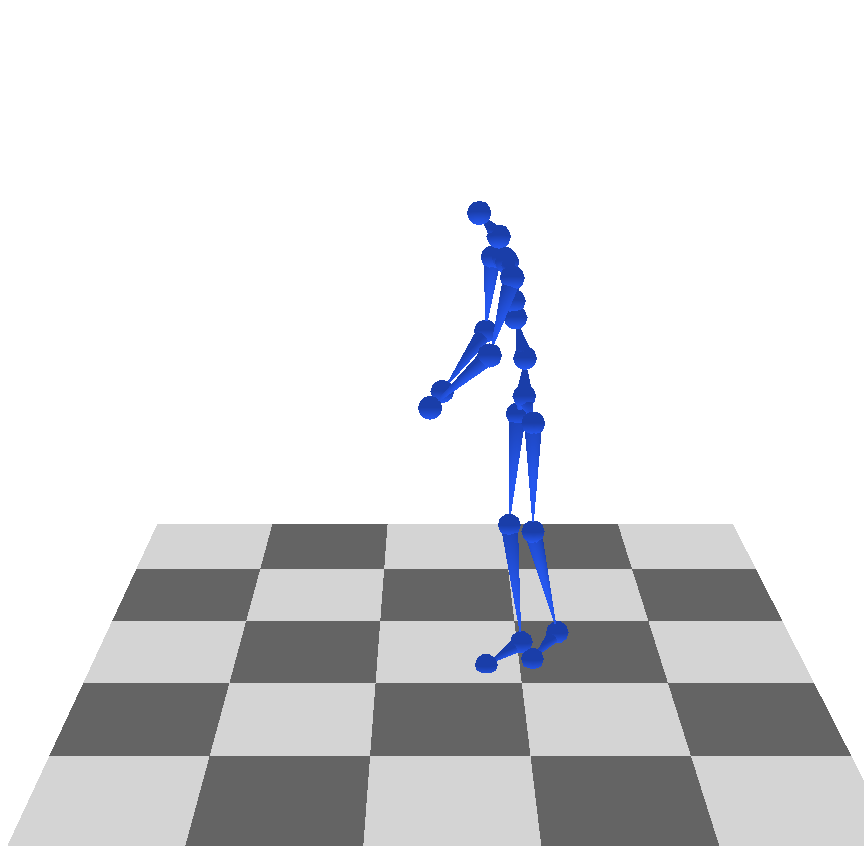} &
    \includegraphics[width=0.18\linewidth]{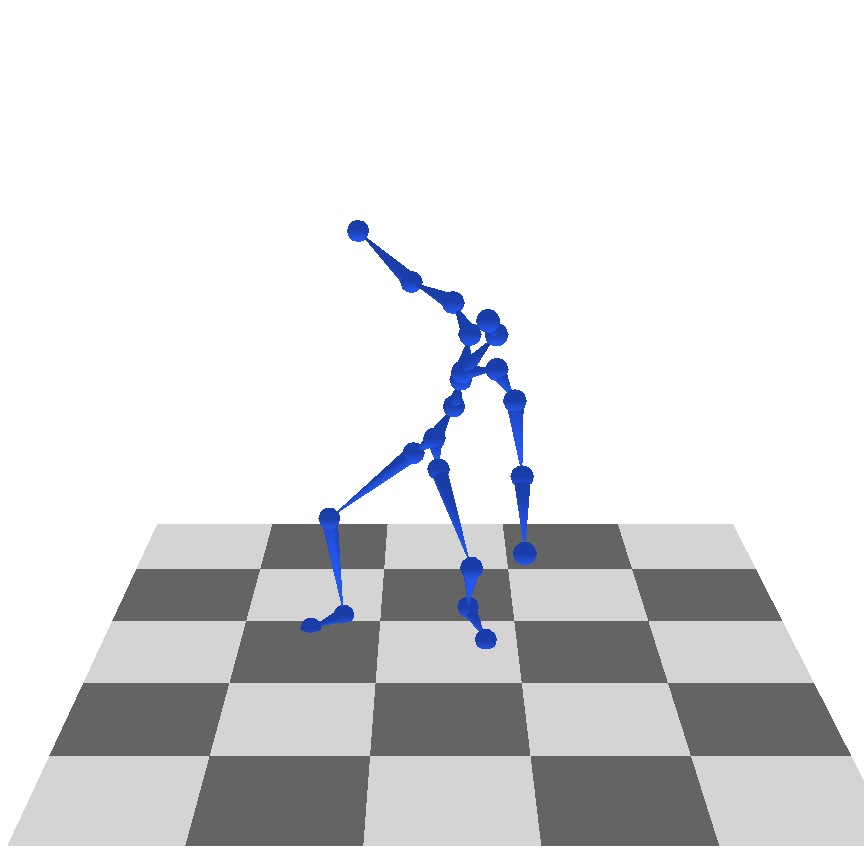} &
    \includegraphics[width=0.18\linewidth]{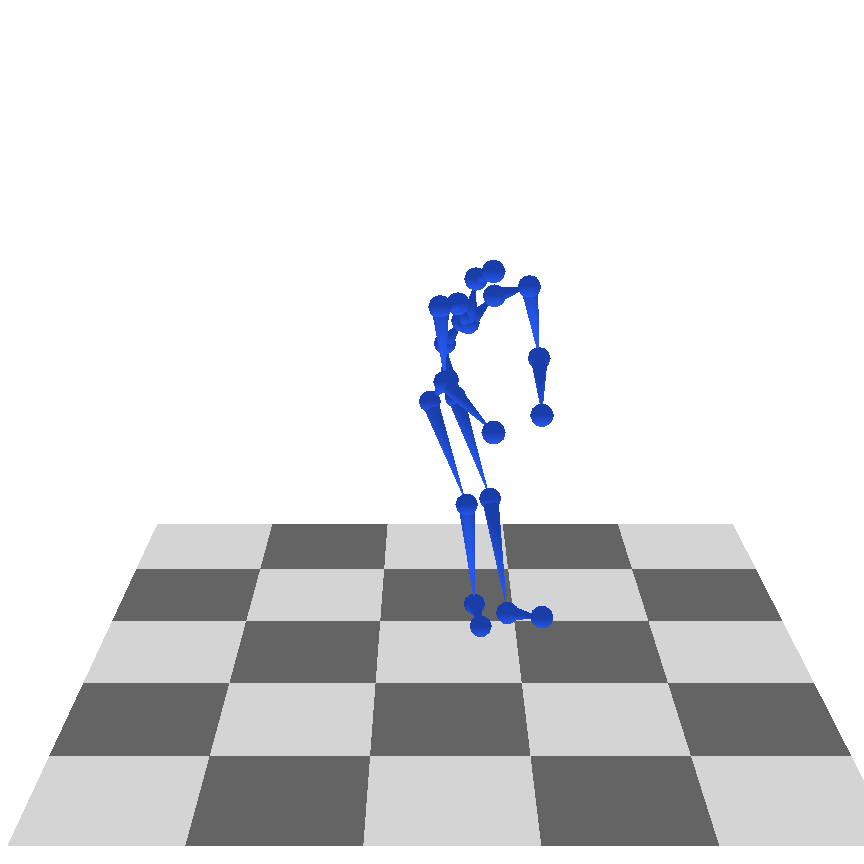} \\
    \rotatebox{90}{\parbox{0.18\linewidth}{\centering Frame 55}} &
    \includegraphics[width=0.18\linewidth]{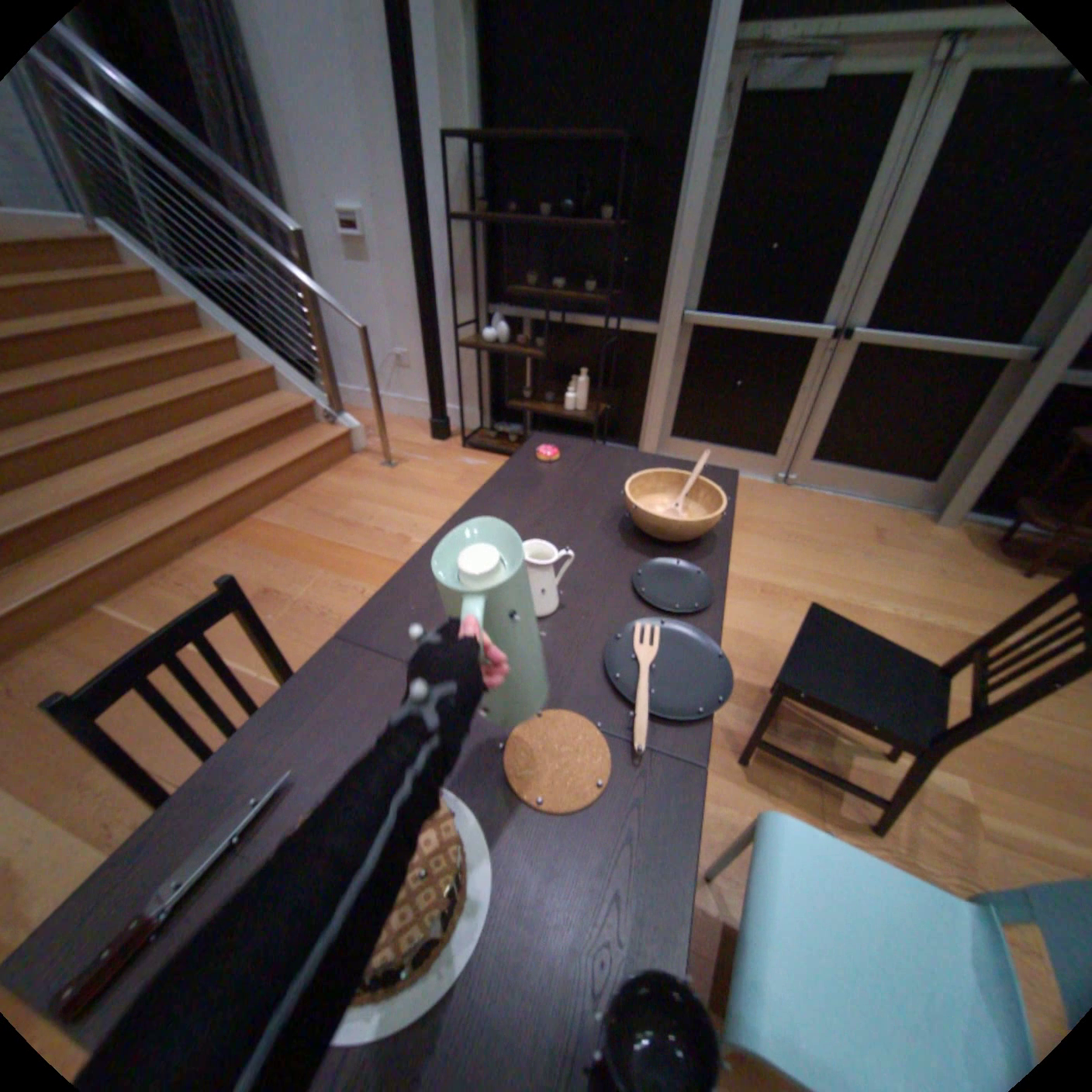} &
    \includegraphics[width=0.18\linewidth]{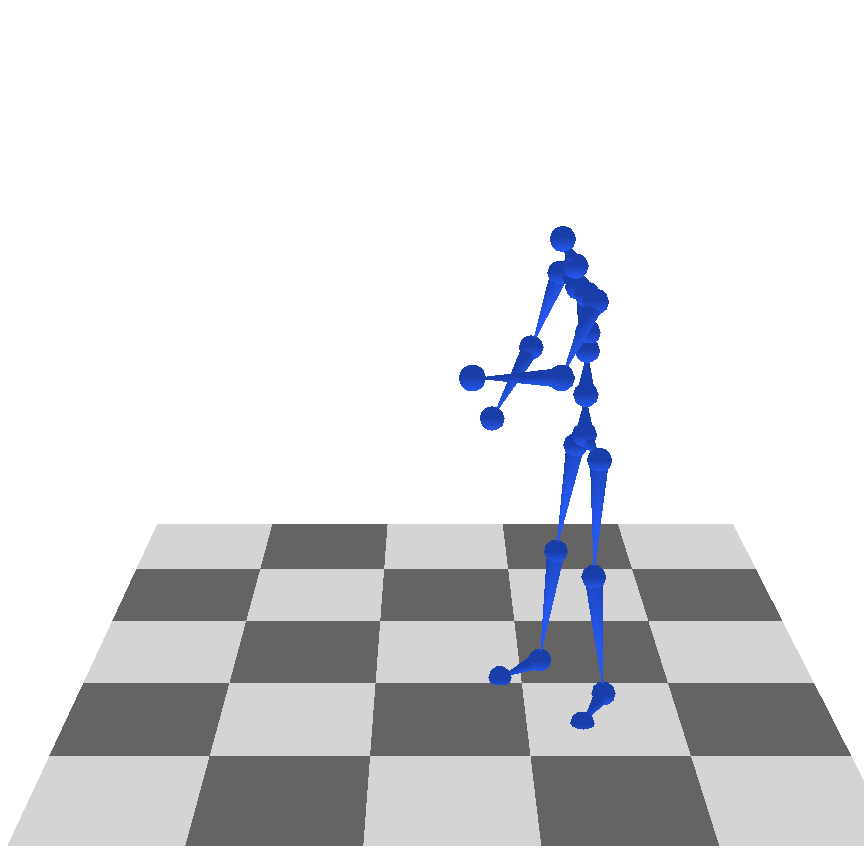} &
    \includegraphics[width=0.18\linewidth]{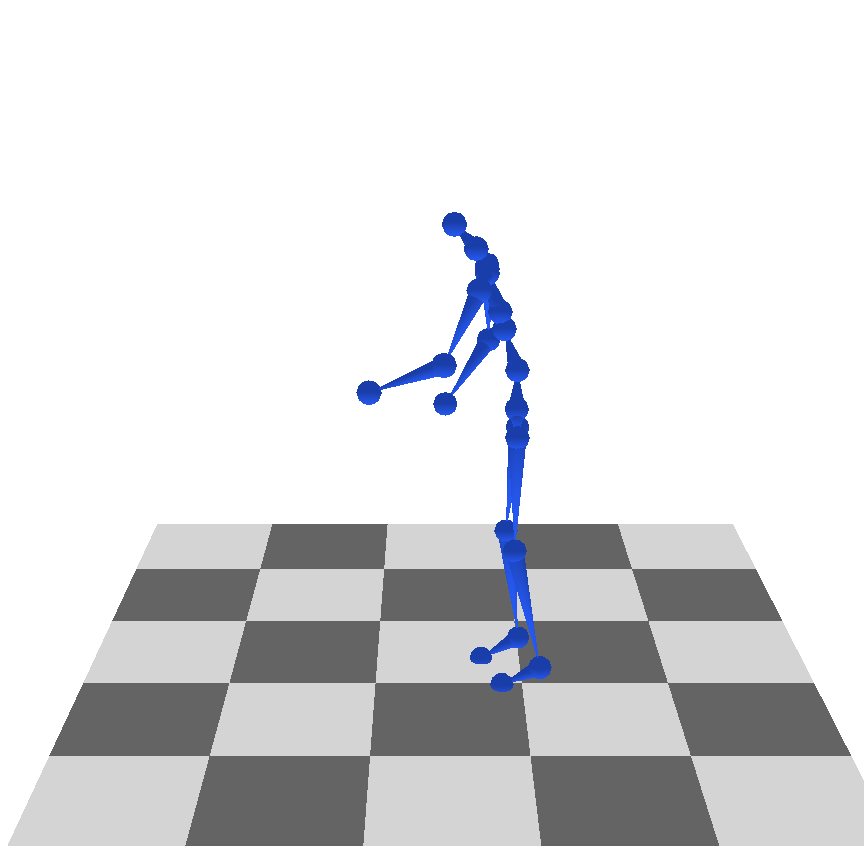} &
    \includegraphics[width=0.18\linewidth]{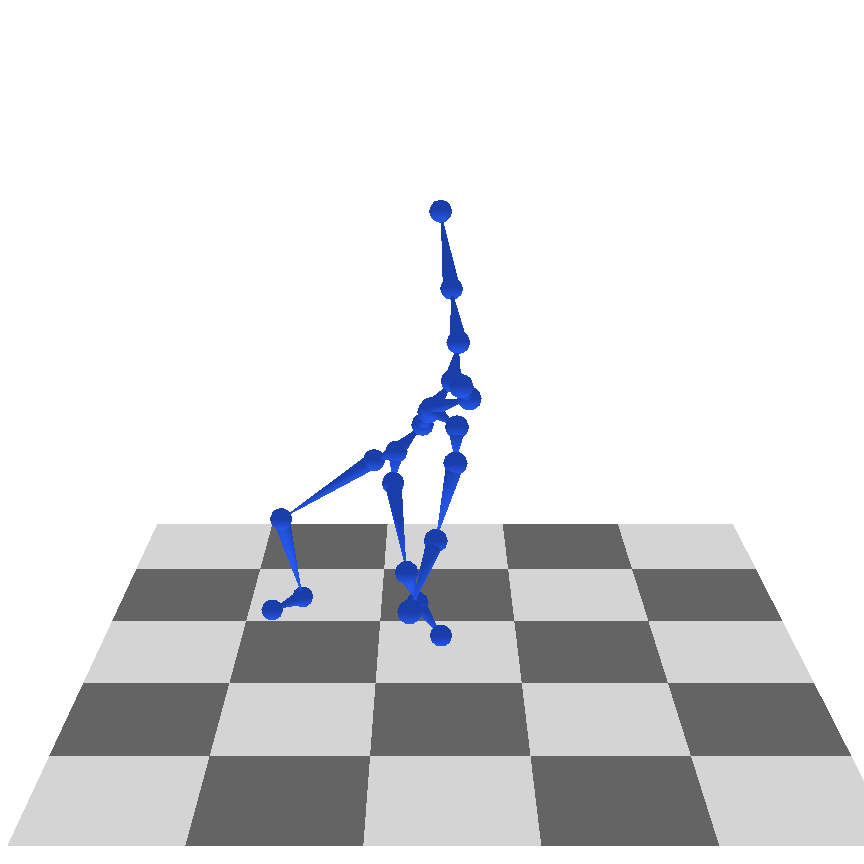} &
    \includegraphics[width=0.18\linewidth]{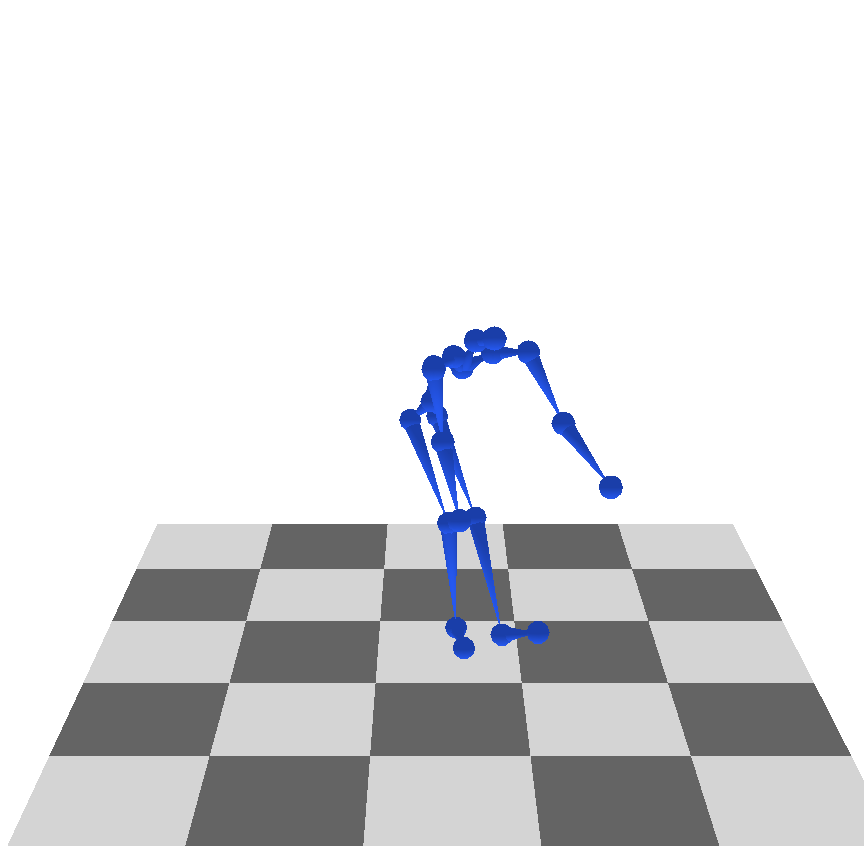} \\
\end{tabular}

\begin{tabular}{cccccc}
    & Input & GT & Ours & \parbox[b]{0.18\linewidth}{\centering EgoAllo \cite{yi2025egoallo} w/ GT cam} & \parbox[b]{0.18\linewidth}{\centering UniEgoMotion \cite{patel2025uniegomotion} w/ GT cam} \\[2pt]
    \rotatebox{90}{\parbox{0.18\linewidth}{\centering Frame 05}} &
    \includegraphics[width=0.18\linewidth]{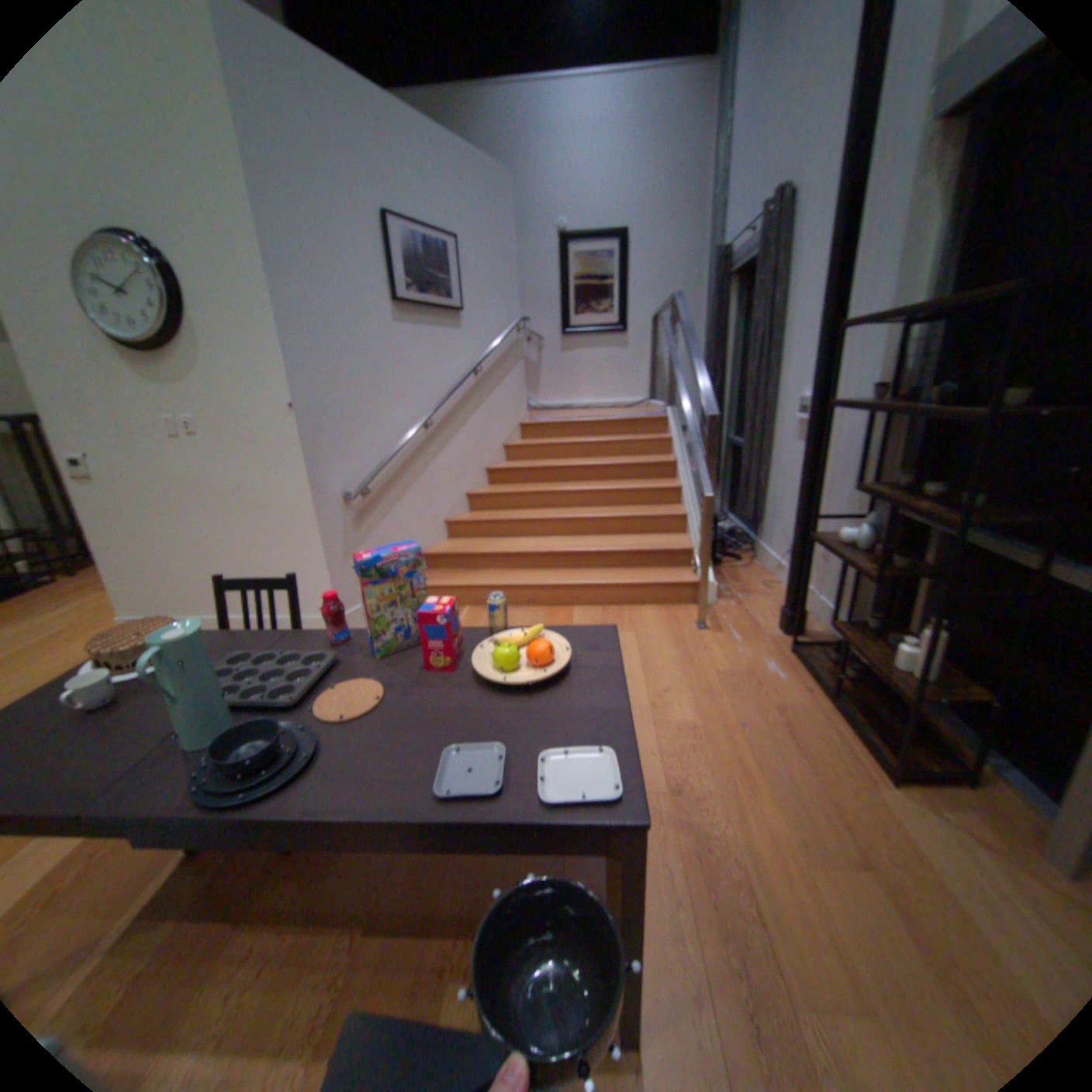} &
    \includegraphics[width=0.18\linewidth]{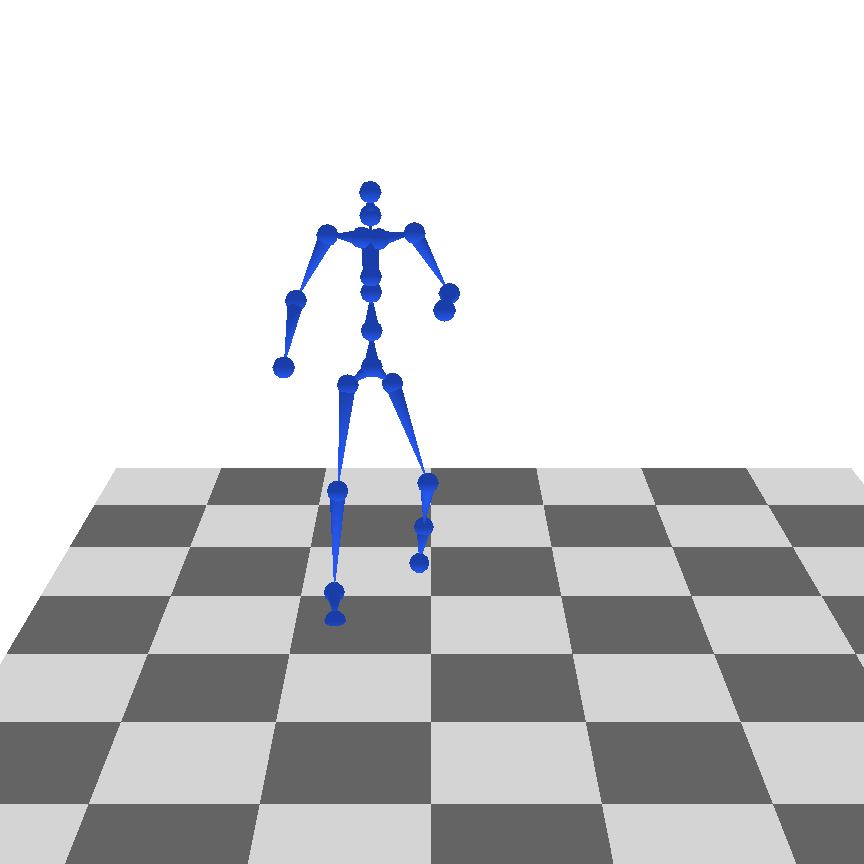} &
    \includegraphics[width=0.18\linewidth]{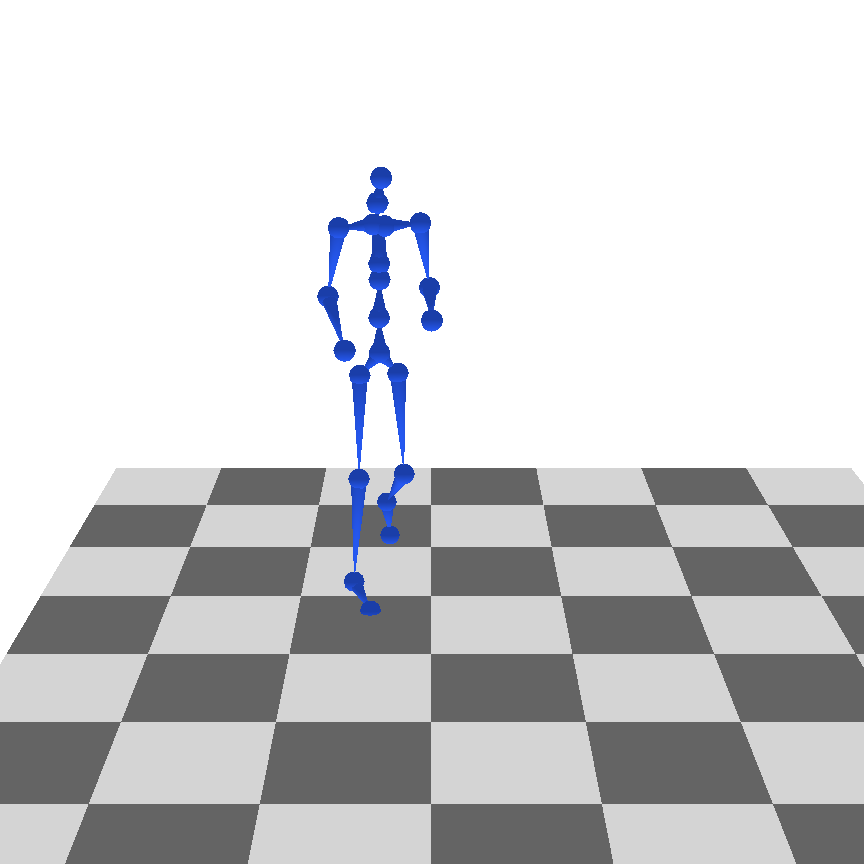} &
    \includegraphics[width=0.18\linewidth]{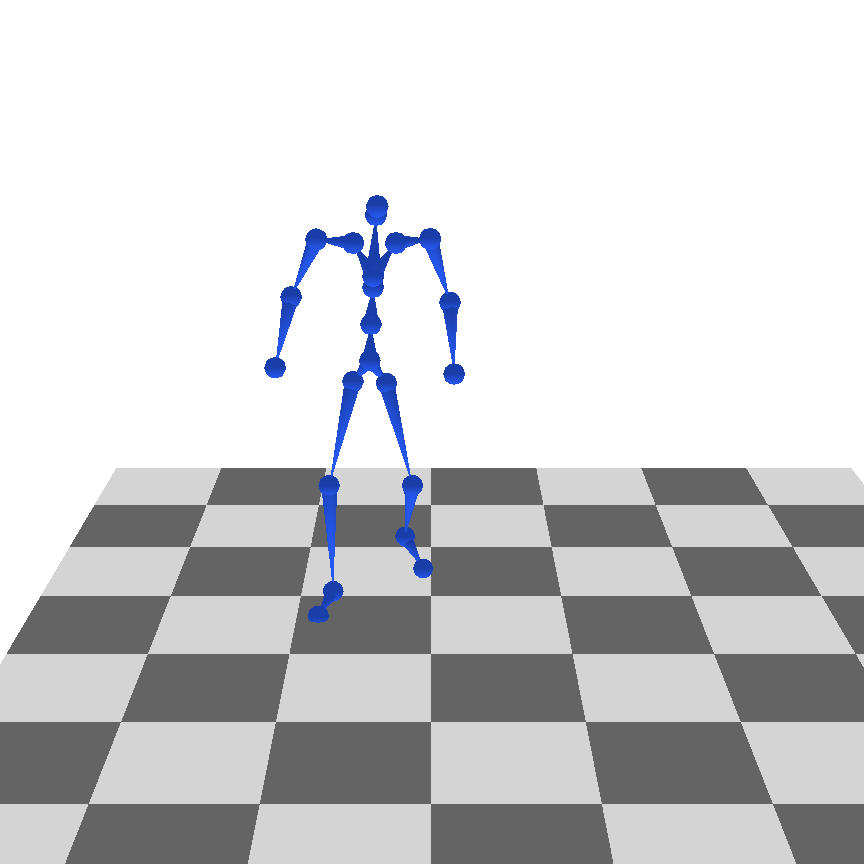} &
    \includegraphics[width=0.18\linewidth]{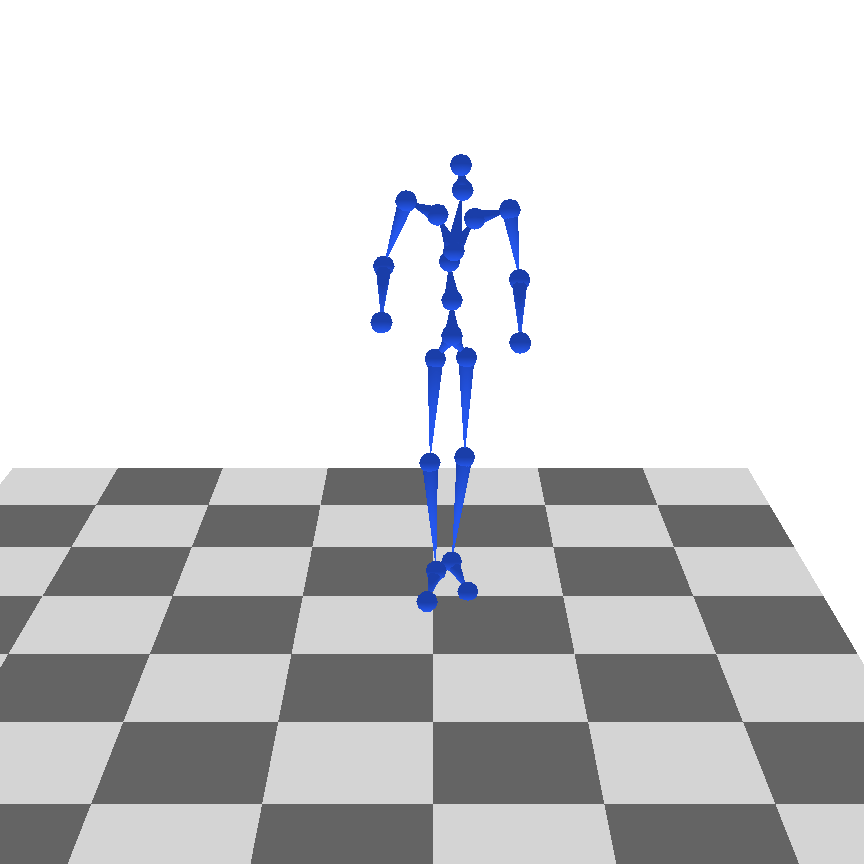} \\
    \rotatebox{90}{\parbox{0.18\linewidth}{\centering Frame 30}} &
    \includegraphics[width=0.18\linewidth]{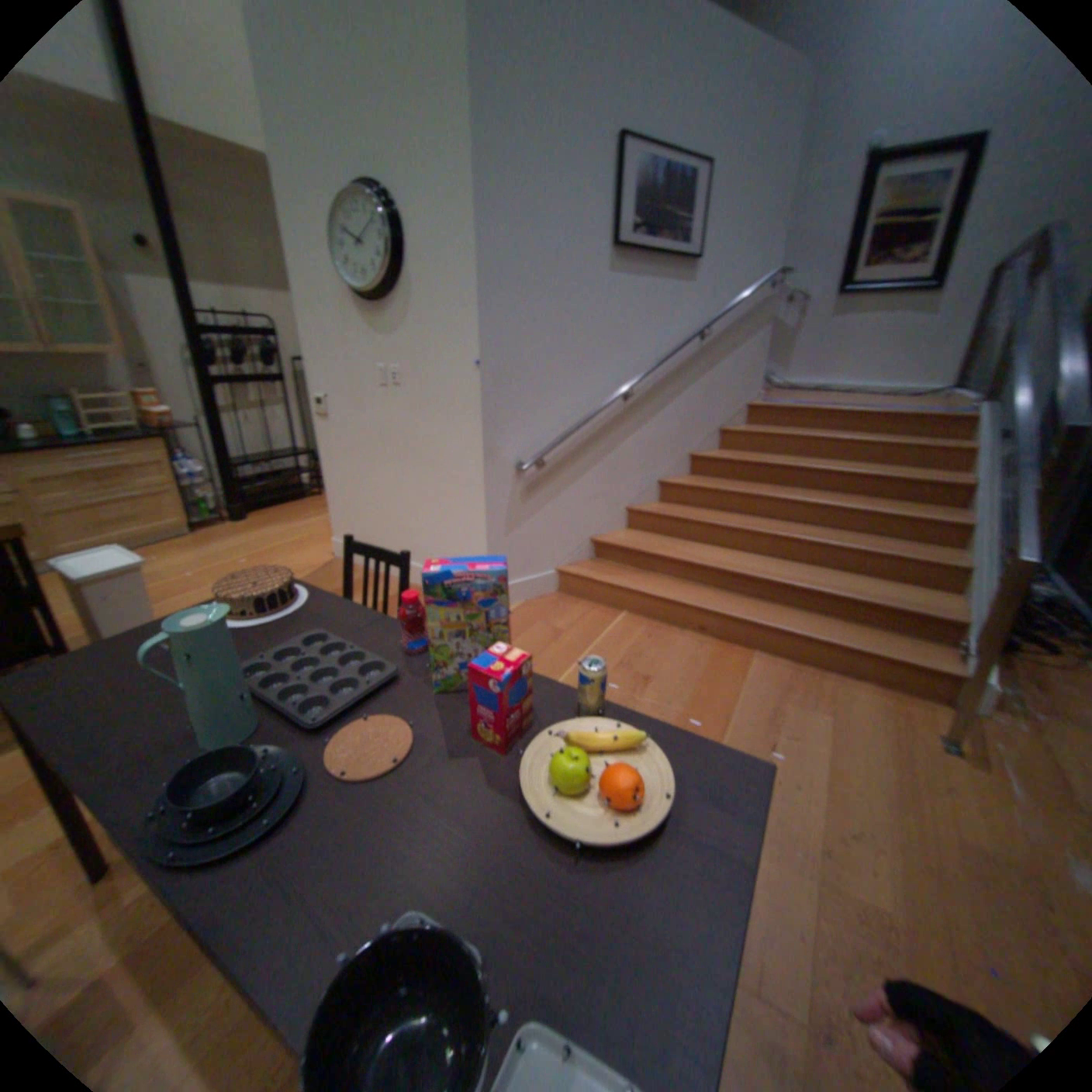} &
    \includegraphics[width=0.18\linewidth]{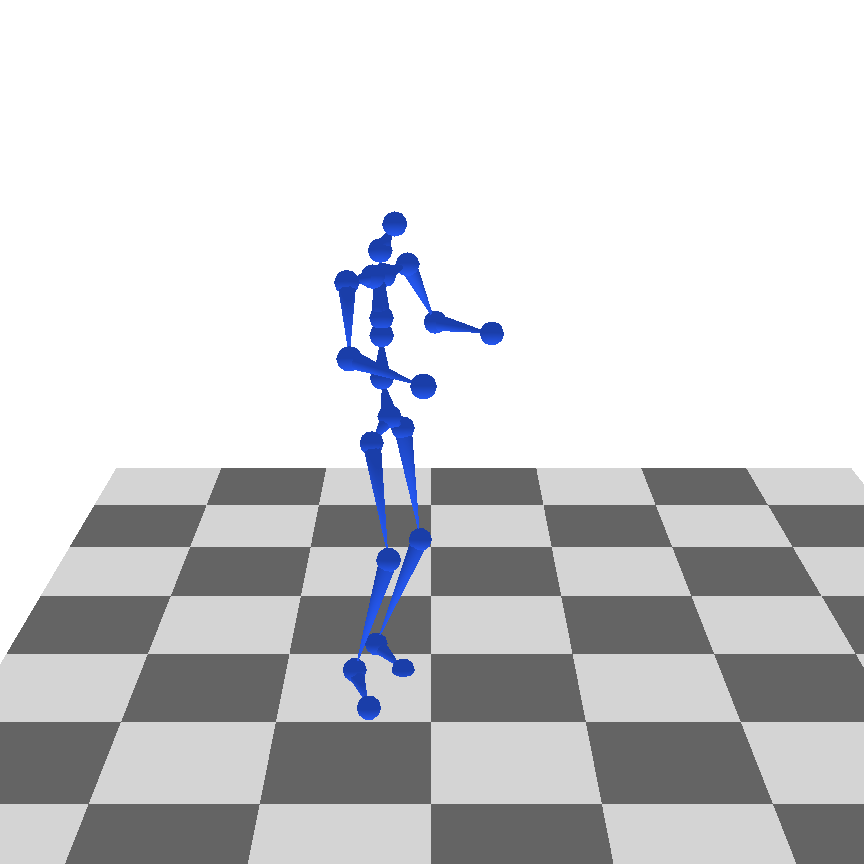} &
    \includegraphics[width=0.18\linewidth]{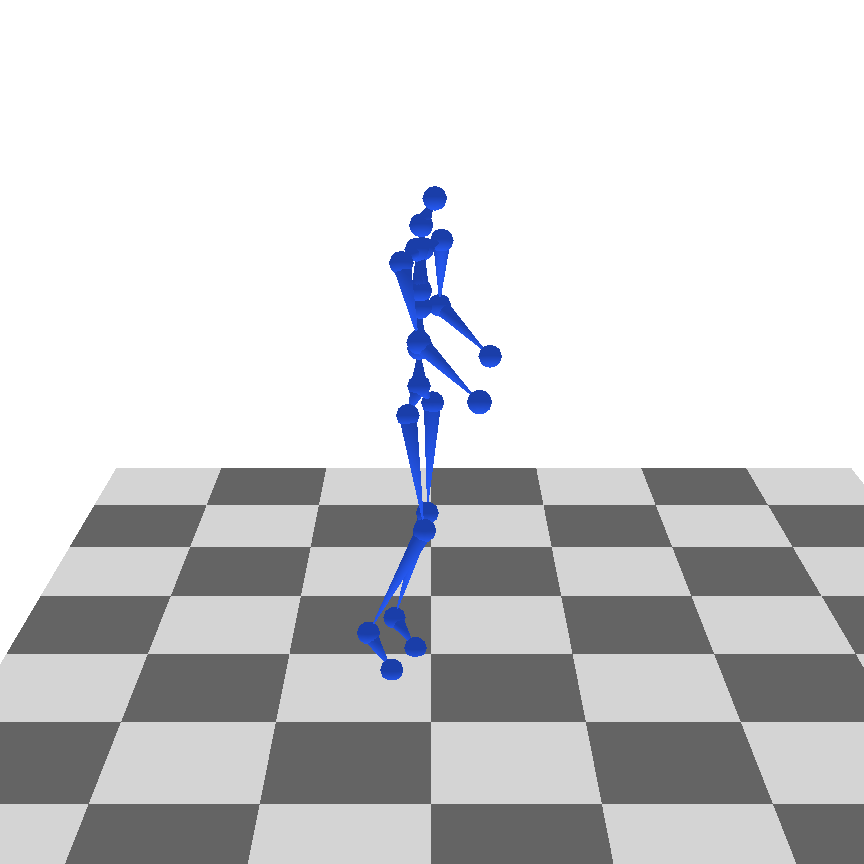} &
    \includegraphics[width=0.18\linewidth]{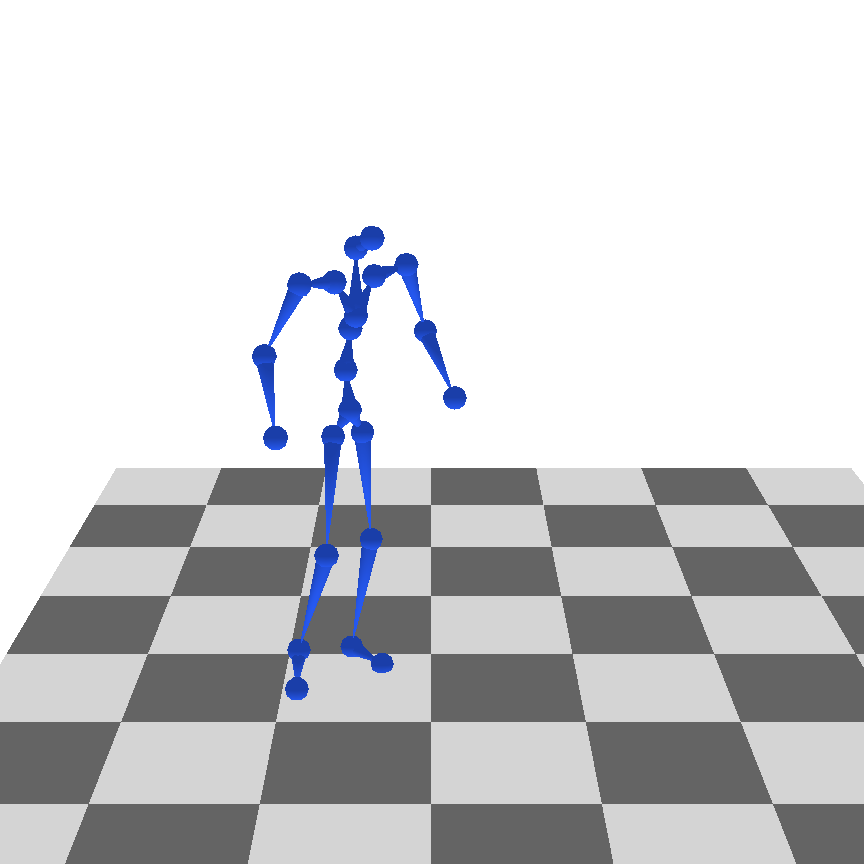} &
    \includegraphics[width=0.18\linewidth]{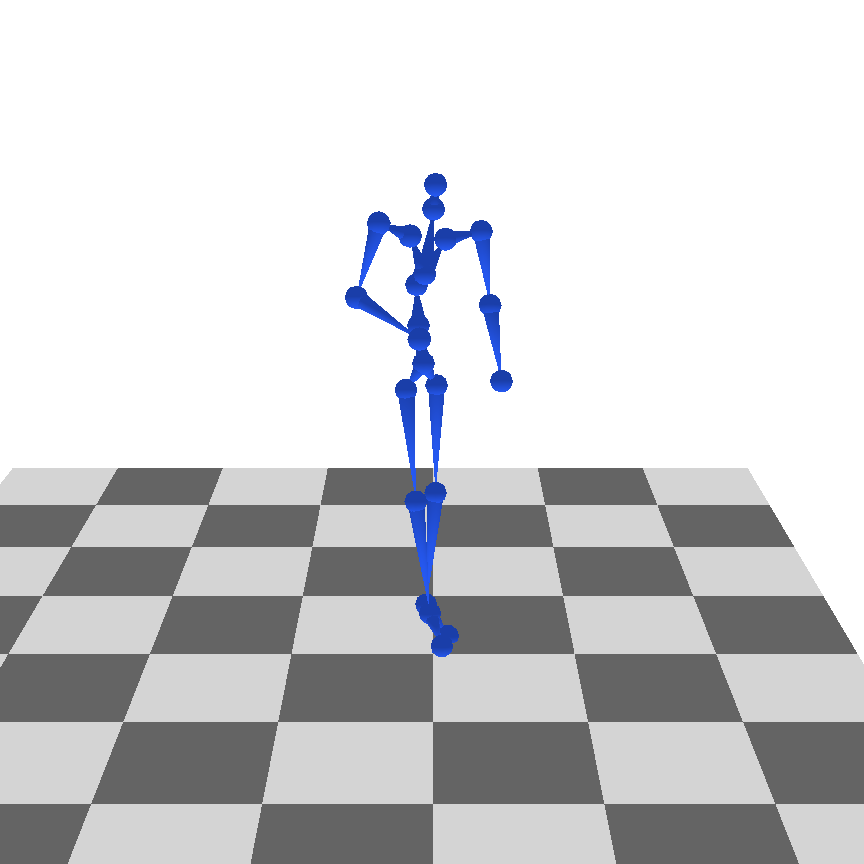} \\
    \rotatebox{90}{\parbox{0.18\linewidth}{\centering Frame 55}} &
    \includegraphics[width=0.18\linewidth]{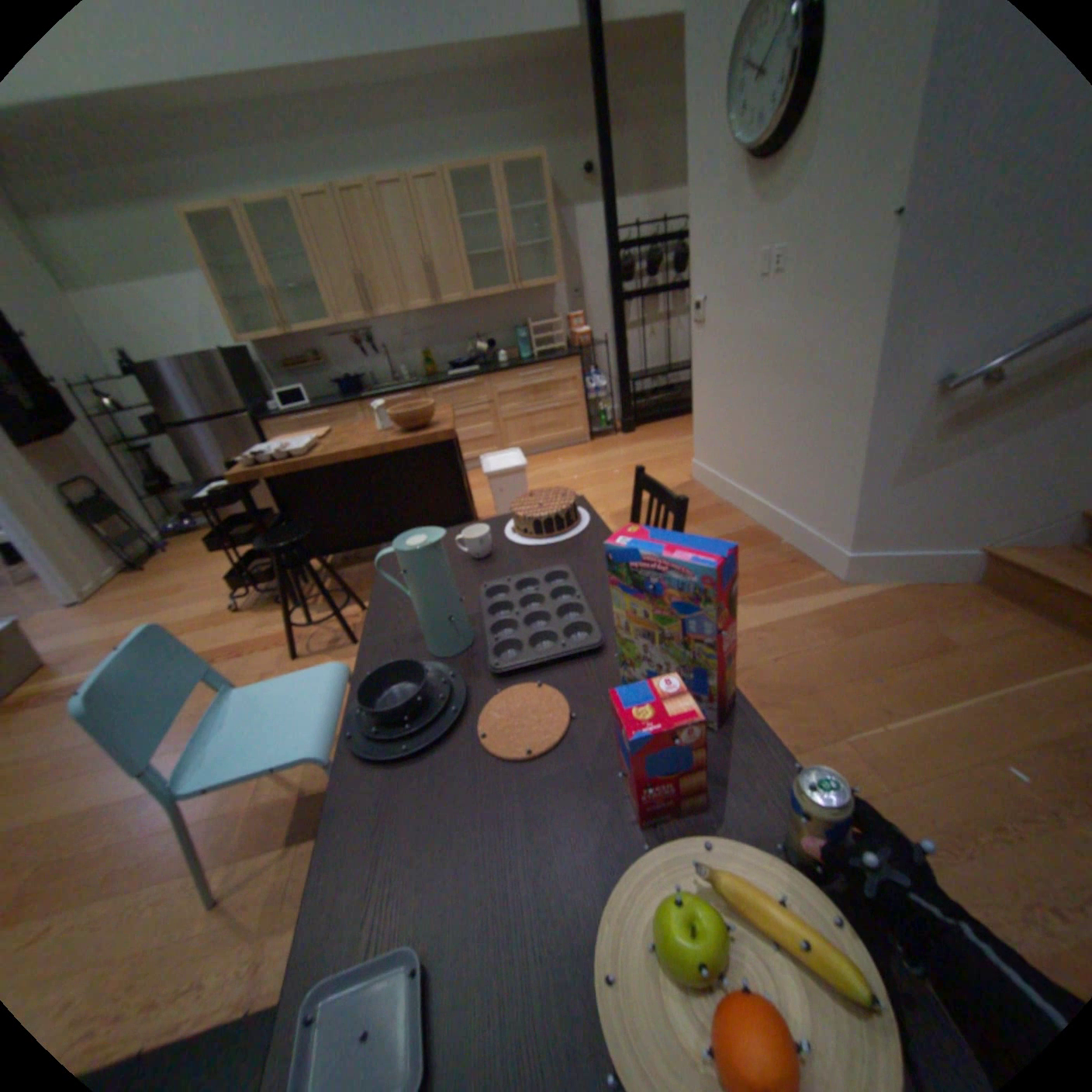} &
    \includegraphics[width=0.18\linewidth]{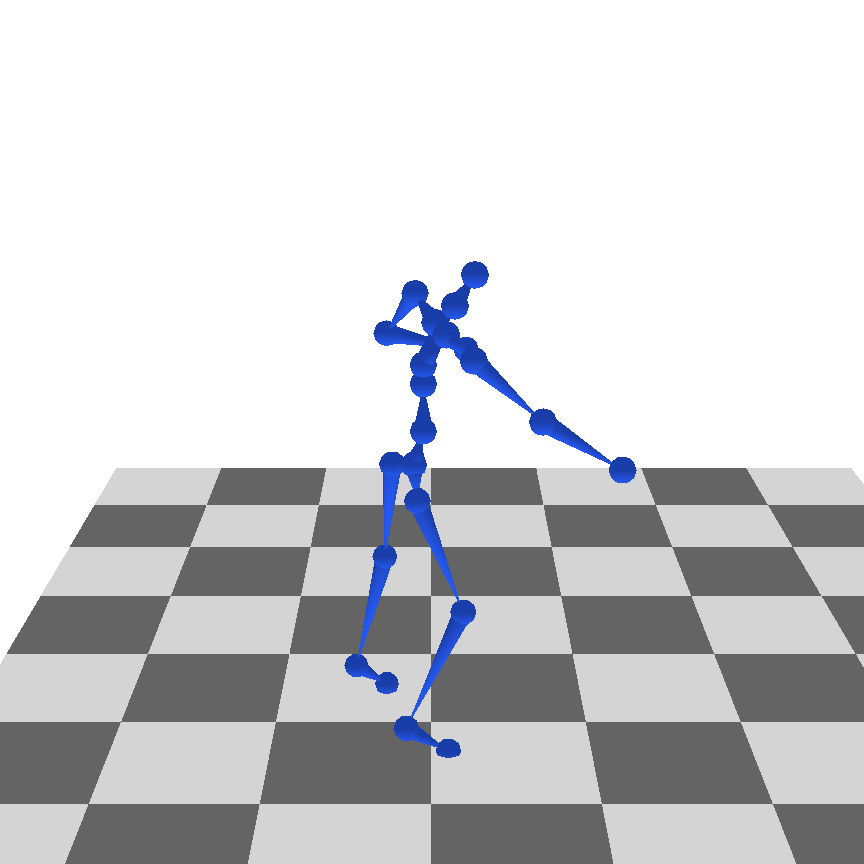} &
    \includegraphics[width=0.18\linewidth]{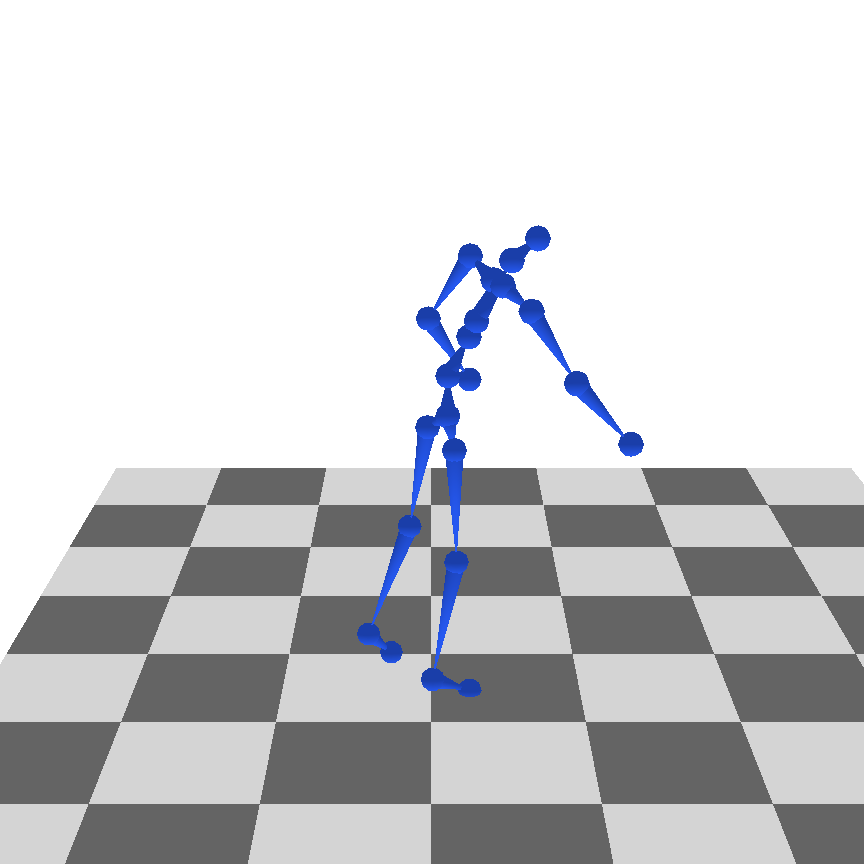} &
    \includegraphics[width=0.18\linewidth]{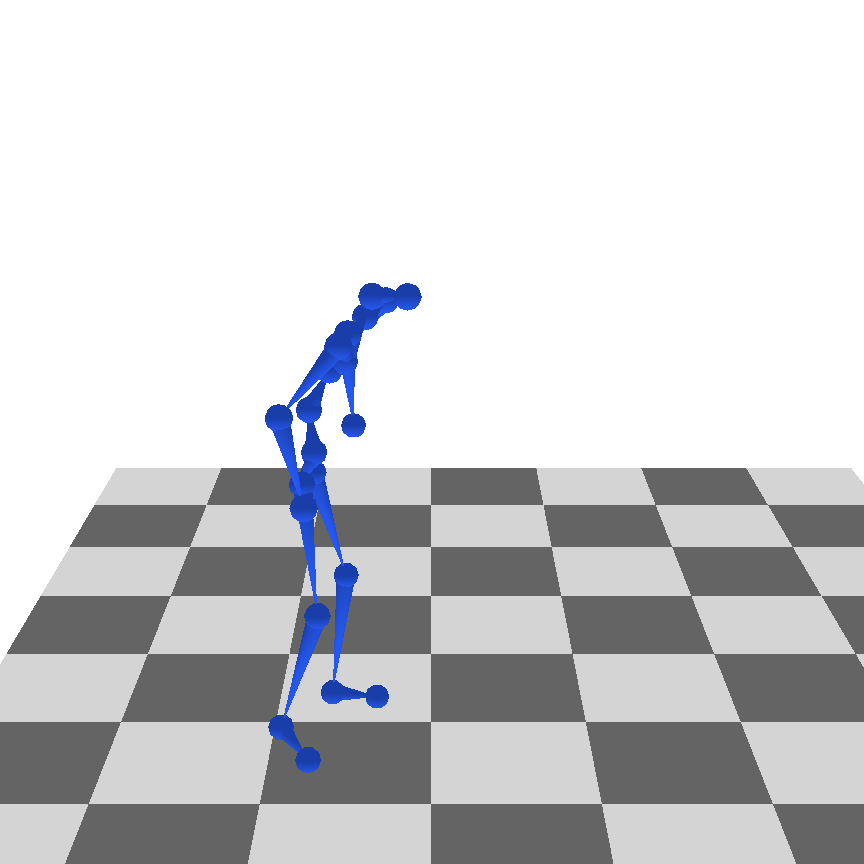} &
    \includegraphics[width=0.18\linewidth]{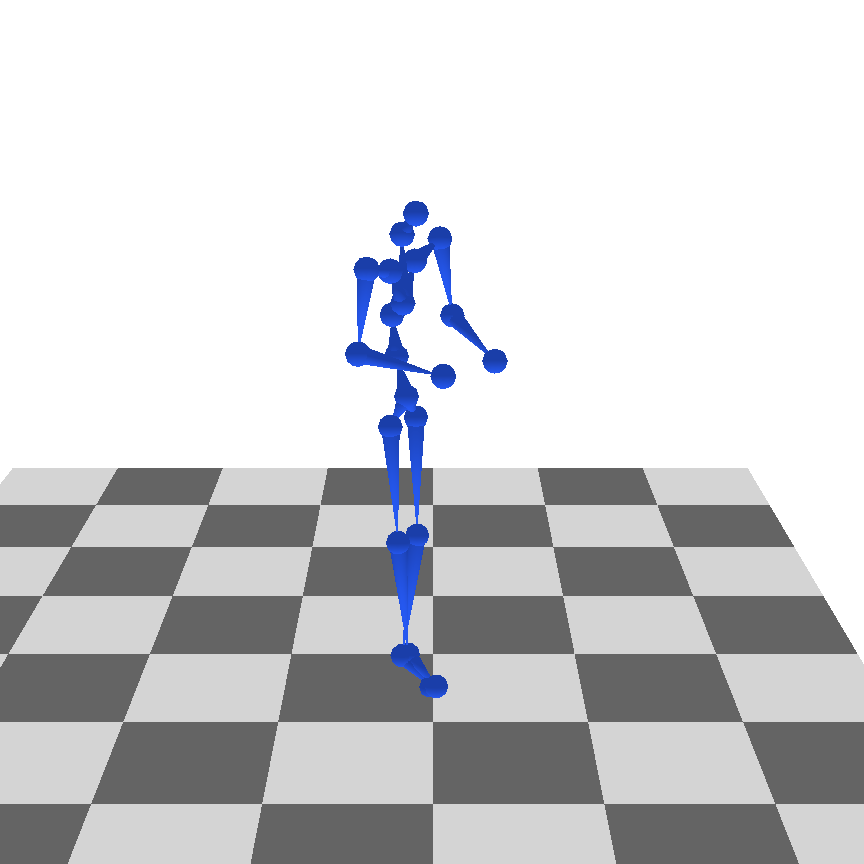} \\
\end{tabular}

\caption{\textbf{Qualitative visualization of egocentric body motion reconstruction.} \method achieves state-of-the-art reconstruction accuracy and semantic alignment using purely monocular egocentric video input. Unlike recent baselines that produce unrealistic ego-body motions when state-of-the-art camera pose estimations (VIPE) fail (\textbf{top}), our approach implicitly learns the joint distribution without relying on explicit camera tracking. This allows our model to produce high-fidelity motions that surpass recent baselines, even when they are given the advantage of ground-truth camera trajectories (\textbf{bottom}).}
\label{fig:body_wireframe}
\end{figure}

In Fig.~\ref{fig:body_wireframe}, we present a detailed frame-by-frame qualitative comparison of egocentric body motion reconstruction on the unseen Aria Digital Twin (ADT) dataset. We compare our method against recent baselines that are conditioned on ground-truth camera trajectories, while our model operates using only monocular egocentric RGB input. As illustrated across multiple timesteps (Frames 05, 30, and 55) spanning diverse indoor daily activities, \method reconstructs human motion that is more faithful to the visual evidence in the input frames. For example, in the first sequence, our method accurately captures the motion of turning right and stop in front of the table, while in the second sequence it correctly reconstructs the action of turning left and reaching for the item.

Notably, \method demonstrates significantly stronger generalization to this unseen dataset compared to UniEgoMotion~\cite{patel2025uniegomotion}, which we attribute to our large-scale multi-modal pretraining. In contrast, baselines such as EgoAllo~\cite{yi2025egoallo} often exhibit structural deviations from the semantics of the input egocentric video. This limitation largely arises because EgoAllo relies on restricted visual signals, primarily leveraging off-the-shelf priors for localized hand guidance rather than modeling the holistic scene context. By instead leveraging full-image semantics directly from raw egocentric video and implicitly learning the joint distribution of body motion and visual observations, \method achieves state-of-the-art ego body motion reconstruction accuracy with strong semantic alignment.

\section{Additional Ablation Studies}

\subsection{Egocentric Hand Motion Reconstruction Ablation}

Tab.~\ref{tab:ablation_hand} extends the body ablation from the main paper to egocentric hand motion reconstruction across four benchmarks: HoloAssist, HOT3D, ARCTIC, and TACO.

\begin{table}[ht]
    \centering
    \scriptsize
    \renewcommand{\arraystretch}{0.91}
    \setlength{\tabcolsep}{0pt}
    \caption{\textbf{Hand reconstruction ablation on multiple benchmarks.} GA/RA/PA denote Global/Relative/Procrustes-aligned MPJPE (mm); lower is better.}
    \label{tab:ablation_hand}
    \begin{tabular*}{\linewidth}{@{\extracolsep{\fill}}lcccccc@{}}
        \toprule
        & \multicolumn{3}{c}{HoloAssist} & \multicolumn{3}{c}{HOT3D} \\
        \cmidrule(lr){2-4} \cmidrule(lr){5-7}
        Methods & GA $\downarrow$ & RA $\downarrow$ & PA $\downarrow$ & GA $\downarrow$ & RA $\downarrow$ & PA $\downarrow$ \\
        \midrule
        \modelFull & \textbf{23.5} & 29.7 & \textbf{10.5} & \textbf{38.2} & \textbf{48.1} & \textbf{13.6} \\
        \modelHandData & 26.4 & \textbf{28.6} & 10.5 & 39.1 & 48.3 & 14.4 \\
        \modelHandTask & 38.7 & 48.7 & 15.7 & 82.0 & 73.9 & 21.7 \\
        \modelAlpha & 34.9 & 43.9 & 13.7 & 60.9 & 58.6 & 16.1 \\
        \modelUniform & 37.7 & 41.2 & 13.2 & 70.1 & 65.0 & 17.8 \\
        \modelTiny & 33.8 & 42.1 & 14.0 & 54.2 & 59.9 & 16.9 \\
        \midrule
        & \multicolumn{3}{c}{ARCTIC} & \multicolumn{3}{c}{TACO} \\
        \cmidrule(lr){2-4} \cmidrule(lr){5-7}
        \modelFull & \textbf{35.1} & \textbf{33.0} & \textbf{13.7} & \textbf{20.3} & \textbf{25.2} & \textbf{9.4} \\
        \modelHandData & 42.2 & 39.2 & 15.5 & 32.8 & 36.9 & 13.0 \\
        \modelHandTask & 75.8 & 77.2 & 22.6 & 53.2 & 52.1 & 17.9 \\
        \modelAlpha & 55.0 & 49.0 & 16.4 & 34.9 & 39.7 & 13.1 \\
        \modelUniform & 67.0 & 60.5 & 19.5 & 42.8 & 48.3 & 15.3 \\
        \modelTiny & 50.8 & 45.6 & 15.8 & 30.0 & 35.8 & 12.1 \\
        \bottomrule
    \end{tabular*}
\end{table}

The hand reconstruction results mirror the trends from the body ablation in the main paper. The task-specific hand specialist (\modelHandTask) fails severely across all four benchmarks, underscoring the critical importance of joint learning over isolated single-modality training. Restricting training to hand-annotated data only (\modelHandData) yields modest degradation, confirming that large-scale datasets without hand annotations still provide useful cross-modal priors through joint distribution learning. Both masking ablations (\modelAlpha, \modelUniform) and the lightweight variant (\modelTiny) consistently underperform \modelFull across all datasets, reinforcing the findings from the main paper.

\subsection{Vision Transformer Encoding Branch}

Quantizing high-frequency visual inputs destroys the fine-grained pixel details necessary for accurate perception and reconstruction. Our full model addresses this by processing raw RGB context through a continuous Vision Transformer branch ($\mathbf{x}_{\text{ViT}}$), bypassing the discrete tokenization bottleneck. Tab.~\ref{tab:vit_branch_ablation} evaluates the impact of removing this branch (w/o $\mathbf{x}_{\text{ViT}}$) across all predicted modalities.

\begin{table*}[ht]
\centering
\scriptsize
\setlength{\tabcolsep}{4pt}
\caption{\textbf{Ablation on ViT encoding branch.} Removing the continuous $\mathbf{x}_{\text{ViT}}$ branch consistently degrades all modalities, confirming its role in preserving high-frequency spatial context beyond what discrete tokens can capture.}
\label{tab:vit_branch_ablation}
\begin{tabular}{l|ccc|c|cc|ccc}
\toprule
& \multicolumn{3}{c|}{Camera} & Gaze & \multicolumn{2}{c|}{Depth} & \multicolumn{3}{c}{Hand (TACO)} \\
\cmidrule(lr){2-4} \cmidrule(lr){5-5} \cmidrule(lr){6-7} \cmidrule(lr){8-10}
Variant & ATE $\downarrow$ & RTE $\downarrow$ & RRE $\downarrow$ & MSE $\downarrow$ & Abs Rel $\downarrow$ & $\delta_{1.25}$ $\uparrow$ & GA $\downarrow$ & RA $\downarrow$ & PA $\downarrow$ \\
\midrule
w/o $\mathbf{x}_{\text{ViT}}$  & 0.018 & \textbf{0.009} & \textbf{1.258} & 0.0233 & 0.296 & 51.7 & 22.0 & 27.0 & 9.6 \\
Full Model               & \textbf{0.015} & \textbf{0.009} & 1.279 & \textbf{0.0211} & \textbf{0.265} & \textbf{56.5} & \textbf{20.3} & \textbf{25.2} & \textbf{9.4} \\
\bottomrule
\end{tabular}
\end{table*}

\begin{itemize}
    \item Depth: Unquantized RGB context is critical for resolving fine 3D geometry, significantly improving both Absolute Relative Error and inlier accuracy ($\delta_{1.25}$).
    \item Gaze Dynamics: Precise localization of visual targets relies on subtle pixel textures. Preserving these details via the continuous ViT noticeably reduces gaze MSE.
    \item Camera Trajectories: Higher-fidelity visual features improve global egomotion consistency, leading to a reduction in Absolute Trajectory Error (ATE).
    \item Hand Reconstruction: Removing the continuous ViT degrades all three metrics on the TACO benchmark, confirming that unquantized visual features are essential for fine-grained 3D hand pose recovery.
\end{itemize}

Overall, the continuous $\mathbf{x}_{\text{ViT}}$ stream successfully bypasses the quantization bottleneck to preserve the essential high-frequency spatial context required across all modalities.

\subsection{VQ-VAE Ablation}
We use the body module as an example to show the choices of VQ-VAE. Increasing codebook size (512, 1024, 2048) progressively reduces reconstruction error (11.06, 8.67, 8.26\,mm), while expanding to 4096 degrades accuracy (9.02\,mm) due to codebook utilization inefficiency. For spatial-temporal convolution, our $[2,3]$ configuration (8.26\,mm) outperforms more aggressive spatial ($[2,7]$, 15.47\,mm) or temporal ($[4,3]$, 13.05\,mm) compressions, achieving the optimal balance between reconstruction fidelity and generalizability.

\end{document}